%% file: main.tex
\documentclass[10pt,twocolumn,letterpaper]{article}

\usepackage[pagenumbers]{cvpr} 

\input{preamble}
\definecolor{cvprblue}{rgb}{0.21,0.49,0.74}
\usepackage[pagebackref,breaklinks,colorlinks,allcolors=cvprblue]{hyperref}

\def\paperID{*****} 
\def\confName{CVPR}
\def\confYear{2026}

\title{Personalized Generative Models for Contextual Debiasing}

\author{
Xinran Liang$^{1}$, 
Esin Tureci$^{1}$, 
Prachi Sinha$^{1}$,
Ye Zhu$^{12}$, 
Vikram V. Ramaswamy$^{1}$, 
Olga Russakovsky$^{1}$ \\
$^{1}$Department of Computer Science, Princeton University \hspace{5mm}
$^{2}$LIX, CNRS, École Polytechnique
}

\begin{document}
\maketitle

\newcommand{\proj}[1]{DecoupleGen}

\newcommand{\esin}[1]{\textcolor{blue}{[Esin: #1]}}

\begin{abstract}
Different visual patterns appear with different frequencies in the world: e.g., beach balls appear on sand more often than they do on a road. These statistics are reflected in vision datasets, and as a result trained models more easily recognize objects in common scenarios. However, recognizing a beach ball on a road may arguably be even more important than recognizing it on sand. We study how to mitigate this discrepancy. Since collecting uncommon images in the real world may be difficult, we explore whether generating images with less frequent contexts can serve as effective training augmentation. A key challenge is guiding generations to remain close to the original dataset distribution while creating diverse images with uncommon contexts. We introduce \textbf{Decoupl}ing Cont\textbf{e}xtual Patterns with \textbf{Gen}erations (\textbf{DecoupleGen}), a method that personalizes text-to-image diffusion models to facilitate coherent synthesis of images with rare contexts while preserving original visual details. The generated images contain semantically meaningful content and remain visually aligned with the original datasets. We further apply verification constraints to ensure relevance of the augmented data. We evaluate our approach on object classification and recognition tasks on complex scene datasets. Our experiments demonstrate consistent improvements over previous approaches, and our analyses identify factors underlying these improvements.
\end{abstract}   
\vspace{-0.5cm}

\iftrue
\input{8page_sections/introduction}

\input{8page_sections/related_work}

\input{8page_sections/pipeline}
\input{8page_sections/experiments}

\input{8page_sections/discussion}
\fi

\section*{Acknowledgment}
This work is supported by NSF Career Award \#2145198 to O.R.. Any opinions, findings, and conclusions or recommendations expressed in this material are those of the authors’ and do not necessarily reflect the views of the National Science Foundation. We would like to thank Princeton VisualAI Lab for discussions and feedback.

{
    \small
    \bibliographystyle{ieeenat_fullname}
    \bibliography{main}
}

\input{8page_sections/appendix}

\end{document}

%% file: 8page_sections/introduction.tex
\section{Introduction}
\label{sec:intro}

Improvements in computer vision models are fueled, in part, by advances in data collection methods, resulting in ever-larger datasets. However, despite diverse curation strategies, these datasets still contain multiple forms of bias~\cite{buolamwini2018gender, shankar2017no, birhane2021large, van2016stereotyping, zhao2021understanding}. One way these biases can manifest is through common co-occurring patterns; for example, skis frequently appear with people, while images of skis alone are rare. 
These \emph{contextual biases} often cause models to fail to generalize to uncommon scenarios; going back to our example, these models often have low accuracy on images of skis without people~\cite{singh2020don}. 
While recent general-purpose vision-language models may exhibit greater robustness to such biases due to large-scale pretraining~\citep{bai2025qwen3,singh2025openai}, many real-world and industrial applications still utilize smaller specialized models given compute and memory constraints~\citep{liao2022kitti,caesar2020nuscenes}, motivating the need to improve generalization of compact classifiers.

\begin{figure}[!t]
    \centering
        \begin{subfigure}[b]{0.24\linewidth}
             \centering
             \includegraphics[width=\linewidth]{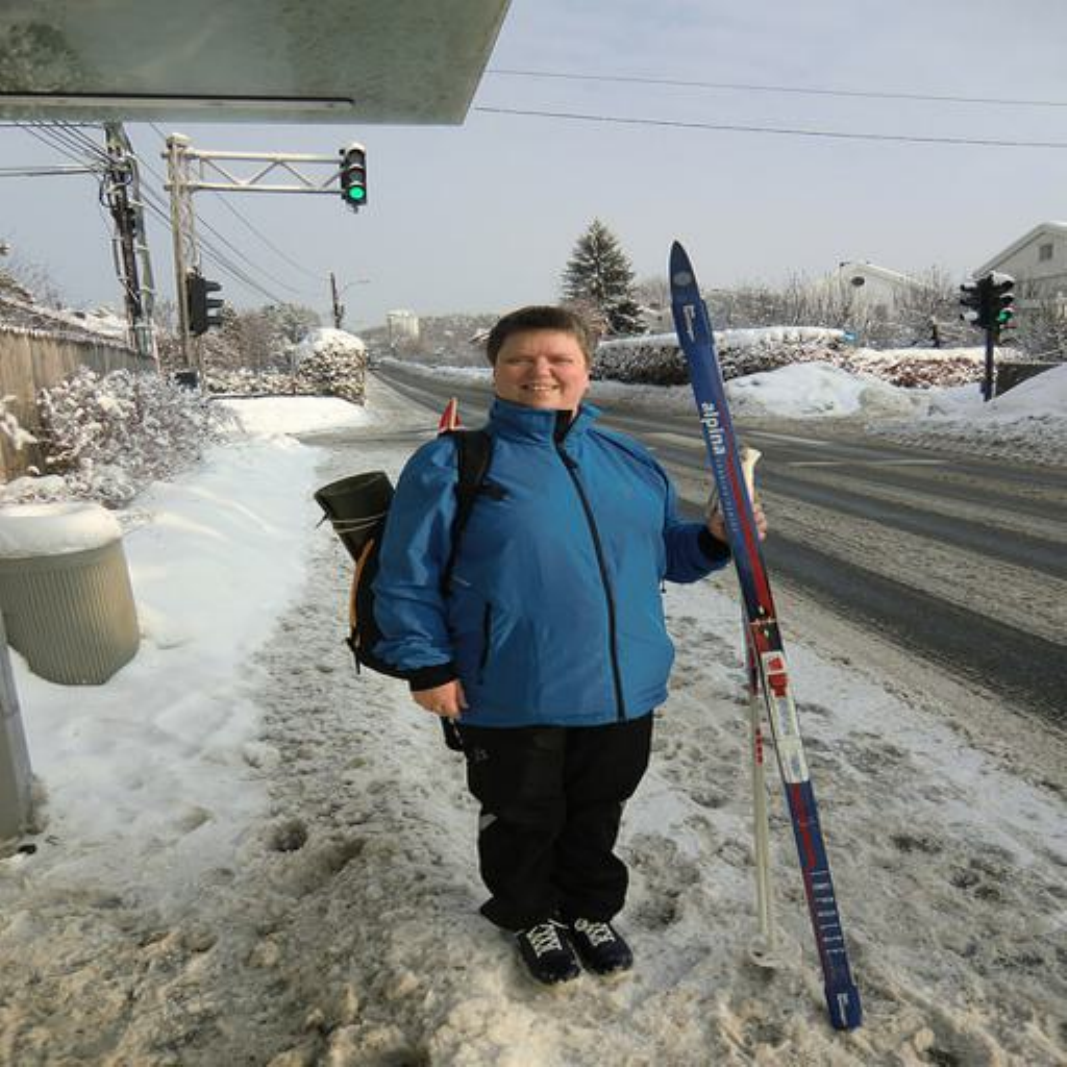}
             \caption{Original}
             \label{fig:pipe_hard_orig}
         \end{subfigure}
        \hspace{-0.02\linewidth}
        \begin{subfigure}[b]{0.24\linewidth}
             \centering
             \includegraphics[width=\linewidth]{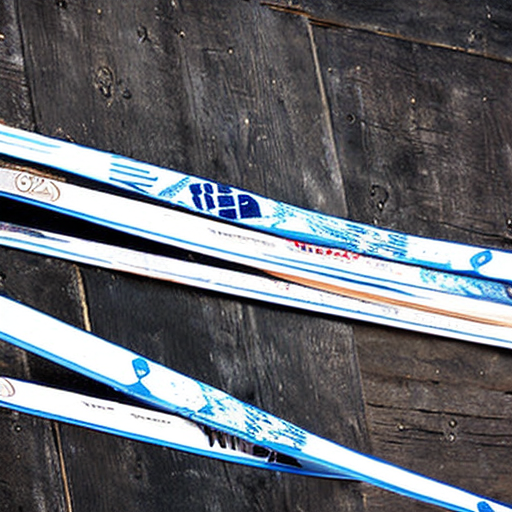}
             \caption{Txt2Img~\citep{rombach2022high}}
             \label{fig:pipe_hard_txt2img}
         \end{subfigure}
         \hspace{-0.02\linewidth}
        \begin{subfigure}[b]{0.24\linewidth}
             \centering
             \includegraphics[width=\linewidth]{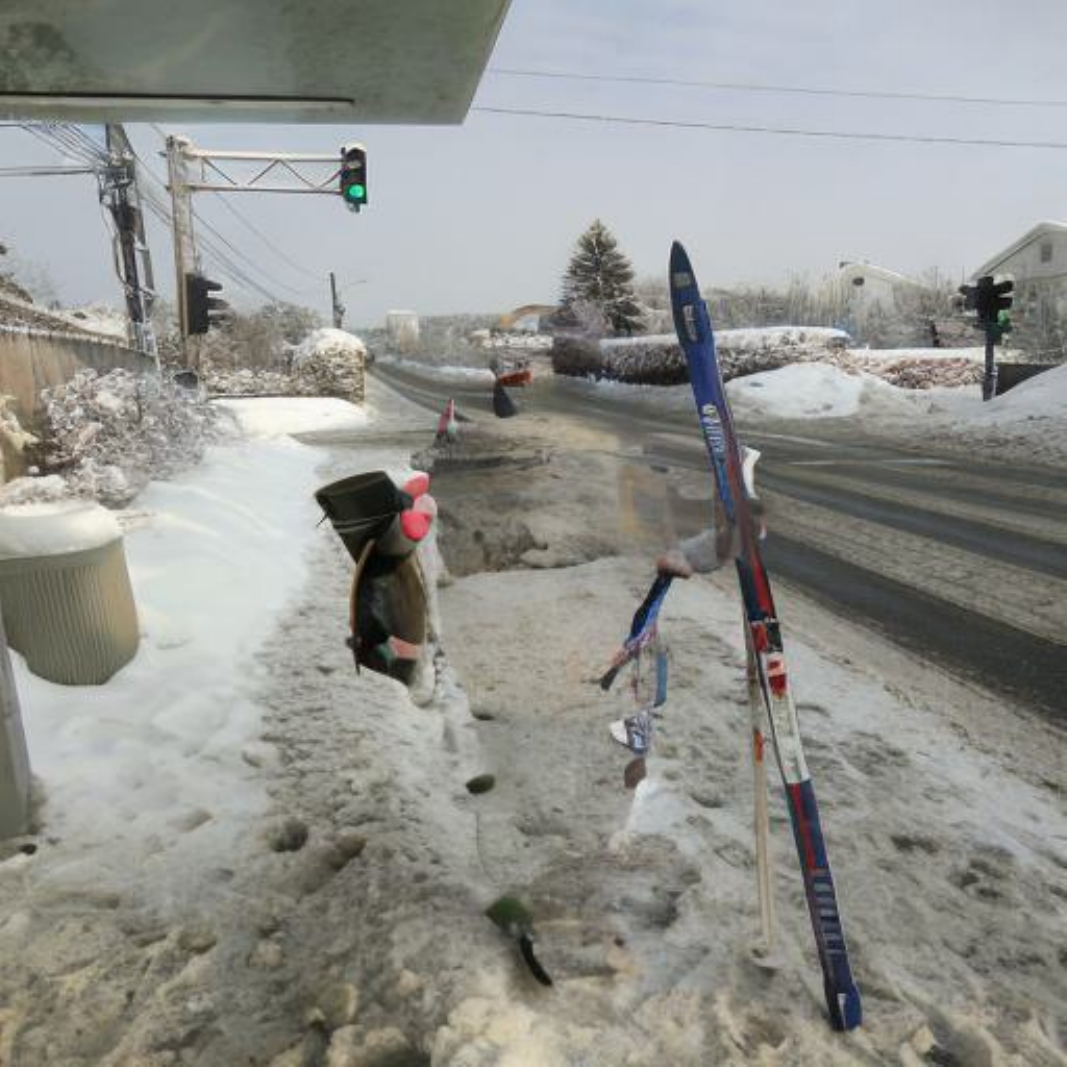}
             \caption{Inpaint~\citep{rombach2022high}}
             \label{fig:pipe_hard_inpaint}
         \end{subfigure}
        \hspace{-0.02\linewidth}
        \begin{subfigure}[b]{0.24\linewidth}
             \centering
             \includegraphics[width=\linewidth]{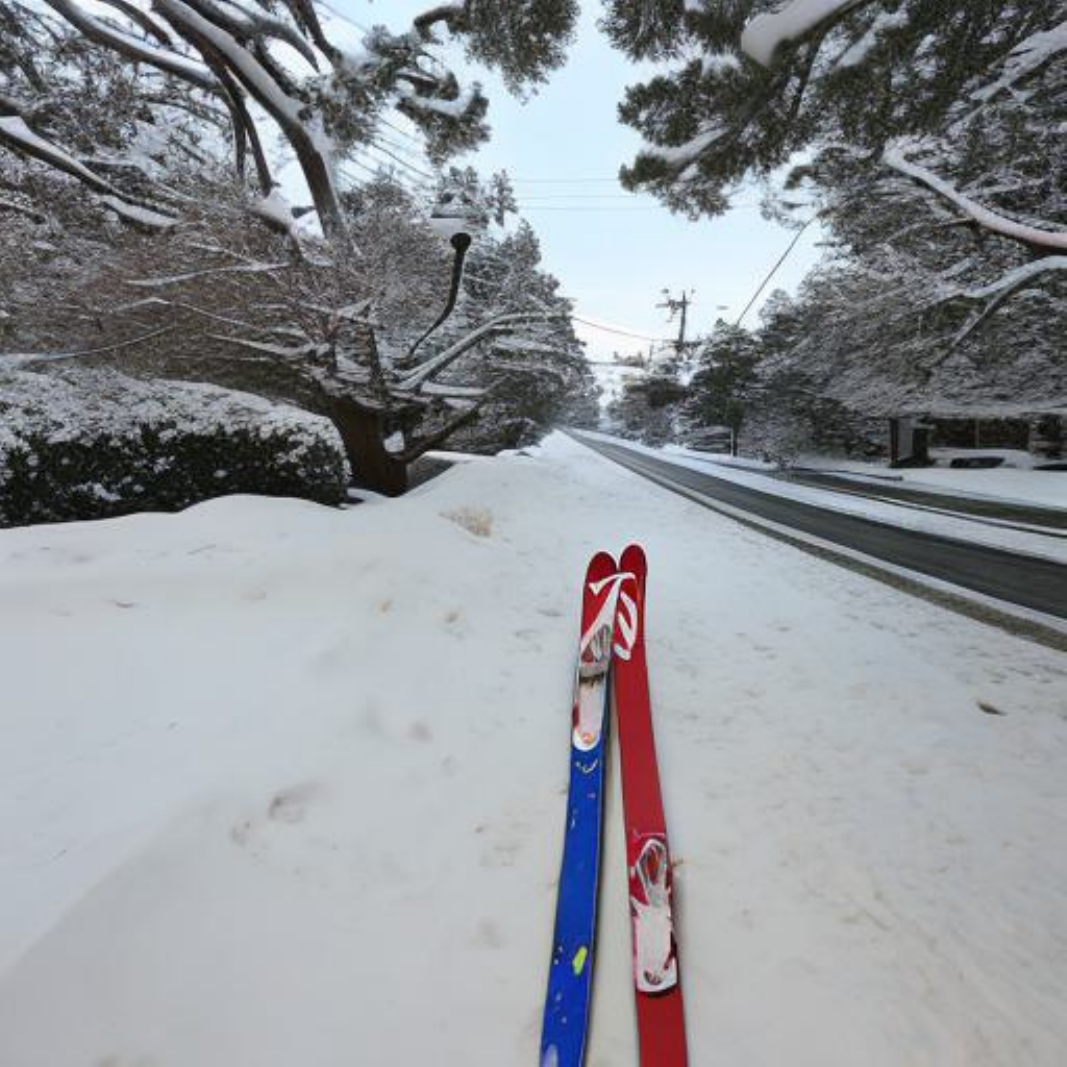}
             \caption{DecoupleGen}
             \label{fig:pipe_hard_ours}
         \end{subfigure}
        \hspace{-0.02\linewidth}
    \caption{\textbf{An example illustrating challenges in contextual debiasing.}
    Our goal is to generate an image of skis without a person. Using common Txt2Img models typically produces images with visual appearances that differ substantially from those in the original dataset (Figure~\ref{fig:pipe_hard_txt2img}). Applying existing local image editing methods, e.g. inpainting, often generates less plausible results (Figure~\ref{fig:pipe_hard_inpaint}). Modifying other elements related to person is necessary to ensure visual coherence of the entire image (Figure~\ref{fig:pipe_hard_ours}).}
    \label{fig:pipe_hard}
    \vspace{-5mm}
\end{figure}

One way to address these biases is through training objectives, such as reweighting, that penalize models for over-relying on contextual cues~\citep{singh2020don, cui2019class,bickel2009discriminative, fernandez2018cost}. However, these methods are limited to using only the original datasets and can often result in poorer performance on in-context samples. 
An alternative way to tackle these biases is to leverage recent advances in generative models~\cite{rombach2022high, ho2020denoising, nichol2021glide, ramesh2021zero, hertz2022prompt} to synthesize images with diverse content. These models can be used to edit images to remove common contexts, or generate images with difficult contexts. Recent works explore the potential to incorporate these generations as data augmentation to improve recognition performance. However, they usually focus on classifying a single object in an image~\citep{azizi2023synthetic, he2022synthetic, dunlap2024diversify, hemmat2023feedback}, thus are not well equipped to handle interactions between multiple objects in real world scenarios, limiting their potential scope and applicability.

Constructing images that contain objects in uncommon contexts using generative models can be challenging. Since objects and scenes usually interact with each other in real images, manipulating contextual patterns requires handling their complex relationships in order to generate natural looking images.
Furthermore, in order for these samples to improve model performance, the resulting generations should ideally remain close to the distribution of the original dataset. Consider the COCO~\cite{caesar2018coco} dataset and, in particular, images containing skis, the person within the image usually interacts with the skis in some way (Figure~\ref{fig:pipe_hard_orig}). When generating images with rare contexts such as skis without people, pre-trained text-to-image (Txt2Img) models often produce images with viewpoints and backgrounds backgrounds that differ substantially from the original dataset (Figure~\ref{fig:pipe_hard_txt2img}), which may not be relevant to downstream tasks. On the other hand, local editing methods, e.g. inpainting~\citep{rombach2022high, birhane2021large}, which takes mask as input to explicitly remove or replace the context, are unable to deal with interactions among objects. Removing the person in this manner results in the skis floating (Figure~\ref{fig:pipe_hard_inpaint}).
Instead, generating a plausible version of the image without the person requires repositioning the skis as well, as shown in Figure~\ref{fig:pipe_hard_ours}.

To address this issue, we introduce \textbf{Decoupl}ing Cont\textbf{e}xtual Patterns with \textbf{Gen}erations (\textbf{DecoupleGen}) (Figure \ref{fig:pipe_main}). We adapt pre-trained diffusion models to downstream datasets by fine-tuning them on images containing common contextual patterns, and then generate samples containing uncommon contextual patterns. Specifically, inspired by personalization methods of diffusion models for subject-driven generations \citep{ruiz2023dreambooth, gal2022image}, we learn new word tokens that capture visual details from original images. We then prompt the fine-tuned diffusion models with the learned text embeddings to manipulate contextual patterns. The resulting generations are natural looking and visually similar to images from the original datasets. To obtain generations more relevant to downstream recognition tasks, we further select synthesized images that do not contain common contextual patterns, ensuring that the additional training samples capture underrepresented groups in the original datasets and supplement out-of-context samples.

We evaluate our method on visual recognition tasks using real-world object and scene datasets \citep{caesar2018coco, he2021towards}. Our experiments show that \proj{} outperforms existing approaches that modify loss functions or leverage generative models for data augmentation by 1.5\%, improving over standard classifier by 14.2\% on the NICO dataset and by 5.4\% on the COCO dataset. Our analyses show two distinctive advantages of \proj{}: first, synthesized images from our method remain closer to the downstream dataset, indicating better alignment; and second, our method is better accounts for complex relationships between objects and scenes depicted in natural images.\footnote{In this work, “personalization” refers to subject-driven visual generation techniques that extract and reproduce specific visual concepts from one or a few reference samples.} 

%% file: 8page_sections/related_work.tex
\section{Related Work}

\paragraph{Addressing model bias via dataset manipulations.} 
Prior work has shown that existing datasets are not fully representative of the real world~\citep{khosla2012undoing, tommasi2017deeper}, causing models trained on them to struggle to generalize to uncommon scenarios. We focus specifically on failures where vision models cannot recognize objects without their co-occurring contexts, a limitation observed in both deep neural network classifiers~\citep{choi2012context, singh2020don} and recent large vision-language models~\citep{li2023evaluating, li2025oric}.
One approach to address these limitations is to rebalance training datasets through methods such as re-weighting~\citep{jiang2020identifying, cui2019class}, which assigns higher loss weights to underrepresented samples, and over-sampling~\citep{bickel2009discriminative, fernandez2018cost}, which duplicates such samples during training. While these methods manipulate the effective training distribution using real images, we instead use synthetic samples generated by diffusion models to achieve similar goals. Unlike \citet{ramaswamy2021fair}, which generates additional samples for facial recognition, our method focuses on manipulating contextual patterns and is therefore more broadly applicable to real-world scene images.

\vspace{-0.5cm}
\paragraph{Leveraging generated data for downstream tasks.} 
With recent advances in generative models, using synthetic data for model training or evaluation is not entirely new. Previous works \citep{jahanian2021generative, liu2022palm, tian2024stablerep, fan2023scaling, tian2023learning} develop self-supervised representation learning methods using generations from GANs \citep{goodfellow2014generative} or diffusion models \citep{ho2020denoising}. Synthetic images and labels from generative models have been used effectively in providing diverse augmentations and improving model performance in classification tasks \citep{he2022synthetic, trabucco2023effective, azizi2023synthetic, dunlap2024diversify, hemmat2023feedback}. Generated images can be used to create test sets for classifier evaluations \citep{prabhu2024lance, zhang2024imagenet}. Our work falls into this category of approaches using generated data for model training. While these works perform generation or editing at the image level for classification or self-supervised tasks, our method instead considers relationships between multiple objects and scenes captured in real images during generation.

\vspace{-0.5cm}
\paragraph{Generative models for local image editing.} 
A natural approach for manipulating contextual patterns is to provide object masks to generative models such as inpainting~\citep{pathak2016context}. The task of local image editing has recently received significant attention and seen substantial progress~\citep{avrahami2022blended, lugmayr2022repaint, andonian2021paint, rombach2022high}. Consider an image of a person jumping on a skateboard in midair. While such models can feasibly modify local content -- for example, the jumping person -- they cannot adequately adjust related information in the remaining regions, such as the skateboard suspended in the air, due to the complex relationships between objects and scenes. In contrast, our method can generate more realistic images with physically plausible scenes (as discussed in Section~\ref{sec:reason_why}).

\begin{figure*}[!ht]
    \centering
    \includegraphics[width=0.8\textwidth]{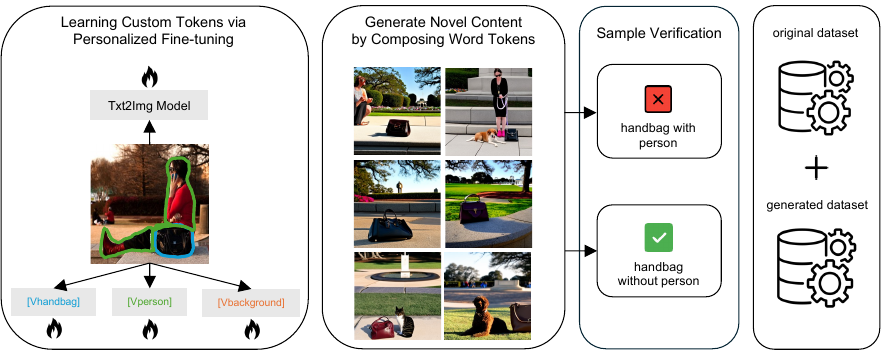}
    \caption{\textbf{Overview of DecoupleGen.}
    We first fine-tune text-to-image diffusion models on each co-occurring image by learning new word tokens to describe visual details. We then synthesize images by combining these learned text embeddings to manipulate contextual patterns. Our generations are visually coherent and maintain visual information similar to original image. Finally, we select relevant and meaningful generations via a verification process and incorporate them in classifier training. 
    }
    \label{fig:pipe_main}
    \vspace{-5mm}
\end{figure*}

\vspace{-0.5cm}
\paragraph{Finetuning diffusion models for personalization.}
Recent works, e.g., Textual Inversion~\citep{gal2022image}, DreamBooth~\citep{ruiz2023dreambooth}, and Custom Diffusion~\citep{kumari2023multi}, explore fine-tuning text-to-image diffusion models for personalization using a small set of images and learning new word embeddings. These methods reconstruct visual details of a single subject -- such as a person, animal, or object -- from one or a few training images and enable diverse generations through different text prompts. Unlike these approaches, which reconstruct the primary subject from training images, our goal is to preserve visual details from regions unrelated to the manipulated class categories and use them to generate new object-context combinations. Following \citet{avrahami2023break}, our method learns new word tokens to describe corresponding regions in the training image and reconstruct similar content during generation (as discussed in Section~\ref{pipe:gen}).

%% file: 8page_sections/pipeline.tex
\vspace{-2mm}
\section{Contextual Debiasing via Generative Model}

We aim to augment the original dataset with generations that do not contain common contextual patterns, while encouraging these generations to be visually coherent and close to the original dataset distribution. To achieve this goal, we perform text-to-image generations while maintaining visual information by adapting personalization methods for diffusion models on individual samples.

We propose \textbf{Decoupl}ing Cont\textbf{e}xtual Patterns with \textbf{Gen}erations (\textbf{DecoupleGen}) and provide an overview of the approach in 
Figure \ref{fig:pipe_main}.
Our method consists of three steps: Starting from real images with common contextual patterns, we first fine-tune text-to-image diffusion models to learn new word tokens that describe different image components (biased object, contextual object, and background). We then guide generation process to create images without common contextual patterns by composing these learned word tokens. Finally, we filter the synthesized images using a pretrained object detector to remove samples that may still contain common contextual patterns. 
We augment the real dataset with generated images for classifier training.

\subsection{Problem Formulation: Image Generation}


We aim to teach the model to separate biased object classes \texttt{b} (e.g., skateboard) from their contextual categories \texttt{c} (e.g., person) by generating images that remove common co-occurrence patterns. Specifically, we formulate contextual debiasing as editing images containing \texttt{(b, c)} by either removing \texttt{c} or replacing it with alternative semantic categories while preserving the remaining visual content.

\noindent \textbf{What is hard about this problem? }
Images encode complex object–scene relationships, making direct edits prone to producing implausible results (e.g. a floating skateboard). Meanwhile, off-the-shelf Txt2Img models often fail to preserve viewpoint and other image-specific details. Our goal is therefore to generate images that reposition objects while maintaining the original visual characteristics.

\subsection{Adapting Personalized Diffusion Models} \label{pipe:gen}

We take inspiration from recent personalization methods \citep{ruiz2023dreambooth, gal2022image, kumari2023multi} that fine-tune special tokens to reconstruct subjects. Specifically, we learn a new ``background'' token that captures visual details of the remaining pixels.

\textit{At training time}, given an image with masks of \texttt{b} and \texttt{c}, we learn a new \texttt{[Vbackground]} token to reconstruct visual details specified by remaining pixels not belonging to either \texttt{b} or \texttt{c}. We use the masks of \texttt{b} and \texttt{c} as input, while the background mask is defined as all remaining pixels outside the two classes. The word tokens and the diffusion model are optimized to reconstruct the image from the text prompt ``a photo of \texttt{[Vhandbag]} and \texttt{[Vperson]} and \texttt{[Vbackground]}". Following \citet{avrahami2023break}, we use the training objective $\mathcal{L} = \mathcal{L}_{\text{maskrec}} + \lambda \mathcal{L}_{\text{crossattn}}$, where
\begin{align*}
    \mathcal{L}_{\text{maskrec}} = \mathbf{E}_{z, m, \epsilon, t} 
    \left[ \Vert \epsilon \odot M_m - \epsilon_{\theta} (z, t, \texttt{V}_m ) \odot M_m \Vert ^2 \right]
\end{align*}
is the variational lower bound loss between the predicted noise $\epsilon_\theta$ and the added noise $\epsilon \sim \mathcal{N}(0, 1)$, where $z$ denotes the noisy latent at diffusion timestep $t$. This objective encourages faithful reconstruction of each masked region $M_m$ from the corresponding text token $\texttt{V}_m$. In addition, 
\begin{align*}
    \mathcal{L}_{\text{crossattn}} = \mathbf{E}_{z, m, t}
    \left[ \Vert \text{CrossAttn}_\theta (z, t, \texttt{V}_m) - M_m \Vert ^2 \right]
\end{align*}
is the cross attention loss that encourages each word token $\texttt{V}_m$ to attend only to its corresponding masked region $M_m$ and not to other regions.

\textit{At inference time}, we generate images from the finetuned diffusion model by composing the learned background token with standard text tokens describing object categories. The custom token \texttt{[Vbackground]} captures contextual visual details and scene appearance, while object categories are represented using regular text tokens. For example, to remove the common contextual class \texttt{person} from handbag images, we use the prompt ``a photo of handbag at \texttt{[Vbackground]}''. To replace the contextual class with another category, we use prompts such as ``a photo of handbag and dog at \texttt{[Vbackground]}'', where \texttt{dog} can be substituted with other alternative classes.

\textit{A verification step} on synthesized samples is helpful, since our approach for manipulating contextual patterns does not guarantee successful generations in all cases. After generation, we annotate each image using an off-the-shelf pre-trained semantic segmentation model \citep{jain2023oneformer} to predict class labels. We discard images that do not have the expected labels during sample selection. 
For example, given an image with both skateboard and person, successful generations containing skateboard but not person will be included. 
This verification process ensures that generated samples contain information useful for downstream tasks. 

Similarly, our pipeline can be applied to image classification tasks in which each image contains an object from a single class. Our goal in this case is to generate images that preserve the visual appearance of objects to the original image but place them in diverse background scenes. Accordingly, our finetuning process learns two new word tokens \texttt{[Vclass]} and \texttt{[Vcontext]}. The word tokens and diffusion models are optimized to reconstruct the image from the text prompt ``a photo of \texttt{[Vclass]} and \texttt{[Vcontext]}" with the same loss function as above. At inference time, we use prompts of the form ``a photo of \texttt{[Vclass]} in \texttt{context-name}" for generation. We use this technique in our experiments, although our pipeline can be integrated with other approaches that learns to reconstruct visual details from masked regions of a single image.

%% file: 8page_sections/experiments.tex
\vspace{-4mm}
\section{Experiments}

We aim to generate synthetic images to increase sampling of the underrepresented regions of the dataset, while keeping generations as close to the original training data distribution as possible. Our goal is to improve model accuracy on underrepresented groups while preserving the accuracy on well-represented groups. In this section, we first demonstrate that our method outperforms existing approaches that modify training objectives on original datasets, as well as approaches that leverage generative models to synthesize additional training samples (Section \ref{sec:baseline}). We then analyze factors contributing to the advantages of our method over alternative generation-based approaches (Section \ref{sec:reason_why}). 

\vspace{-0.5cm}
\paragraph{Implementation Details.}
For image generation, we use Stable-Diffusion-v2 \citep{rombach2022high} as the base model with default hyperparameters. 
For classifier training, we initialize a ResNet50 \citep{he2016deep} with ImageNet \citep{russakovsky2015imagenet} pre-trained weights and perform end-to-end training on each downstream dataset. For all classification models, we tune learning rate and weight decay hyperparameters based on validation performance and report the best results.

\subsection{Quantitative Comparisons} \label{sec:baseline}

\subsubsection{Image Classification}

\paragraph{Experimental setup.} 
We use NICO dataset \citep{he2021towards} for single-object image classification. The dataset contains 60 object classes across 6 contexts, with a highly imbalanced distribution of object-context combinations (which we treat as groups). Our goal is to improve performance on underrepresented groups. We apply our pipeline independently to each training image, learning image-specific \texttt{[Vclass]} and \texttt{[Vcontext]} tokens during fine-tuning. At inference time, we preserve the learned \texttt{[Vclass]} token while replacing \texttt{[Vcontext]} with alternative textual context descriptions to generate new object-context combinations.

Following previous work \citet{hemmat2023feedback}, we synthesize images such that the combined dataset is balanced across groups, resulting in a total of 229k synthetic samples consistently used across different methods. We do not further apply verification constraints, since we found that samples generated by all methods considered in our experiments are visually plausible for single-object generation.

We compare our method against classifiers trained with different categories of learning algorithm \citep{yang2023change}: (1) subgroup robust methods: \textbf{GroupDRO} \citep{sagawa2019distributionally}, \textbf{LISA} \citep{yao2022improving}; (2) domain invariant learning: \textbf{IRM} \citep{arjovsky2019invariant}; (3) imbalanced learning: \textbf{BSoftmax} \citep{ren2020balanced}, \textbf{CRT} \citep{kang2019decoupling}; and (4) data augmentation: \textbf{Mixup} \citep{zhang2017mixup}. We use benchmarking results reported in previous work \citep{yang2023change}.
We also compare to models that utilize generative models to obtain additional training samples: \textbf{Txt2Img} with captions formatted as ``a photo of \texttt{class-name} in \texttt{context-name}" as text prompt, and \textbf{Feedback Guidance} \citep{hemmat2023feedback} that uses the same captions as Txt2Img and additional entropy guidance from a standard classifier. We implemented both methods and our reproduced results match those reported in previous work \citep{hemmat2023feedback}.

To evaluate model performance, we use overall accuracy (\textbf{Acc}) and worst group accuracy (\textbf{WGA}) following previous work \citep{yang2023change, hemmat2023feedback}. Overall accuracy is computed over all samples, while worst-group accuracy is computed over all subgroup accuracies.

\vspace{-0.5cm}
\paragraph{Experimental results.} 
Our results in Table \ref{tab:base_nico} show that \proj{} either matches or outperforms existing approaches on NICO dataset, while also improving upon the overall accuracy of the standard classifier, which previous methods failed to achieve.
\proj{} achieves an overall accuracy of $85.8\% \pm 0.1$, compared to $85.3\% \pm 0.3$ for the standard classifier, while matching the state-of-the-art worst-group accuracy of $49.2\%$, previously achieved by the feedback-guided generation approach~\citep{hemmat2023feedback}.

\begin{table}[h]
    \centering
    \resizebox{\linewidth}{!}{
    \begin{tabular}{lcccc}
        \toprule
        Method & Num Generations & Acc & WGA \\
        \midrule
         Standard & 0 & 85.3 $\pm$ 0.3 & 35.0 $\pm$ 4.1 \\
         Mixup \citep{zhang2017mixup} & 0 & 84.0 $\pm$ 0.6 & 42.7 $\pm$ 1.4 \\
         GroupDRO \citep{sagawa2019distributionally} & 0 & 83.2 $\pm$ 0.4 & 37.8 $\pm$ 1.8 \\
         LISA \citep{yao2022improving} & 0 & 84.7 $\pm$ 0.3 & 42.7 $\pm$ 2.2 \\
         IRM \citep{arjovsky2019invariant} & 0 & 84.4 $\pm$ 0.7 & 40.0 $\pm$ 0.0 \\
         BSoftmax \citep{ren2020balanced} & 0 & 84.0 $\pm$ 0.5 & 40.4 $\pm$ 0.3 \\
         CRT \citep{kang2019decoupling} & 0 & 85.2 $\pm$ 0.3 & 43.3 $\pm$ 2.7 \\
         \midrule
         Txt2Img \citep{rombach2022high} & $229k$ & 85.2 $\pm$ 0.1 & 33.5 $\pm$ 1.1 \\
         Feedback Guidance \citep{hemmat2023feedback} & $229k$ & 85.3 $\pm$ 0.3 & \textbf{49.2} $\pm$ 1.0 \\
         DecoupleGen (ours) & $229k$ & \textbf{85.8} $\pm$ 0.1 & \textbf{49.2} $\pm$ 0.7 \\
         \bottomrule
    \end{tabular}}
    \caption{\textbf{Quantitative comparisons on NICO dataset.}
    Our method achieves competitive or better performance in WGA compared to previous work, without sacrificing in accuracy.}
    \label{tab:base_nico}
    \vspace{-5mm}
\end{table}

\subsubsection{Object Recognition}

\paragraph{Experimental setup.} 
We use COCO-Stuff \citep{caesar2018coco} dataset, containing complex real-world images, for multi-object recognition. \citet{singh2020don} defined a bias score to identify biased categories and their co-occurring context; we use the same set of biased classes for evaluation. Our goal is to improve model performance on images in which biased classes appear without their common contextual labels.

To generate samples that manipulate contextual patterns, we follow the semantic label hierarchy of the original dataset to identify sets of similar categories within the same superclass for replacement. We generate 1 output from each replacement or removal prompt to avoid duplicates and encourage diversity in generated semantic content. We then consider a generation successful if its annotation labels contain only \texttt{b} (target class) without \texttt{c} (common context label), and retain only the successful generations. This process results in 96k generated samples in total.

We evaluate classifier performance using average precision (AP) for each class. We consider several evaluation metrics following previous work \citep{singh2020don}: (1) \textbf{Exclusive mAP} computed over 20 biased classes on images where  \texttt{b} \textit{never} co-occurs with  \texttt{c}; (2) \textbf{Cooccur mAP} computed over 20 biased classes on images where  \texttt{b} \textit{always} co-occurs with  \texttt{c}; (3) \textbf{Unbiased mAP} computed over the remaining 60 object classes in COCO; and (4) \textbf{All mAP} is computed over all 171 object and context classes.

\begin{table*}[!ht]
    \centering
    \resizebox{0.9\linewidth}{!}{
    \begin{tabular}{lccccc}
    \toprule
        Method & Num Generations & Exclusive mAP & Cooccur mAP & Unbiased mAP & All mAP \\
        \midrule
        Standard & 0 & 24.10 $\pm$ 0.13 & \textbf{64.80} $\pm$ 0.08 & \textbf{72.54} $\pm$ 0.04 & \textbf{55.97} $\pm$ 0.02 \\
        Real Cooccur & 0 & 22.69 $\pm$ 0.53 & 64.00 $\pm$ 0.57 & 71.28 $\pm$ 0.24 & 54.82 $\pm$ 0.31 \\
        Class balancing loss \citep{cui2019class} & 0 & 25.03 $\pm$ 0.06 & 64.67 $\pm$ 0.05 & 72.35 $\pm$ 0.04 & 55.79 $\pm$ 0.02 \\
        Negative penalty \citep{singh2020don} & 0 & 23.71 $\pm$ 0.03 & 64.61 $\pm$ 0.07 & 72.19 $\pm$ 0.05 & 55.68 $\pm$ 0.02 \\
        Weighted loss \citep{singh2020don} & 0 & \textbf{28.21} $\pm$ 0.15 & 59.89 $\pm$ 0.16 & 72.13 $\pm$ 0.01 & 55.44 $\pm$ 0.00 \\
        Remove cooccur labels \citep{singh2020don} & 0 & 24.17 $\pm$ 0.06 & 64.59 $\pm$ 0.06 & 71.59 $\pm$ 0.04 & 55.30 $\pm$ 0.01 \\
        Remove cooccur images \citep{singh2020don} & 0 & 28.09 $\pm$ 0.34 & 60.09 $\pm$ 0.08 & 72.25 $\pm$ 0.04 & 55.50 $\pm$ 0.04 \\
        CAM \citep{singh2020don} & 0 & 25.09 $\pm$ 0.23 & 64.45 $\pm$ 0.04 & 72.45 $\pm$ 0.04 & 55.87 $\pm$ 0.01 \\ 
        Feature split \citep{singh2020don} & 0 & 26.76 $\pm$ 0.52 & 64.10 $\pm$ 0.01 & 72.39 $\pm$ 0.07 & 55.74 $\pm$ 0.02 \\
        \midrule
        Txt2Img Cooccur \citep{rombach2022high} & $18k$ & 24.66 $\pm$ 0.36 & 64.41 $\pm$ 0.08 & 72.87 $\pm$ 0.14 & 56.11 $\pm$ 0.20 \\
        Txt2Img Exclusive \citep{rombach2022high} & $18k$ & 25.01 $\pm$ 1.04 & 65.26 $\pm$ 0.09 & 72.66 $\pm$ 0.02 & 55.96 $\pm$ 0.04 \\
        Txt2Img Target Class \citep{rombach2022high} & $18k$ & 23.90 $\pm$ 0.40 & 64.87 $\pm$ 0.15 & 72.44 $\pm$ 0.02 & 55.92 $\pm$ 0.02 \\
        SD Inpainting \citep{rombach2022high} & $18k$ & 25.34 $\pm$ 1.13 & 64.77 $\pm$ 0.09 & 72.61 $\pm$ 0.01 & 55.85 $\pm$ 0.05 \\
        Blended LD \citep{avrahami2022blended} & $18k$ & 24.87 $\pm$ 0.57 & 64.77 $\pm$ 0.05 & 72.71 $\pm$ 0.01 & 55.94 $\pm$ 0.01 \\
        Feedback Guidance \citep{hemmat2023feedback} & $18k$ & 25.55 $\pm$ 0.01 & 65.01 $\pm$ 0.06 & 72.68 $\pm$ 0.01 & 56.11 $\pm$ 0.01 \\
        DecoupleGen (ours) & $18k$ & \textbf{27.20} $\pm$ 0.32 & \textbf{65.56} $\pm$ 0.04 & \textbf{73.09} $\pm$ 0.03 & \textbf{56.37} $\pm$ 0.04 \\
        \midrule
        DecoupleGen (ours) & $96k$ & \textbf{29.52} $\pm$ 0.21 & \textbf{66.43} $\pm$ 0.04 & \textbf{73.52} $\pm$ 0.05 & \textbf{57.00} $\pm$ 0.06 \\
    \bottomrule
    \end{tabular}}
    \caption{\textbf{Quantitative comparisons on COCO dataset.}
    We compare the performance of \proj{} against methods that modify training objectives on the original dataset, where we include all successful generations from our method, totaling 96k for classifier training (top). We also compare \proj{} performance to different generation-based approaches, where we keep the number of additional samples for each target biased class consistent across methods, totaling 18k generations (bottom). \proj{} achieves the best performance across all evaluation metrics compared to existing approaches in both settings. The best result within each block is bolded.}
    \vspace{-5mm}
\label{tab:baseline_coco}
\end{table*}

\vspace{-0.5cm}
\paragraph{Comparison with loss-function based approaches.}
In addition to the \textbf{Standard} classifier that is trained on the original dataset with supervised loss, we compare our method against two sets of previous work. The first set includes approaches that modify training objectives on original datasets in order to improve performance on underrepresented groups. Following \citet{singh2020don}, we consider \textbf{Class balancing loss}, \textbf{Negative penalty}, \textbf{Weighted loss}, \textbf{Remove cooccur labels}, \textbf{Remove coocccur images}, \textbf{CAM}, which uses class activation map \citep{zhou2016cam} as weak localization supervision during classifier training, and \textbf{Feature split}, which enforces classifier to learn separate feature spaces for objects and their contexts. 

When comparing against these approaches utilizing alternative loss functions, we use all successful generations containing uncommon contextual patterns (96k) to augment the original training set in classifier training and report the results in Table~\ref{tab:baseline_coco}~(top half). 
Our method achieves better performance on all evaluation metrics compared to these loss-function based approaches.
In particular, we observe a $1.3\%$ improvement in exclusive mAP, increasing from $28.21\%$ with the re-weighting objective to $29.52\%$ with our method, as well as a $1.6\%$ improvement in cooccur mAP, increasing from $64.80\%$ for the standard classifier to $66.43\%$ with our method.
While previous approaches improve performance on exclusive images at the expense of cooccur, unbiased, and all mAP, our method achieves consistent improvements across all evaluation metrics.

\vspace{-0.5cm}
\paragraph{Comparison with other generation-based approaches.} Since we generates additional training samples for downstream tasks, we compare against several alternative generative approaches. We first consider \textbf{Txt2Img} with three variants, where captions describing the target biased classes (\textbf{Txt2Img Target Class}), co-occurring images (\textbf{Txt2Img Cooccur}), or exclusive images (\textbf{Txt2Img Exclusive}) are used to generate synthetic images~\citep{li2023blip, rombach2022high}. We also compare against \textbf{Feedback Guidance}~\citep{hemmat2023feedback}, which uses the same captions as \textbf{Txt2Img Target Class} together with guidance from a standard classifier, since classifier entropy reduces to classification loss in the multi-label classification setting. In addition, we compare against \textbf{Stable Diffusion Inpainting}, an inpainting model trained to reconstruct masked regions from the remaining unmasked areas~\citep{rombach2022high}, and \textbf{Blended Latent Diffusion}~\citep{avrahami2023blended}, where editing masks are applied in the down-sampled latent space and masked and unmasked regions are blended throughout the diffusion process.

To compare the effectiveness of different generation-based methods, we fix the number of generated samples added to each target biased class across all baselines, matching the number of co-occurring images in the original dataset (18k in total). This is motivated by the difficulty of successfully removing or replacing contextual classes using local image editing methods (discussed further in Section~\ref{sec:reason_why}). We apply a consistent filtering procedure across all methods considered in our experiments, retaining only successful generations.

Table \ref{tab:baseline_coco} shows that \proj{} improves over the standard method with 18k additional images, and outperforms other generative data augmentation approaches across all evaluation metrics. The largest improvement is observed in exclusive mAP: DecoupleGen improves over Feedback Guidance by $1.6\%$, achieving an mAP of $27.2\%$ compared to $25.6\%$. Thi improvement in exclusive mAP is especially noteworthy, indicating that our generations provide useful information for underrepresented groups in the dataset. 

\vspace{-1mm}
\subsection{Why are \proj{} Images Better?} \label{sec:reason_why}

Our goal in this section is to gain insight into the advantages provided by the personalized fine-tuning approach compared to alternative generative methods.
Through qualitative analysis, we first illustrate with examples that common local image editing models perform poorly on the contextual debiasing problem for complex object and scene images. We then analyze limitations of popular pre-trained Txt2Img models in ensuring that generations remain close to the downstream dataset distribution.

\begin{figure}[!ht]
    \centering
        \includegraphics[width=0.24\linewidth]{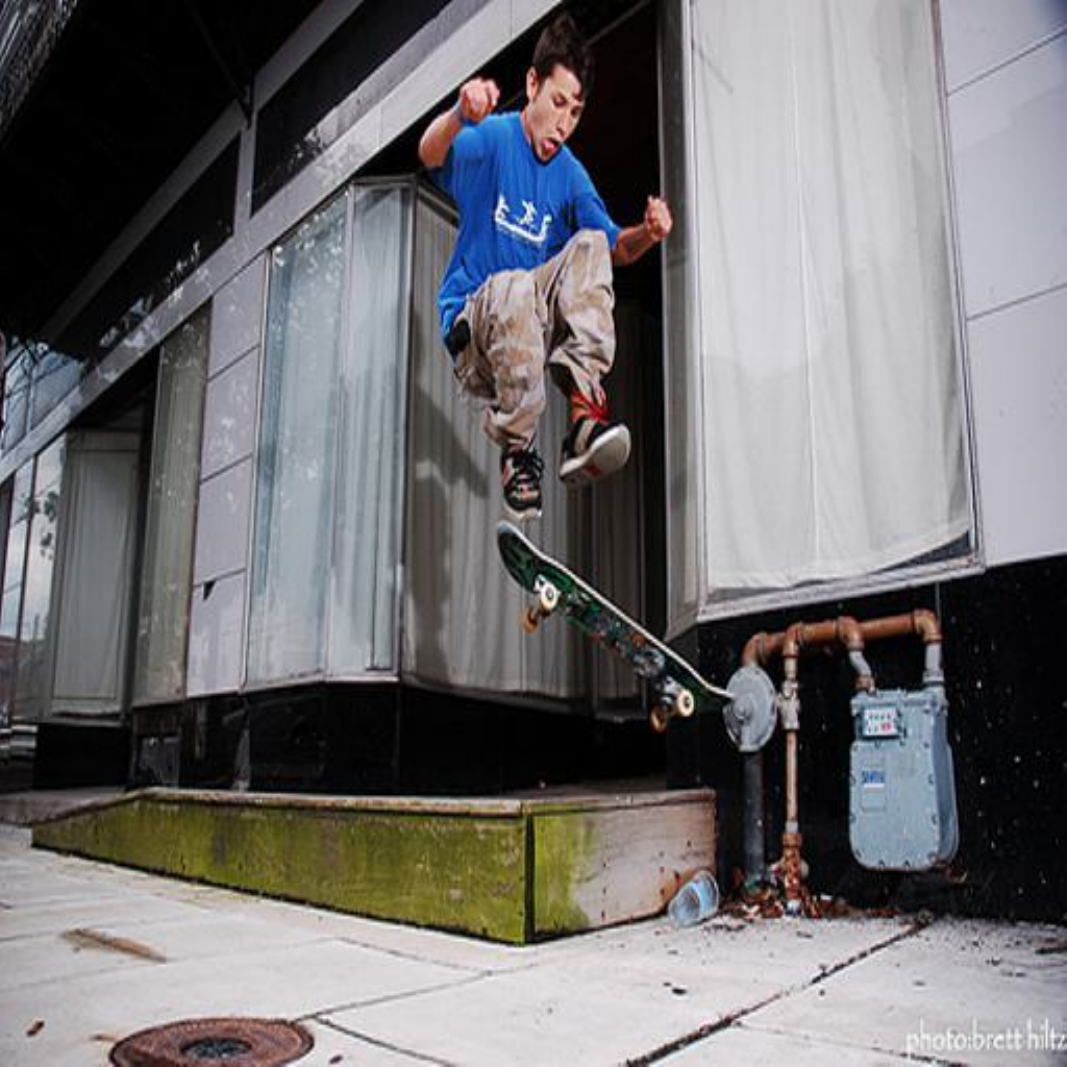}
        \hspace{-0.02\linewidth}
        \includegraphics[width=0.24\linewidth]{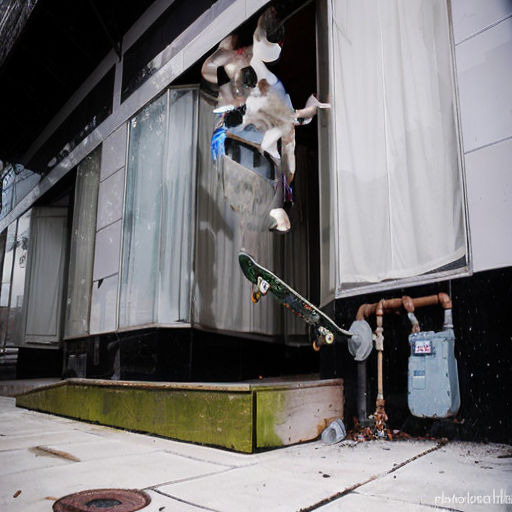}
        \hspace{-0.02\linewidth}
        \includegraphics[width=0.24\linewidth]{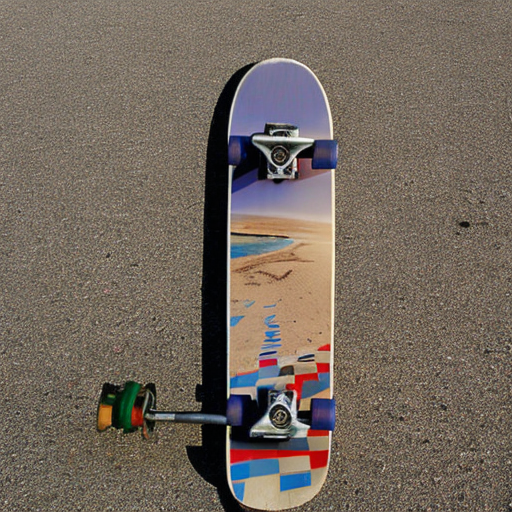}
        \hspace{-0.02\linewidth}
        \includegraphics[width=0.24\linewidth]{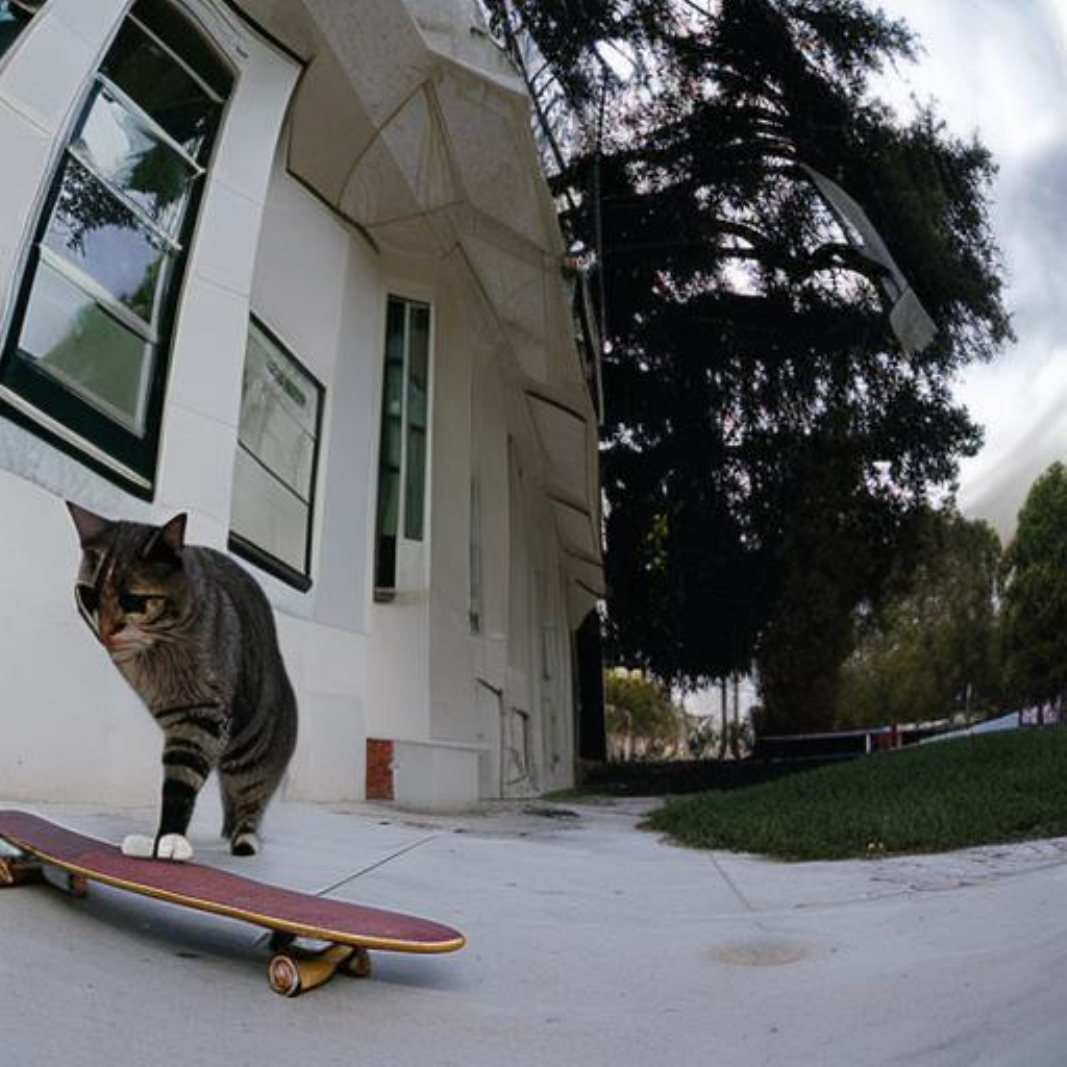}
        \hspace{-0.02\linewidth}
        \includegraphics[width=0.24\linewidth]{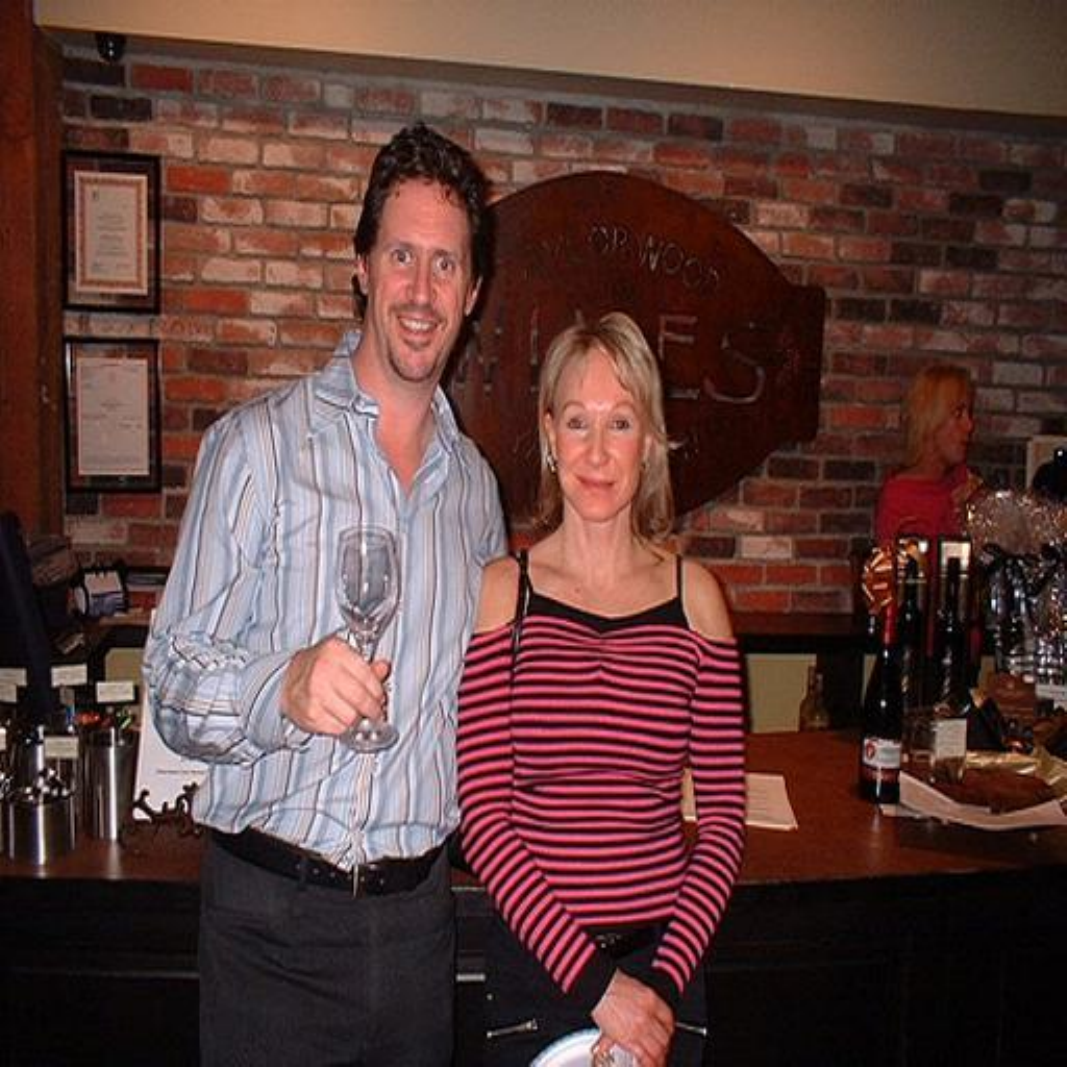}
        \hspace{-0.02\linewidth}
        \includegraphics[width=0.24\linewidth]{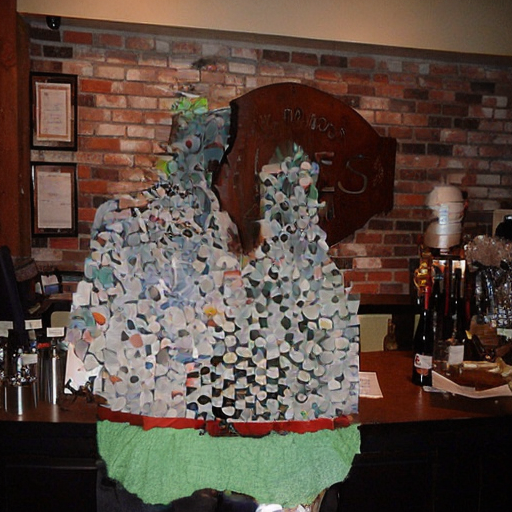}
        \hspace{-0.02\linewidth}
        \includegraphics[width=0.24\linewidth]{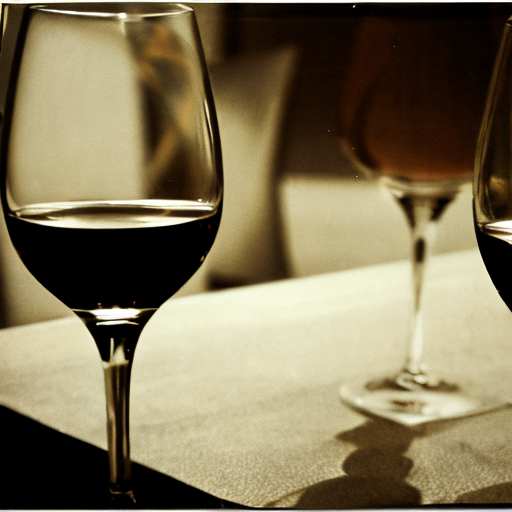}
        \hspace{-0.02\linewidth}
        \includegraphics[width=0.24\linewidth]{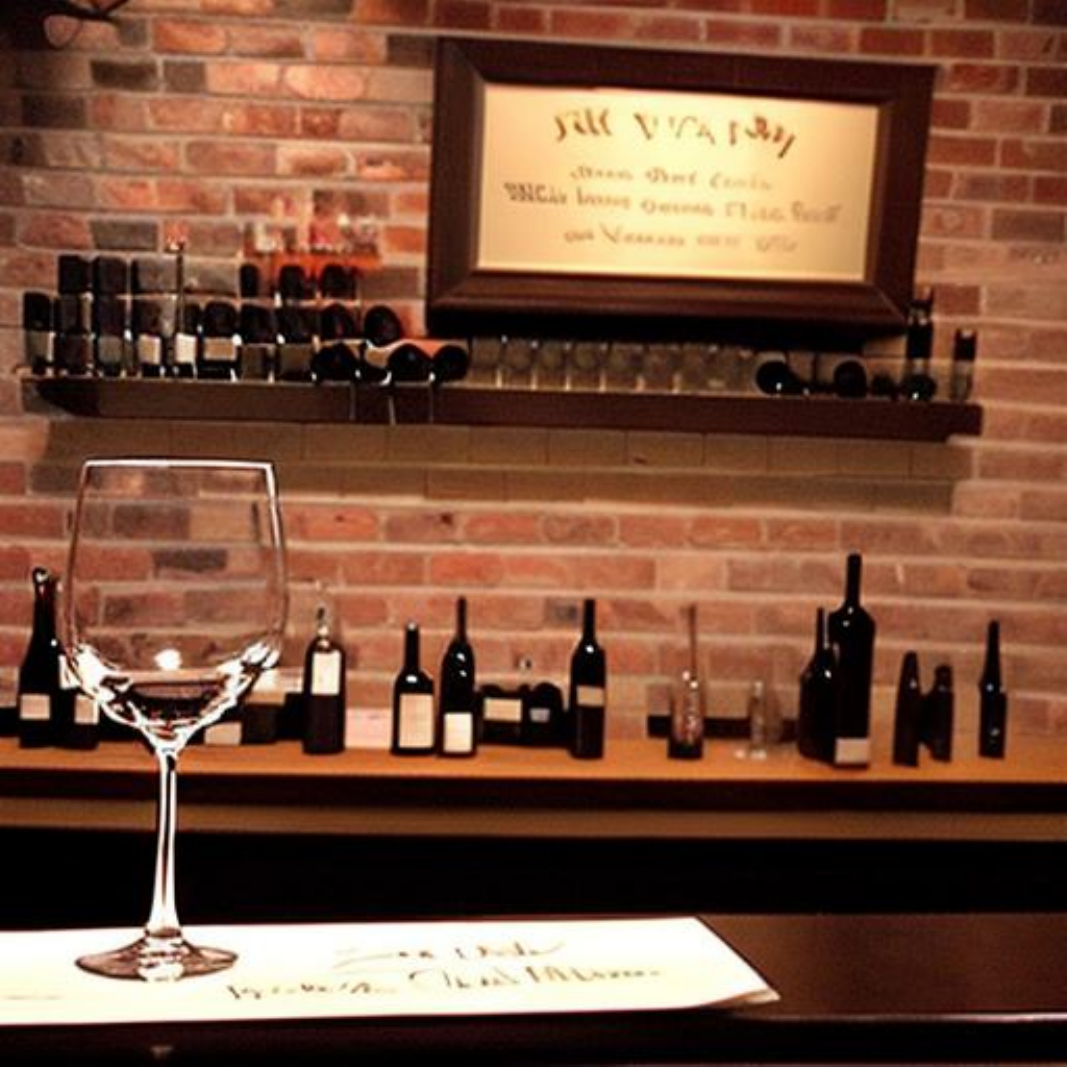}
        \hspace{-0.02\linewidth}
        \includegraphics[width=0.24\linewidth]{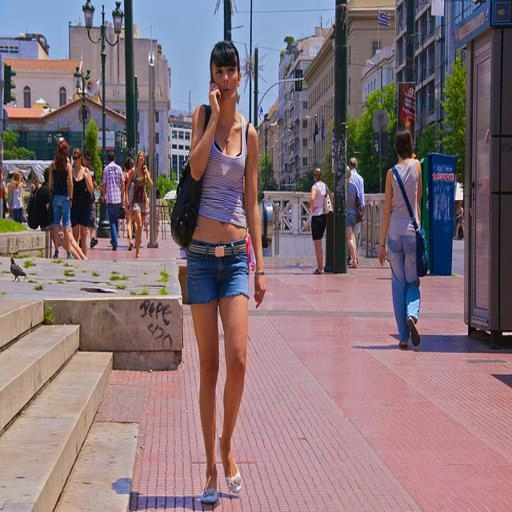}
        \hspace{-0.02\linewidth}
        \includegraphics[width=0.24\linewidth]{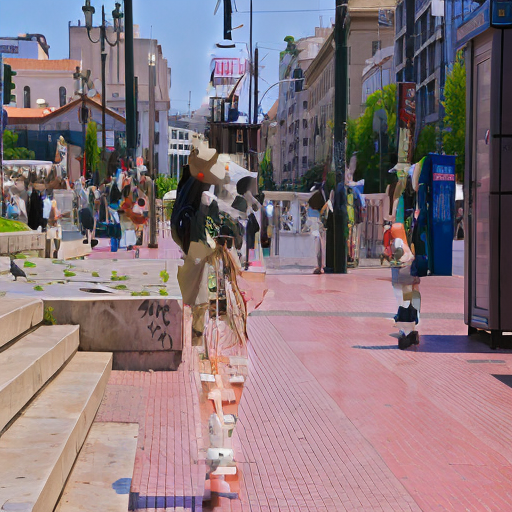}
        \hspace{-0.02\linewidth}
        \includegraphics[width=0.24\linewidth]{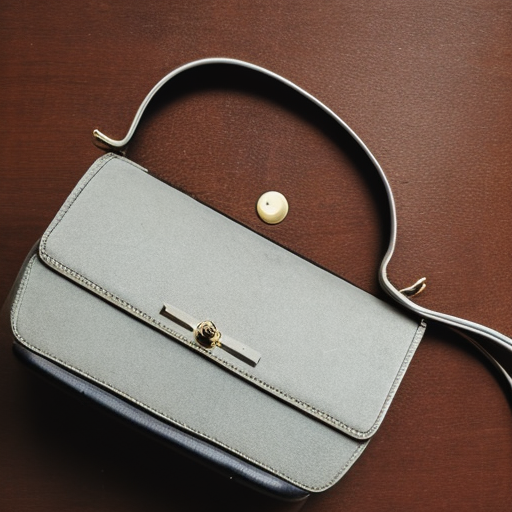}
        \hspace{-0.02\linewidth}
        \includegraphics[width=0.24\linewidth]{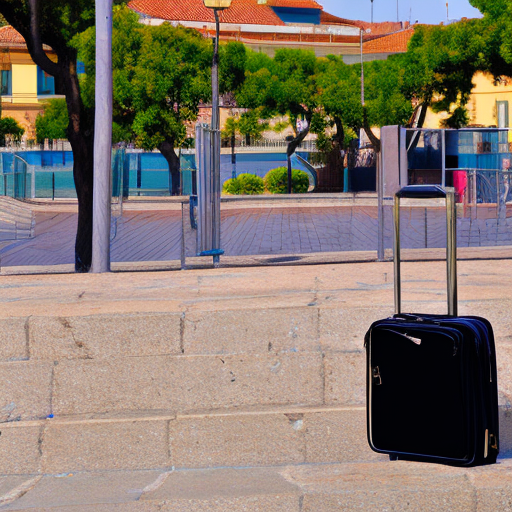}
        \hspace{-0.02\linewidth}
        \begin{subfigure}[b]{0.24\linewidth}
            \includegraphics[width=\linewidth]{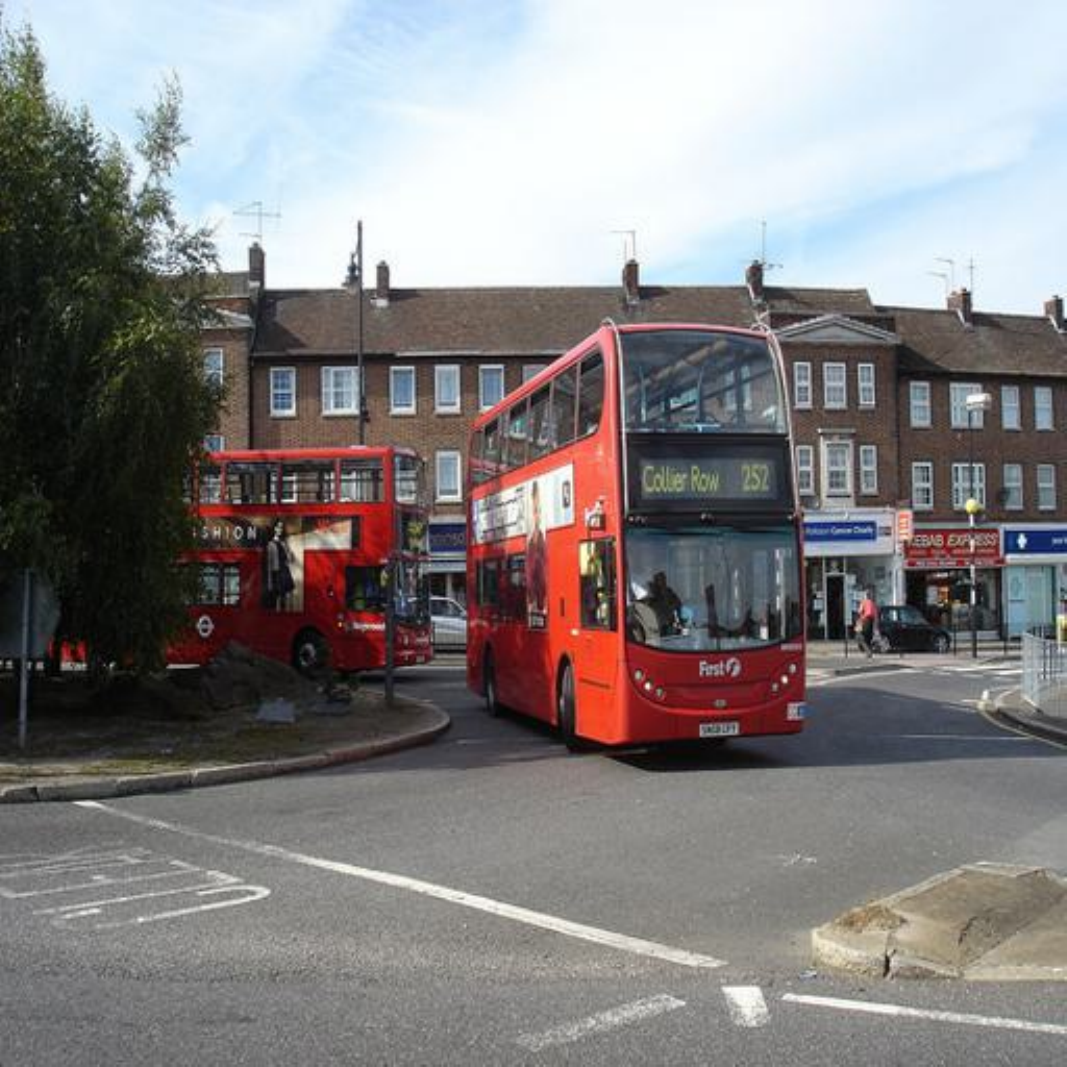}
            \caption{Original}
        \end{subfigure}
        \hspace{-0.02\linewidth}
        \begin{subfigure}[b]{0.24\linewidth}
            \includegraphics[width=\linewidth]{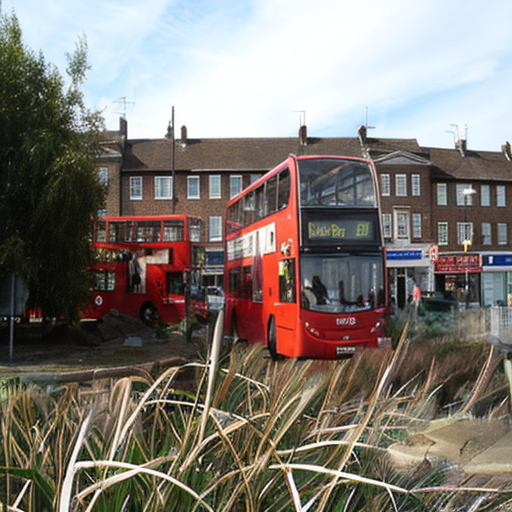}
            \caption{Blended LD}
            \label{fig:qualex_local}
        \end{subfigure}
        \hspace{-0.02\linewidth}
        \begin{subfigure}[b]{0.24\linewidth}
            \includegraphics[width=\linewidth]{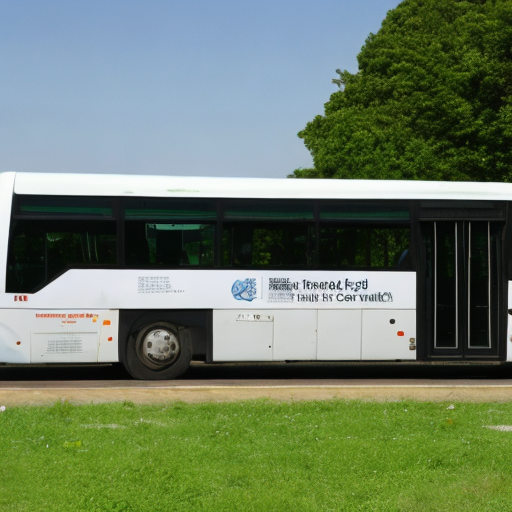}
            \caption{Txt2Img FG}
            \label{fig:qualex_txt2img}
        \end{subfigure}
        \hspace{-0.02\linewidth}
        \begin{subfigure}[b]{0.24\linewidth}
            \includegraphics[width=\linewidth]{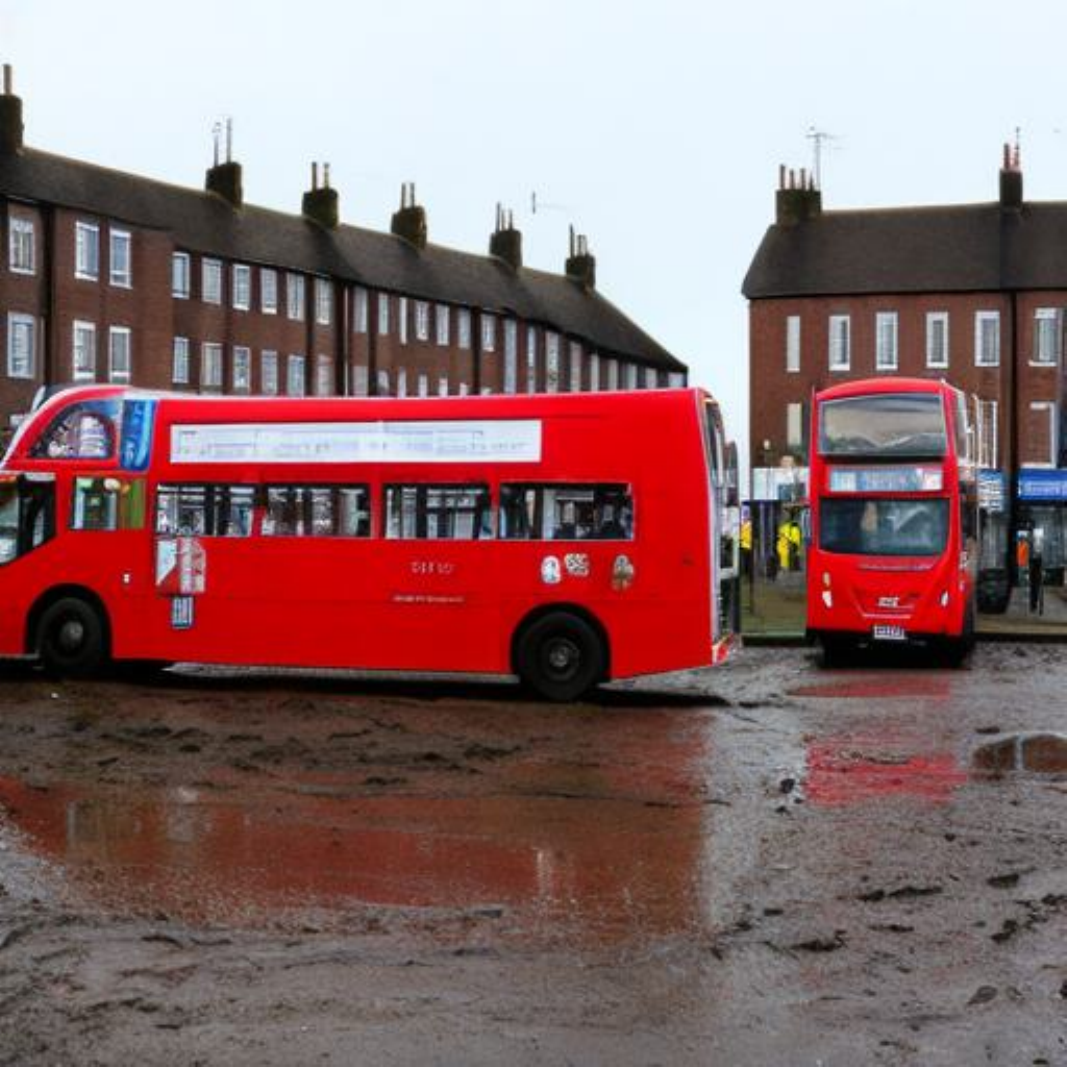}
            \caption{\proj}
            \label{fig:qualex_ours}
        \end{subfigure}
    \caption{\textbf{Qualitative comparisons.}
    Local image editing models, e.g. Blended LD \citep{avrahami2023blended}, often fail to keep generate reasonable and coherent images of complex scenes. Feedback-guided generations \citep{hemmat2023feedback} fail to preserve similar visual appearance in natural images. Instead, DecoupleGen can address issues when person interacts with objects in the image and when generation involves transforming into a different background scene. }
    \label{fig:qualex_main}
    \vspace{-6mm}
\end{figure}

\vspace{-0.5cm}
\paragraph{Limitations of local image-editing models.} 
We found that a major limitation of local image editing models is their low success rate. Through qualitative inspection, we observe that many generated images either contain visual artifacts caused by local modifications, resulting in lower image quality, or fail to produce semantically reasonable content that reflects natural scenes. Consequently, these generations are intuitively less helpful for downstream classifier performance. Figure~\ref{fig:qualex_main} shows several such examples.

When a person interacts with objects in the original image, modifying the person often requires repositioning the objects as well. For example, local image editing models may generate images in which a wine glass appears to float in the air (Figure~\ref{fig:qualex_local}), whereas our approach can place the wine glass onto different surfaces, producing physically more plausible images (Figure~\ref{fig:qualex_ours}). Similarly, when the generation process involves changing the background scene, local image editing models often fail to maintain global visual consistency. In contrast, our approach can modify contexts such as changing a road into mud while preserving image quality and avoiding the visual artifacts observed in alternative approaches.

\vspace{-0.5cm}
\paragraph{Pretrained Txt2Img models fail to adapt to downstream dataset distribution.}
We analyze synthetic images generated by \proj{} and by pre-trained Txt2Img models, including Txt2Img Target Class and Feedback Guidance. We compare their distributions to real samples from the validation split of the original dataset using UMAP~\citep{mcinnes2018umap}.

\begin{figure}[!ht]
    \centering
    \includegraphics[width=0.49\linewidth]{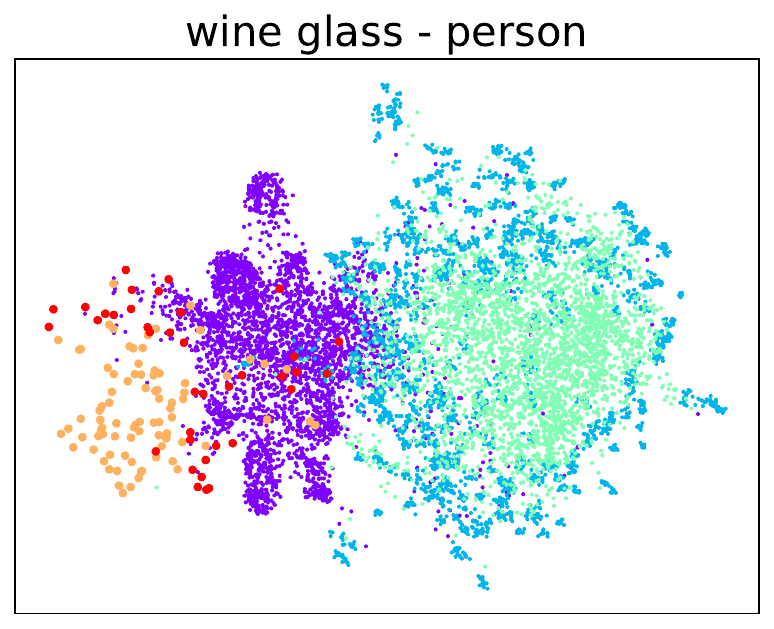}
    \includegraphics[width=0.49\linewidth]{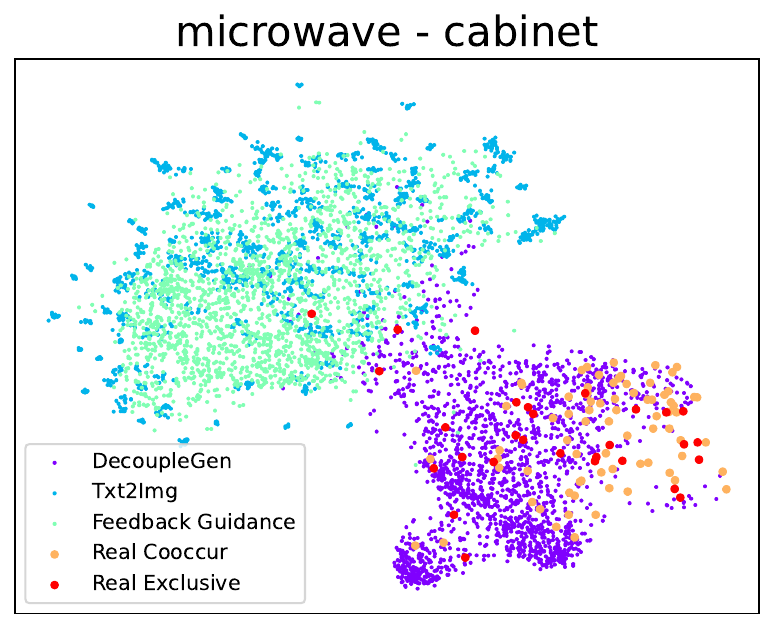}
    \caption{\textbf{UMAP visualizations of real samples and generations.}
    We plot real exclusive and cooccur images from validation split, and synthetic samples from different generation approaches. Generations from DecoupleGen overall stay closer to real data distribution than those from other methods, thus providing information more relevant to downstream tasks.}
    \label{fig:umap_coco_main}
    \vspace{-5mm}
\end{figure}

In Figure~\ref{fig:umap_coco_main}, we show UMAP projections for two representative object classes, \texttt{wine glass} and \texttt{microwave}, which have high co-occurrence with \texttt{person} and \texttt{cabinet}, respectively, in the real dataset. The UMAP distributions of the real dataset show that, for \texttt{wine glass}, exclusive and co-occurring images form relatively distinct clusters. For \texttt{microwave}, however, the exclusive and co-occurring images are less clearly separated.

In both cases, compared to \proj{} generations, Target Class and Feedback-Guided generations based on pre-trained Txt2Img models produce images that lie significantly farther from the real data distribution. This supports our intuition that, without constraining the generation process, images synthesized by pre-trained models can exhibit a substantial distribution gap from the real dataset in our downstream task. These results highlight the importance of the personalization-based fine-tuning process in \proj{}, which learns new word tokens to preserve visual details from real samples and thereby better adapts generative models to downstream datasets.

Figure~\ref{fig:qualex_main} additionally shows qualitative examples. From visual inspection, we observe that \proj{} produces images containing objects in complex scenes that better mimic real-world scenarios (Figure~\ref{fig:qualex_ours}). In contrast, Feedback-Guided generations based on Txt2Img models depict the target objects more prominently within simpler scenes and exhibit visual appearances that differ substantially from the real dataset (Figure~\ref{fig:qualex_txt2img}).

We note that, while images synthesized by \proj{} generally remain closer to real samples, they also expand beyond the original data distribution, with some samples lying farther from real exclusive images. One contributing factor may be limitations of our fine-tuning process in accurately learning new word tokens that capture fine-grained visual details, especially when target objects are very small (we discuss failure cases in the Supplementary Material). Another reason is that our method can generate novel combinations of objects and context classes that do not exist in the real dataset, enabling expansion beyond the original samples. For example, a cat or dog sitting near a handbag in Figure~\ref{fig:pipe_main} represents semantic combinations that may rarely appear in real samples.

\subsection{Ablation and Analysis}

\paragraph{Effect of selecting useful generations}
To study the effect of the verification step for selecting informative generations, we compare using only successful generations that manipulate contextual patterns against randomly selecting from all generated samples. We conduct experiments on the COCO dataset and fix the number of additional samples for each class to match the number of co-occurring images. Table~\ref{tab:ablate_select} shows that filtering out synthetic samples containing common contextual patterns provides more useful information for model performance, resulting in a $1.47\%$ improvement in accuracy on exclusive images.

\begin{table}[!h]
    \centering
    \resizebox{\linewidth}{!}{
    \begin{tabular}{lcccc}
    \toprule
        & Exclusive mAP & Cooccur mAP & Unbiased mAP & All mAP \\
        \midrule
        w/o selection & 25.73 $\pm$ 0.04 & 65.18 $\pm$ 0.15 & 72.72 $\pm$ 0.06 & 56.16 $\pm$ 0.05 \\
        w/ selection & \textbf{27.20} $\pm$ 0.32 & \textbf{65.56} $\pm$ 0.04 & \textbf{73.09} $\pm$ 0.03 & \textbf{56.37} $\pm$ 0.04 \\
    \bottomrule
    \end{tabular}}
    \caption{\textbf{Effect of selecting useful generations.}
    Excluding unsuccessful manipulations of contextual patterns improves training.}
    \label{tab:ablate_select}
    \vspace{-3mm}
\end{table}

\vspace{-0.5cm}
\paragraph{Decomposing model accuracy and bias score}
In addition to mAP as an evaluation metric, \citet{singh2020don} proposed a bias score computed for each class, defined as the relative ratio between classifier-predicted probabilities for images containing an object with and without its co-occurring context. Intuitively, a higher bias score indicates a larger performance drop when the object appears without its common context. We analyze the bias scores for the top biased pairs identified by \citet{singh2020don} and find that our method improves the bias score for only 3 out of 14 classes, while the remaining classes show little change. Interestingly, the feature split method improves the bias scores for 8 out of 14 classes.
We note, however, that feature split method often reduces overall accuracy, with a $0.29\%$ drop in exclusive mAP and a $0.71\%$ drop in cooccur mAP compared to the standard model, which can lead to improved bias scores. In contrast, our method improves both cooccur mAP by $2.24\%$ and exclusive mAP by $3.68\%$ relative to the standard model, resulting in unchanged bias scores in some cases (more details in the Supplementary Material).

\begin{figure}[!h]
    \centering
    \includegraphics[width=0.9\linewidth]{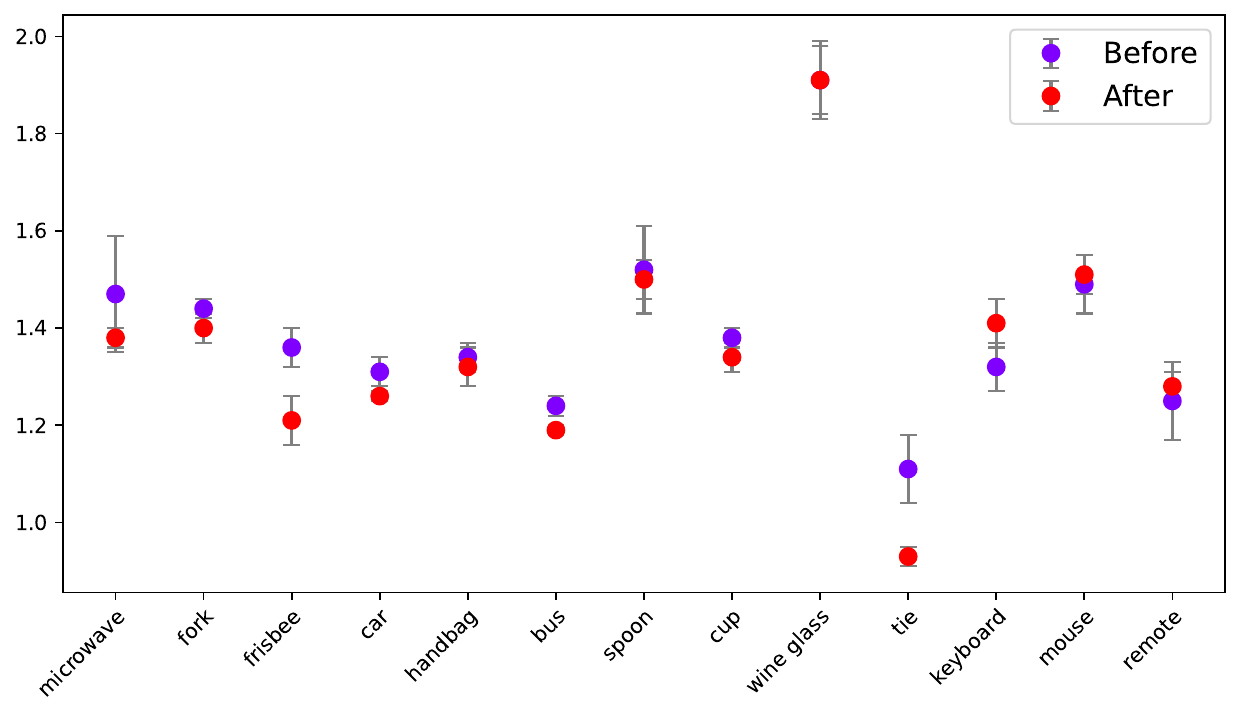}
    \caption{\textbf{Bias scores for biased object classes and their second-highest co-occurring context classes.}
    Even without explicitly targeting these contextual patterns, \proj{} maintains or slightly improves the corresponding bias scores. }
    \label{fig:shift_biasscore}
    \vspace{-5mm}
\end{figure}

In addition to the bias scores of object-context pairs targeted by our method, we also track the bias scores of the second most co-occurring context class. Specifically, for each object class, we consider the context class with the second-highest bias score, excluding the highest-scoring context already addressed by augmentation. Figure~\ref{fig:shift_biasscore} compares these bias scores before and after our pipeline, i.e., between the standard classifier and DecoupleGen.

We found that these bias scores either improve or remain similar after applying \proj{} (lower bias score values indicate less bias), with improvements observed in 9 out of 13 classes. This demonstrates that \proj{} does not negatively affect, and can even improve, bias scores without explicitly targeting these co-occurring biases.

\begin{figure}[!h]
    \centering
    \includegraphics[width=\linewidth]{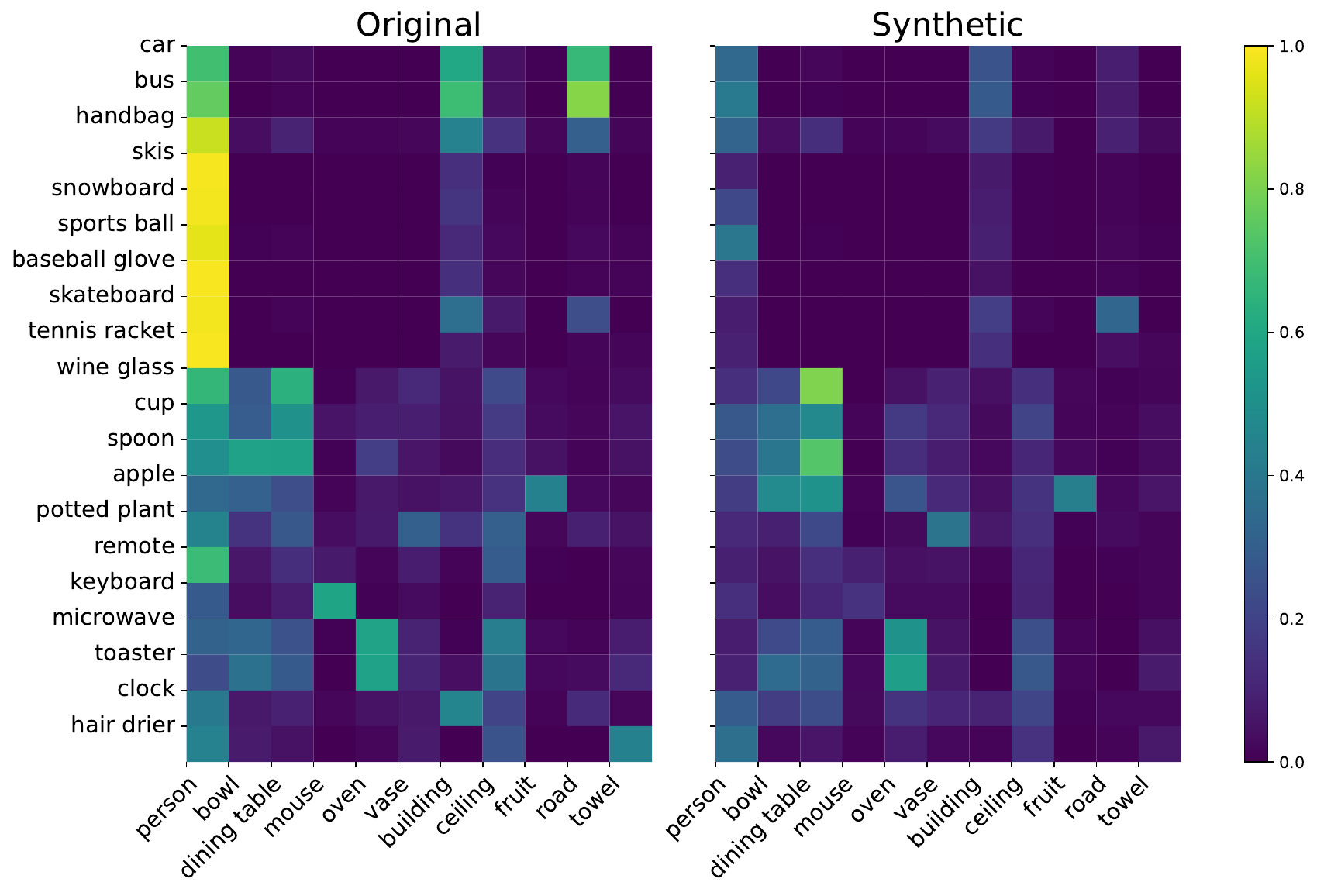}
    \caption{ \textbf{Conditional distribution of biased object classes (y-axis) and cooccurring contexts (x-axis).} Common contextual patterns in original datasets are reduced, while generations also introduce new cooccurring patterns.}
    \label{fig:shift_classlabel}
    \vspace{-5mm}
\end{figure}

\vspace{-0.3cm}
\paragraph{Newly introduced common contextual patterns.}
Since our approach targets only the context label with the highest bias score for each object class, we further investigate whether including generated samples negatively affects other contextual patterns or introduces additional contextual bias. Figure~\ref{fig:shift_classlabel} shows the conditional probability distribution of observing co-occurring contexts (x-axis) for each biased object category (y-axis) in both the original dataset and the synthesized samples.

We observe that many original co-occurrence patterns are reduced to substantially lower probabilities, for example the co-occurrence of \texttt{person} with \texttt{handbag}, \texttt{snowboard}, and \texttt{baseball glove}. However, newly introduced co-occurrence patterns also emerge, such as \texttt{dining table} with \texttt{wine glass} and \texttt{apple} with \texttt{dining table}, possibly due to implicit biases in pre-trained generative models. While we do not address these secondary biases in this work, they could be mitigated in future work through approaches such as tracking the class distribution of the augmented dataset during the image generation process or employing iterative generation strategies to synthesize additional images. Despite the presence of these newly introduced biases, \proj{} improves classifier performance compared to existing approaches.

%% file: 8page_sections/discussion.tex
\vspace{-1mm}
\section{Discussions}
In this work, we introduce \textbf{\proj}, a contextual debiasing pipeline that adapts diffusion model personalization to learn new word tokens for visual details and generate images without common contextual patterns. Our experiments show that incorporating generated samples from our method achieves competitive or superior performance on downstream image classification and object recognition tasks. Through qualitative analysis, we further demonstrate two key advantages of \proj{} over alternative generation approaches: it produces visually consistent and semantically meaningful images, and its generations better align with the downstream dataset distribution.

\textbf{Limitations and future directions.}
\proj{} fine-tunes the diffusion model on each image to capture visual details, which can be time-consuming for large-scale datasets. A promising future direction is to improve the scalability of such approaches without degrading generation quality or downstream model performance. Finally, although our experimental results show that \proj{} significantly improves over previous work, there remains a substantial gap in classification accuracy between uncommon and common samples, motivating further investigation into how to better reduce this discrepancy.


%% file: 8page_sections/appendix.tex
\clearpage
\setcounter{page}{1}
\maketitlesupplementary

\section{Experimental Details}

\subsection{Descriptions of Previous Approaches}

We provide more detailed descriptions of existing approaches that we compared against for each downstream task below:

\vspace{-5mm}
\paragraph{Image Classification}
We compare \proj{} to a set of approaches mentioned in \citep{yang2023change, hemmat2023feedback}.
\textbf{GroupDRO} \citep{sagawa2019distributionally} minimizes the worst-case group loss by reweighting training examples based on their group representations. \textbf{Mixup} \citep{zhang2017mixup} performs data augmentation by linearly interpolating pairs of training samples and their labels, resulting in smoothier decision boundaries. \textbf{LISA} \citep{yao2022improving} learns invariant predictors through strategically augmenting training data with interpolations between samples. \textbf{Balanced Softmax} \citep{ren2020balanced} adjusts standard cross entropy loss by reweighting logits inversely proportional to class frequencies in training dataset. \textbf{Classifier Retraining} \citep{kang2019decoupling} decouples representation and classifier learning, which finetunes classification layer using class balanced sampling with a fixed visual feature.

\vspace{-3mm}
\paragraph{Object Recognition}
We compare \proj{} to set of approaches proposed and mentioned in \citep{singh2020don}.
\textbf{Class balancing loss} \citep{cui2019class} treats scenarios where biased categories occur exclusively as long-tail classes. \textbf{Negative penalty} assigns high penalty when model predicts the context category in cases where a biased category occurs exclusively. \textbf{Weighted loss} applies 10 times higher weights to loss values when biased categories occur exclusively. \textbf{Remove co-occurring labels} modifies true labels of co-occurring images by removing context categories $c$ for each $b$ in training datasets. \textbf{Remove co-occurring images} modifies training datasets by removing all instances where biased categories co-occur with its context classes. \textbf{CAM} modifies training objective by using class activation map \citep{zhou2016cam} as location supervision. \textbf{Feature split} separates feature space of classifier by enforcing them to attend to objects and their context.

\subsection{Implementation Details}

\paragraph{Image Generation}
We use publicly available Stable Diffusion v2 in Huggingface \citep{wolf2019huggingface}, with default hyperparameters as the base generative model. In our image generation step, we finetune text encoder and diffusion model unet with learning rate of word tokens $10^{-4}$ and learning rate of diffusion models $10^{-6}$. We follow implementation details and experimental configurations in~\citet{avrahami2023break}.

\vspace{-5mm}
\paragraph{NICO dataset} 
We finetune Stable Diffusion v2 for $200$ iterations in both finetuning stages. We search for these hyperparameters across $\{ 10^{-4}, 5 \times 10^{-4} \}$ and $\{ 10^{-6}, 5 \times 10^{-6}, 10^{-5} \}$ for both learning rate and $\{ 200, 300, 400 \}$ for training iterations. We adopt LoRA \citep{hu2021lora} as parameter-efficient fine-tuning technique.

For image classification model, following previous work \citep{yang2023change, hemmat2023feedback}, we search over 10 hyperparameter configurations and select the best based on validation performance. We train classifiers for 50 epochs and report test performance.

\vspace{-5mm}
\paragraph{COCO dataset}
We finetune Stable Diffusion v2 for $300$ iterations in both finetuning stages. We search for best hyperparameters among the same set as in NICO dataset.

For multi-lable classification model, following previous work \citep{singh2020don}, we train classifiers with batch size of 200, learning rate 0.1, and cosine scheduler for a total of 100 epochs. We search across different values of batch size $\{ 128, 200, 256 \} $, learning rate schedules at [60] or [50, 75] epochs, and pick final values based on validation performance. In both cases, we utilizes standard cross entropy loss without modifying supervised learning objectives.

\vspace{-5mm}
\paragraph{PASCAL dataset}
The PASCAL dataset~\citep{mottaghi2014pascal} contains 59 classes, including 20 object classes. Following prior work \citep{singh2020don}, we identify 10 biased classes and their most cooccurring context categories based on bias scores (Table~\ref{tab:pascal_setup}). Since the dataset does not provide a semantic hierarchy of labels, following COCO we manually construct sets of semantically similar classes for replacement (Table~\ref{tab:pascal_classlist}). We use same set of hyperparameters for model training in both image generation stage and classification stage.

\begin{table}[!h]
    \centering
    \resizebox{0.6\linewidth}{!}{
    \begin{tabular}{ccc}
    \toprule
        Biased Class & Cooccur Class & Bias Score \\
        \midrule
        car & road & 1.18 $\pm$ 0.01 \\
        aeroplane & sky & 1.30 $\pm$ 0.06 \\
        tvmonitor & wall & 1.28 $\pm$ 0.08 \\
        cow & grass & 1.35 $\pm$ 0.05 \\
        table & wall & 1.40 $\pm$ 0.04 \\
        bus & building & 1.64 $\pm$ 0.06 \\
        chair & floor & 1.77 $\pm$ 0.06 \\
        train & ground & 1.85 $\pm$ 0.07 \\
        sofa & wall & 2.02 $\pm$ 0.08 \\
        boat & water & 35.39 $\pm$ 4.41 \\
        \bottomrule
    \end{tabular}}
    \caption{Top 10 biased and cooccurring pairs among 20 object classes identified on PASCAL dataset \citep{mottaghi2014pascal}.}
    \label{tab:pascal_setup}
    \vspace{-5mm}
\end{table}

\begin{table}[!h]
    \centering
    \resizebox{0.8\linewidth}{!}{
    \begin{tabular}{l|l}
    \toprule
        road & snow, grass, water, ground, track \\
        sky & building, ceiling, wall (indoor) \\
        wall & ceiling, curtain, door, fence, window \\
        grass & sidewalk, track, ground, road, snow \\
        building & bench, chair, fence, sign \\
        floor & ground, platform, road, sidewalk, track \\
        ground & grass, platform, road, sidewalk, track, water \\
        water & grass, platform, road, rock, wood, track \\
        \bottomrule
    \end{tabular}}
    \caption{List of original cooccurring classes (left) and their semantically similar classes used for replacement (right).}
    \label{tab:pascal_classlist}
    \vspace{-5mm}
\end{table}

\newpage
\section{Additional Experimental Results} \label{app:exp_result}

\subsection{Image classification}

\begin{figure}[!h]
    \centering
    \includegraphics[width=0.9\linewidth]{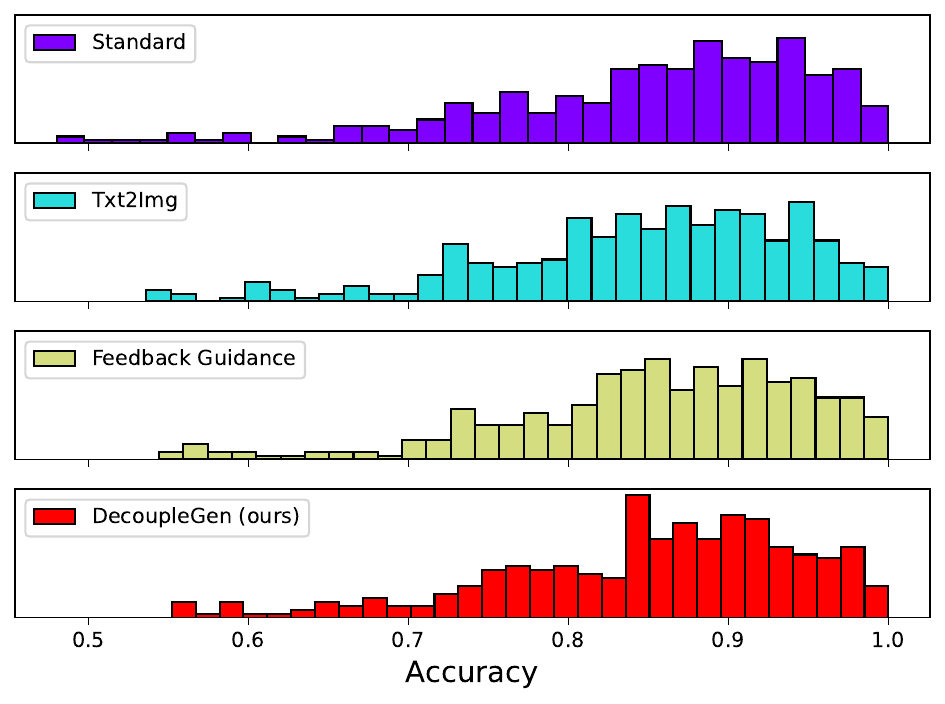}
    \caption{\textbf{Histogram of group-wise accuracies on NICO dataset.} 
    All generation approaches achieve similar performance, all of which improve from standard classifier.}
    \label{fig:group_accs_nico}
    \vspace{-5mm}
\end{figure}

\paragraph{Decompose group-wise model accuracy on NICO dataset.}
In NICO dataset, in addition to overall accuracy and worst group accuracy, we take a closer look at classifier accuracies on object-context combinations, after removing groups that contain few data points in validation split (less than 25 samples in our experiments), since model evaluations may not generalize well for these groups. As shown in Figure \ref{fig:group_accs_nico}, we observe similar performance between different generation approaches, all of which improves from standard classifier. Overall DecoupleGen achieves similar performance compared to other approaches.

\subsection{Object Recognition}

\paragraph{Decompose class-wise model accuracy and bias score on COCO dataset.}
In COCO dataset, we further look at performance of accuracy metric and bias score metric on each class with initial bias score at least 1.5 on average from standard classifier, after removing classes that contain few data points in validation split (less than 30 samples in our experiments). These classes mostly overlap with top biased classes identified in \citep{singh2020don}. We report results in Table \ref{tab:class_accs_coco} and Table \ref{tab:bias_score_coco} respectively.

As shown in Table \ref{tab:class_accs_coco}, \proj{} shows significant improvements in 13 out of 14 classes on both exclusive and cooccur mAP, while feature split only improves in 4 classes on exclusive mAP and results in performance drop in 4 classes on both exclusive and cooccur mAP. However, as shown in Table \ref{tab:bias_score_coco}, feature split shows significant improvements in 8 out of 14 classes in bias scores, while our method only improves 3 classes. This demonstrates that advantages of \proj{} mainly comes in better recognition performance in terms of model accuracy instead of bias score.

In particular, bias score metric that measures relative accuracy drop between cooccur and exclusive images may not capture complete information about classifier performance. While feature split results in better bias score, it penalizes model accuracy on cooccur, i.e. overrepresented, images more than exclusive images, resulting in no improvement in its accuracy from standard classifier. Instead, \proj{} improves $3.68\%$ in exclusive mAP and $2.24\%$ in cooccur mAP compared to standard classifier, with more significant benefits on exclusive, i.e. underrepresented, images.

Additionally, our results in Table \ref{tab:class_accs_coco} show that \proj{} achieves much better performance in exclusive and cooccur mAP compared to Feedback Guidance approach \citep{hemmat2023feedback} based on Txt2Img models, which only improves 2 out of 14 classes in exclusive mAP and 4 classes in cooccur mAP. As discussed in the main paper, such discrepancy is possibly due to that Txt2Img models fail to generate samples that maintain similar to original datasets compared to our approach, therefore providing limited information helpful to downstream tasks. 

\paragraph{Quantitative Comparisons on PASCAL dataset~\citep{mottaghi2014pascal}.}
We report experimental results in Table~\ref{tab:baseline_pascal}. Using the same number of synthetic samples (2k), \proj{} improves over other generation-based baselines by 3.01\%, increasing exclusive mAP from 58.48\% to 61.49\%. By further including more generations (8k), \proj{} improves over the best reweighting-based baseline by 1.26\%, increasing exclusive mAP from 63.00\% to 64.26\%. DecoupleGen outperforms existing approches while maintaining comparable cooccur mAP. Together with results on COCO dataset, this shows that \proj{} is consistently beneficial on common real world scene datasets.

\begin{table}[!ht]
    \centering
    \resizebox{0.95\linewidth}{!}{
    \begin{tabular}{lccc}
    \toprule
        Method & Num Gens & Exclusive mAP & Cooccur mAP \\
        \midrule
        Standard & 0 & 57.56 $\pm$ 0.68 & 86.65 $\pm$ 0.08 \\
        Real Cooccur & 0 & 57.30 $\pm$ 0.58 & 85.66 $\pm$ 0.48 \\
        Class balancing loss & 0 & 60.08 $\pm$ 0.44 & 86.74 $\pm$ 0.11 \\
        Negative penalty & 0 & 57.56 $\pm$ 0.55 & 86.77 $\pm$ 0.14 \\
        Weighted loss & 0 & \textbf{63.00} $\pm$ 0.39 & 86.23 $\pm$ 0.11 \\
        Remove cooccur labels & 0 & 57.68 $\pm$ 0.32 & \textbf{86.84} $\pm$ 0.14 \\
        Remove cooccur images & 0 & 61.47 $\pm$ 0.41 & 86.41 $\pm$ 0.16 \\
        CAM & 0 & 60.55 $\pm$ 0.27 & 86.64 $\pm$ 0.15 \\ 
        Feature split & 0 & 60.08 $\pm$ 0.27 & 86.68 $\pm$ 0.16 \\
        \midrule
        Txt2Img Cooccur & $2k$ & 58.14 $\pm$ 2.05 & 85.11 $\pm$ 0.34 \\
        Txt2Img Exclusive & $2k$ & 57.78 $\pm$ 0.71 & 85.32 $\pm$ 0.84 \\
        Txt2Img Target Class & $2k$ & 57.33 $\pm$ 0.96 & 84.90 $\pm$ 1.44 \\
        SD Inpainting & $2k$ & 57.69 $\pm$ 1.87 & 85.36 $\pm$ 0.43 \\
        Blended LD & $2k$ & 57.51 $\pm$ 1.54 & 84.78 $\pm$ 0.38 \\
        Feedback Guidance & $2k$ & 58.48 $\pm$ 1.33 & 85.59 $\pm$ 0.44 \\
        DecoupleGen (ours) & $2k$ & \textbf{61.49} $\pm$ 0.90 & \textbf{85.60} $\pm$ 0.47\\
        \midrule
        DecoupleGen (ours) & $8k$ & \textbf{64.26} $\pm$ 0.75 & 85.08 $\pm$ 0.32 \\
    \bottomrule
    \end{tabular}}
    \caption{\textbf{Quantitative comparisons on PASCAL dataset.}
    \proj{} achieves better performance in exclusive mAP and maintain similar performance in cooccur mAP compared to existing approaches.}
\label{tab:baseline_pascal}
\vspace{-5mm}
\end{table}

\begin{table*}[!h]
    \centering
    \resizebox{\textwidth}{!}{
    \begin{tabular}{ll|cccccc}
        \toprule
        Bias class & Cooccur class & \multicolumn{2}{c}{Feature split \citep{singh2020don}} & \multicolumn{2}{c}{Feedback guidance \citep{hemmat2023feedback}} & \multicolumn{2}{c}{DecoupleGen (ours)} \\
        & & Exclusive mAP & Cooccur mAP & Exclusive mAP & Cooccur mAP & Exclusive mAP & Cooccur mAP \\
        \midrule
        microwave & cabinet & 34.53 $\pm$ 1.50 \textcolor{blue}{(+3.58)} & 58.30 $\pm$ 0.73 \textcolor{gray}{(-0.95)} & 30.02 $\pm$ 0.38 \textcolor{gray}{(-0.93)} & 62.75 $\pm$ 1.43 \textcolor{blue}{(+3.51)} & \textbf{37.66} $\pm$ 1.32 \textcolor{blue}{(+6.71)} & \textbf{68.35} $\pm$ 0.16 \textcolor{blue}{(+9.11)} \\
        
        potted plant & wall & 21.51 $\pm$ 0.97 \textcolor{blue}{(+2.00)} & 46.95 $\pm$ 0.76 \textcolor{red}{(-4.77)} & 20.18 $\pm$ 0.17 \textcolor{blue}{(+0.67)} & 54.13 $\pm$ 0.58 \textcolor{blue}{(+2.41)} & \textbf{24.01} $\pm$ 0.24 \textcolor{blue}{(+4.50)} & \textbf{57.53} $\pm$ 0.35 \textcolor{blue}{(+5.81)} \\
        
        fork & dining table & 12.00 $\pm$ 0.46 \textcolor{gray}{(-0.61)} & 46.07 $\pm$ 0.24 \textcolor{gray}{(+0.67)} & 11.91 $\pm$ 0.39 \textcolor{gray}{(-0.70)} & 45.19 $\pm$ 0.58 \textcolor{gray}{(-0.21)} & \textbf{16.05} $\pm$ 0.47 \textcolor{blue}{(+3.44)} & \textbf{49.29} $\pm$ 0.52 \textcolor{blue}{(+3.89)} \\
        
        frisbee & grass & 50.28 $\pm$ 1.16 \textcolor{red}{(-3.22)} & 84.64 $\pm$ 0.20 \textcolor{gray}{(-0.45)} & 52.03 $\pm$ 0.20 \textcolor{gray}{(-1.47)} & 85.61 $\pm$ 0.38 \textcolor{gray}{(+0.52)} & \textbf{60.06} $\pm$ 0.36 \textcolor{blue}{(+6.56)} & \textbf{86.51} $\pm$0.23 \textcolor{blue}{(+1.42)} \\
        
        car & road & 38.17 $\pm$ 0.30 \textcolor{blue}{(+1.83)} & 78.49 $\pm$ 0.10 \textcolor{red}{(-0.67)} & 36.06 $\pm$ 0.26 \textcolor{gray}{(-0.28)} & 79.35 $\pm$ 0.18 \textcolor{gray}{(+0.20)} & \textbf{39.73} $\pm$0.36 \textcolor{blue}{(+3.40)} & \textbf{81.21} $\pm$ 0.25 \textcolor{blue}{(+2.06)} \\
        
        handbag & person & 2.78 $\pm$ 0.67 \textcolor{gray}{(-0.03)} & 41.06 $\pm$ 0.23 \textcolor{gray}{(-0.18)} & 2.56 $\pm$ 0.18 \textcolor{red}{(-0.25)} & 41.22 $\pm$ 0.18 \textcolor{gray}{(-0.02)} & \textbf{3.87} $\pm$ 0.41 \textcolor{blue}{(+1.06)} & \textbf{43.15} $\pm$ 0.45 \textcolor{blue}{(+1.91)} \\
        
        bus & road & 44.72 $\pm$ 0.63 \textcolor{blue}{(+3.56)} & 83.95 $\pm$ 0.21 \textcolor{red}{(-1.61)} & 40.93 $\pm$ 0.49 \textcolor{gray}{(-0.23)} & 85.18 $\pm$ 0.18 \textcolor{gray}{(-0.37)} & 44.88 $\pm$ 1.07 \textcolor{blue}{(+3.72)} & \textbf{86.71} $\pm$ 0.16 \textcolor{blue}{(+1.15)} \\
        
        spoon & bowl & 12.01 $\pm$ 0.50 \textcolor{red}{(-1.99)} & 34.11 $\pm$ 0.46 \textcolor{red}{(-1.34)} & 13.52 $\pm$ 0.36 \textcolor{gray}{(-0.48)} & 36.22 $\pm$ 1.34 \textcolor{gray}{(+0.77)} & \textbf{16.03} $\pm$ 0.22 \textcolor{blue}{(+2.03)} & \textbf{37.05} $\pm$ 1.20 \textcolor{blue}{(+1.60)} \\
        
        cup	& dining table & 25.41 $\pm$ 0.29 \textcolor{red}{(-4.28)} & 63.06 $\pm$ 0.20 \textcolor{gray}{(+1.00)} & 29.71 $\pm$ 0.23 \textcolor{gray}{(+0.02)} & 62.90 $\pm$ 0.47 \textcolor{gray}{(+0.84)} & \textbf{32.72} $\pm$ 0.08 \textcolor{blue}{(+3.03)} & \textbf{65.22} $\pm$ 0.28 \textcolor{blue}{(+3.15)} \\
        
        wine glass & person & 36.86 $\pm$ 0.14 \textcolor{gray}{(0.43)} & 55.94 $\pm$ 0.78 \textcolor{gray}{(-1.15)} & 36.34 $\pm$ 0.34 \textcolor{gray}{(-0.08)} & 56.63 $\pm$ 0.31 \textcolor{gray}{(-0.45)} & \textbf{42.06} $\pm$ 0.78 \textcolor{blue}{(+5.64)} & \textbf{59.18} $\pm$ 0.39 \textcolor{blue}{(+2.10)} \\
        
        tie & clothes & 10.06 $\pm$ 0.41 \textcolor{gray}{(-0.09)} & 36.60 $\pm$ 0.38 \textcolor{gray}{(+0.20)} & 9.86 $\pm$ 0.86 \textcolor{gray}{(-0.29)} & 38.50 $\pm$ 0.41 \textcolor{blue}{(+2.10)} & \textbf{11.29} $\pm$ 0.46 \textcolor{blue}{(+1.14)} & 39.35 $\pm$ 1.29 \textcolor{blue}{(+2.95)} \\
        
        keyboard & mouse & 46.30 $\pm$ 0.31 \textcolor{gray}{(-0.43)} & 83.73 $\pm$ 0.54 \textcolor{gray}{(-0.67)} & 47.59 $\pm$ 0.43 \textcolor{blue}{(+0.85)} & 85.32 $\pm$ 0.19 \textcolor{blue}{(+0.92)} & \textbf{49.90} $\pm$ 0.17 \textcolor{blue}{(+3.17)} & \textbf{85.72} $\pm$ 0.15 \textcolor{blue}{(+1.32)} \\
        
        mouse & keyboard & 24.88 $\pm$ 0.58 \textcolor{gray}{(0.54)} & 79.20 $\pm$ 0.18 \textcolor{gray}{(-0.25)} & 23.51 $\pm$ 0.26 \textcolor{gray}{(-0.83)} & 80.45 $\pm$ 0.20 \textcolor{gray}{(+0.99)} & \textbf{31.10} $\pm$ 0.57 \textcolor{blue}{(+6.76)} & \textbf{81.04} $\pm$ 0.50 \textcolor{blue}{(+1.59)} \\
        
        remote & person & 19.36 $\pm$ 0.47 \textcolor{red}{(-3.98)} & 70.44 $\pm$ 0.52 \textcolor{gray}{(-0.03)} & 22.39 $\pm$ 0.34 \textcolor{red}{(-0.95)} & 71.18 $\pm$ 0.68 \textcolor{gray}{(+0.71)} & \textbf{24.32} $\pm$ 0.76 \textcolor{gray}{(+0.98)} & 70.63 $\pm$ 0.73 \textcolor{gray}{(+0.17)} \\
        \midrule
        \multicolumn{2}{l|}{Mean} & 26.49 $\pm$ 0.29 \textcolor{gray}{(-0.29)} & 61.86 $\pm$ 0.12 \textcolor{gray}{(-0.71)} & 26.66 $\pm$ 0.18 \textcolor{gray}{(-0.13)} & 63.22 $\pm$ 0.30 \textcolor{gray}{(+0.65)} & \textbf{30.46} $\pm$ 0.11 \textcolor{blue}{(+3.68)} & \textbf{64.81} $\pm$ 0.26 \textcolor{blue}{(+2.24)} \\
        \bottomrule
     \end{tabular}}
    \caption{\textbf{Comparisons of model accuracy on each class.} 
    We report both exclusive and cooccur mAP for each class, and change in mAP values compared to standard classifier trained on original dataset, where \textcolor{blue}{blue} indicates improvement, \textcolor{red}{red} indicates drop, and \textcolor{gray}{gray} indicates similar performance within error bars. 
    Bolded values indicate significantly better performance for each class among different approaches.
    DecoupleGen achieves better performance and more significant improvements in model accuracy in most cases.}
    \label{tab:class_accs_coco}
    \vspace{-3mm}
\end{table*}

\begin{table}[!h]
    \centering
    \resizebox{\linewidth}{!}{
    \begin{tabular}{ll|cc}
        \toprule
        Bias class & Cooccur class & Feature split \citep{singh2020don} & DecoupleGen (ours) \\
        \midrule
        microwave & cabinet & \textbf{1.36} $\pm$ 0.02 \textcolor{blue}{(-0.13)} & 1.60 $\pm$ 0.02 \textcolor{red}{(+0.11)} \\
        potted plant & wall & \textbf{1.25} $\pm$ 0.03 \textcolor{blue}{(-0.34)} & 1.58 $\pm$ 0.04 \textcolor{gray}{(-0.01)} \\
        fork & dining table & 1.75 $\pm$ 0.08 \textcolor{gray}{(+0.08)} & \textbf{1.60} $\pm$ 0.06 \textcolor{gray}{(-0.06)} \\
        frisbee & grass & 1.73 $\pm$ 0.06 \textcolor{gray}{(+0.03)} & \textbf{1.64} $\pm$ 0.04 \textcolor{gray}{(-0.06)} \\
        car & road & \textbf{1.37} $\pm$ 0.04 \textcolor{blue}{(-0.38)} & 1.61 $\pm$ 0.02 \textcolor{blue}{(-0.14)} \\
        handbag & person & 1.63 $\pm$ 0.13 \textcolor{gray}{(-0.13)} & 1.72 $\pm$ 0.12 \textcolor{gray}{(-0.05)} \\
        bus & road & \textbf{1.55} $\pm$ 0.04 \textcolor{blue}{(-0.21)} & 1.78 $\pm$ 0.03 \textcolor{gray}{(+0.02)} \\
        spoon & bowl & 1.58 $\pm$ 0.08 \textcolor{blue}{(-0.28)} & 1.60 $\pm$ 0.03 \textcolor{blue}{(-0.27)} \\
        cup & dining table & \textbf{1.62} $\pm$ 0.11 \textcolor{blue}{(-0.28)} & 1.87 $\pm$ 0.05 \textcolor{gray}{(-0.02)} \\
        wine glass & person & 1.75 $\pm$ 0.02 \textcolor{blue}{(-0.15)} & \textbf{1.63} $\pm$ 0.01 \textcolor{blue}{(-0.27)} \\
        tie & clothes & 1.91 $\pm$ 0.09 \textcolor{gray}{(+0.00)} & 1.91 $\pm$ 0.05 \textcolor{gray}{(-0.01)} \\
        keyboard & mouse & \textbf{1.72} $\pm$ 0.03 \textcolor{blue}{(-0.24)} & 1.92 $\pm$ 0.05 \textcolor{gray}{(-0.04)} \\
        mouse & keyboard & 2.04 $\pm$ 0.09 \textcolor{gray}{(+0.05)} & \textbf{1.87} $\pm$ 0.07 \textcolor{gray}{(-0.13)} \\
        remote & person & 2.13 $\pm$ 0.13 \textcolor{gray}{(+0.10)} & 1.86 $\pm$ 0.12 \textcolor{gray}{(-0.17)} \\
        \bottomrule
    \end{tabular}}
    \caption{\textbf{Bias score of each class and its cooccurring context.} 
    We report bias scores for each class, and changes in bias score values compared to standard classifier trained on original dataset, where \textcolor{blue}{blue} indicates improvements, \textcolor{red}{red} indicates drop, and \textcolor{gray}{gray} indicates similar performance within error bars.
    Bolded values indicate significantly better performance for each class among different approaches.
    Feature split shows more improvements in bias scores compared to our method.}
    \label{tab:bias_score_coco}
    \vspace{-5mm}
\end{table}


\subsection{Additional Qualitative Results}

\begin{figure}[!h]
    \centering
        \includegraphics[width=0.24\linewidth]{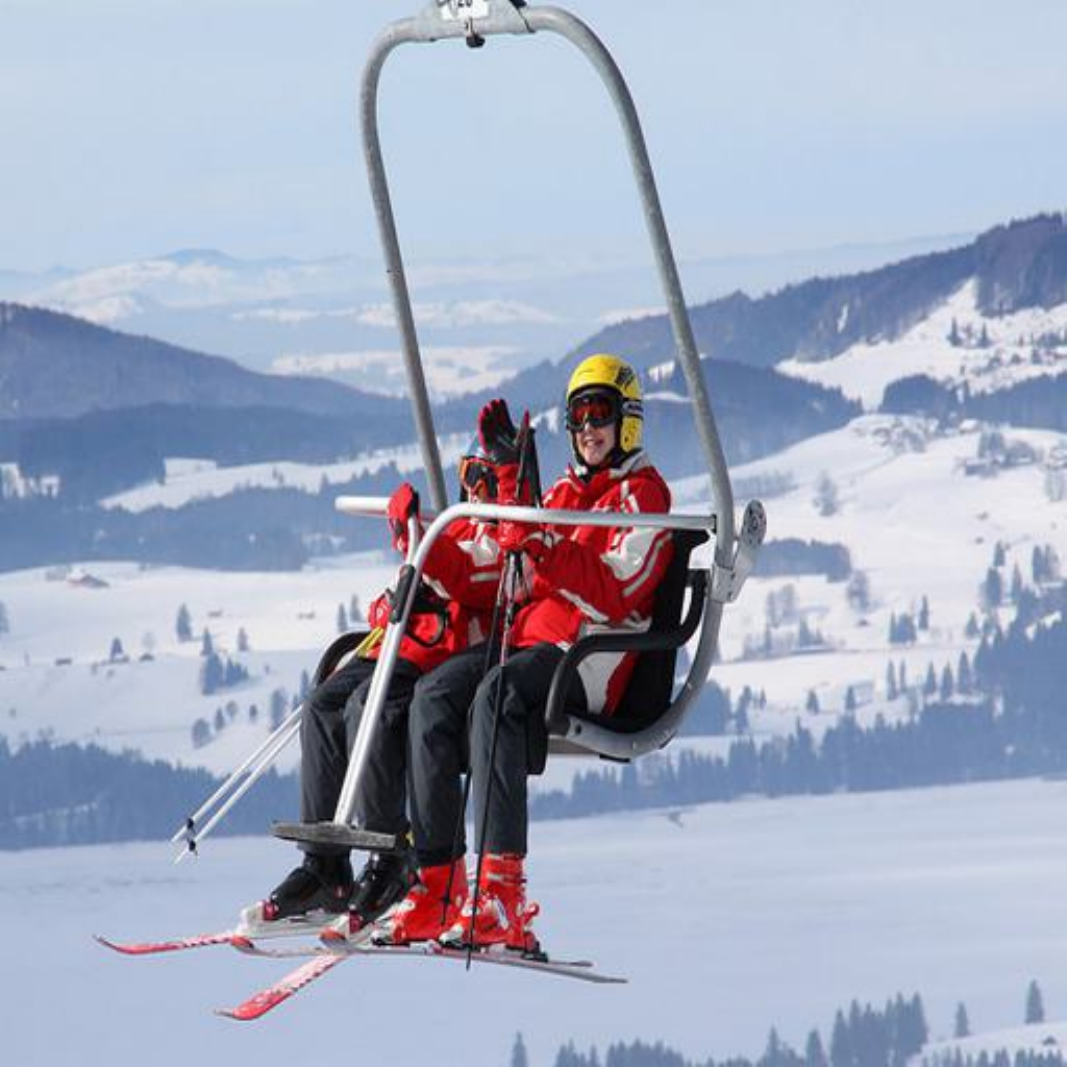}
        \hspace{-0.02\linewidth}
        \includegraphics[width=0.24\linewidth]{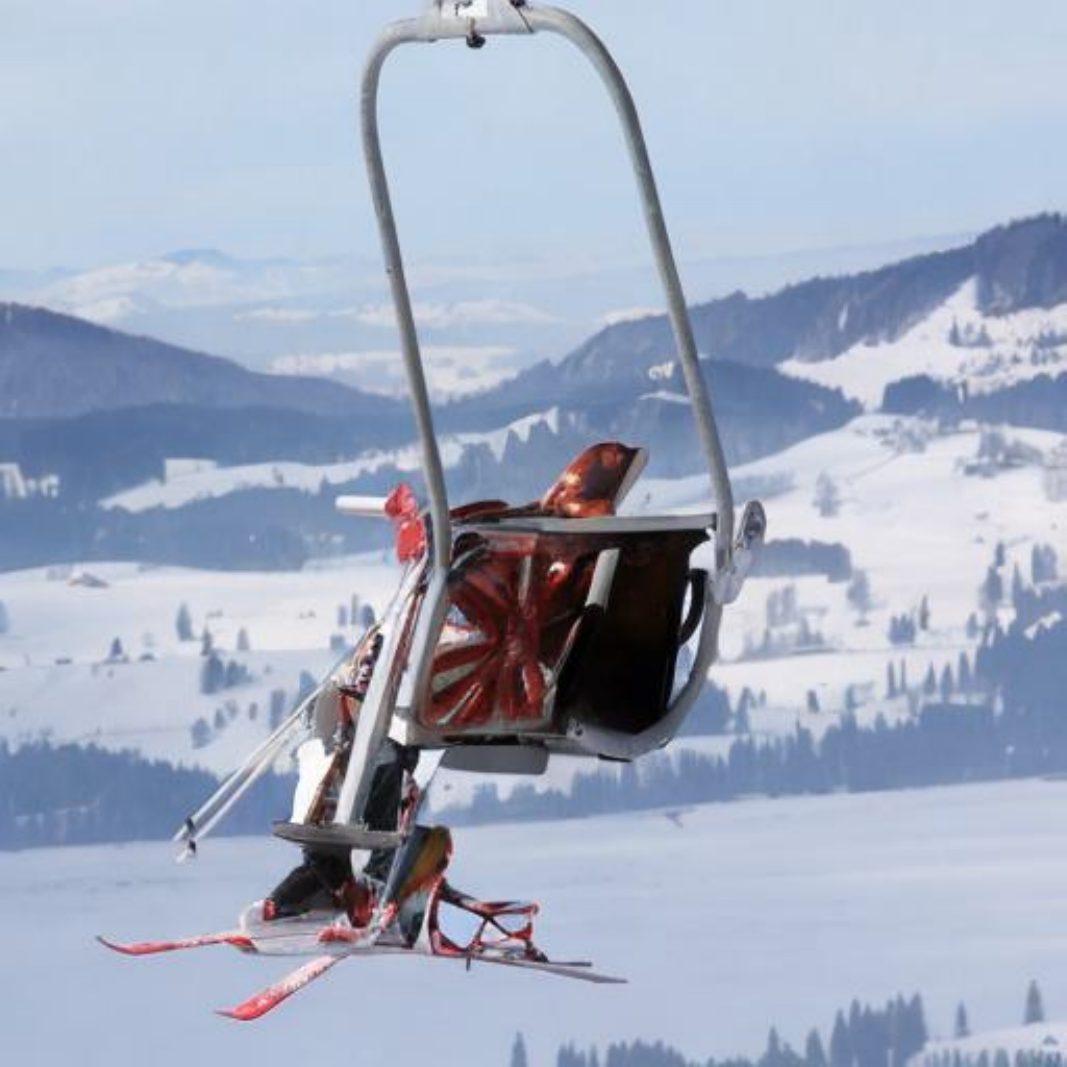}
        \hspace{-0.02\linewidth}
        \includegraphics[width=0.24\linewidth]{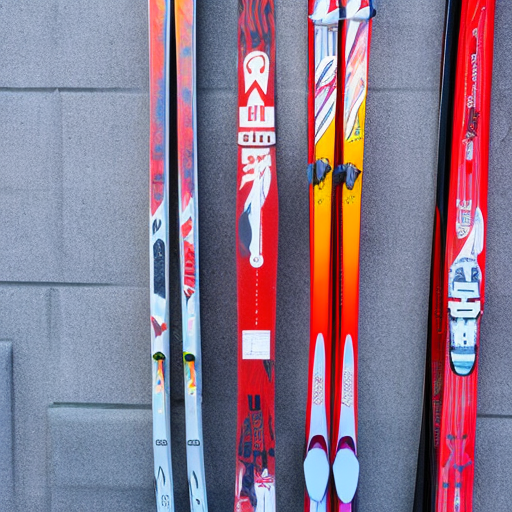}
        \hspace{-0.02\linewidth}
        \includegraphics[width=0.24\linewidth]{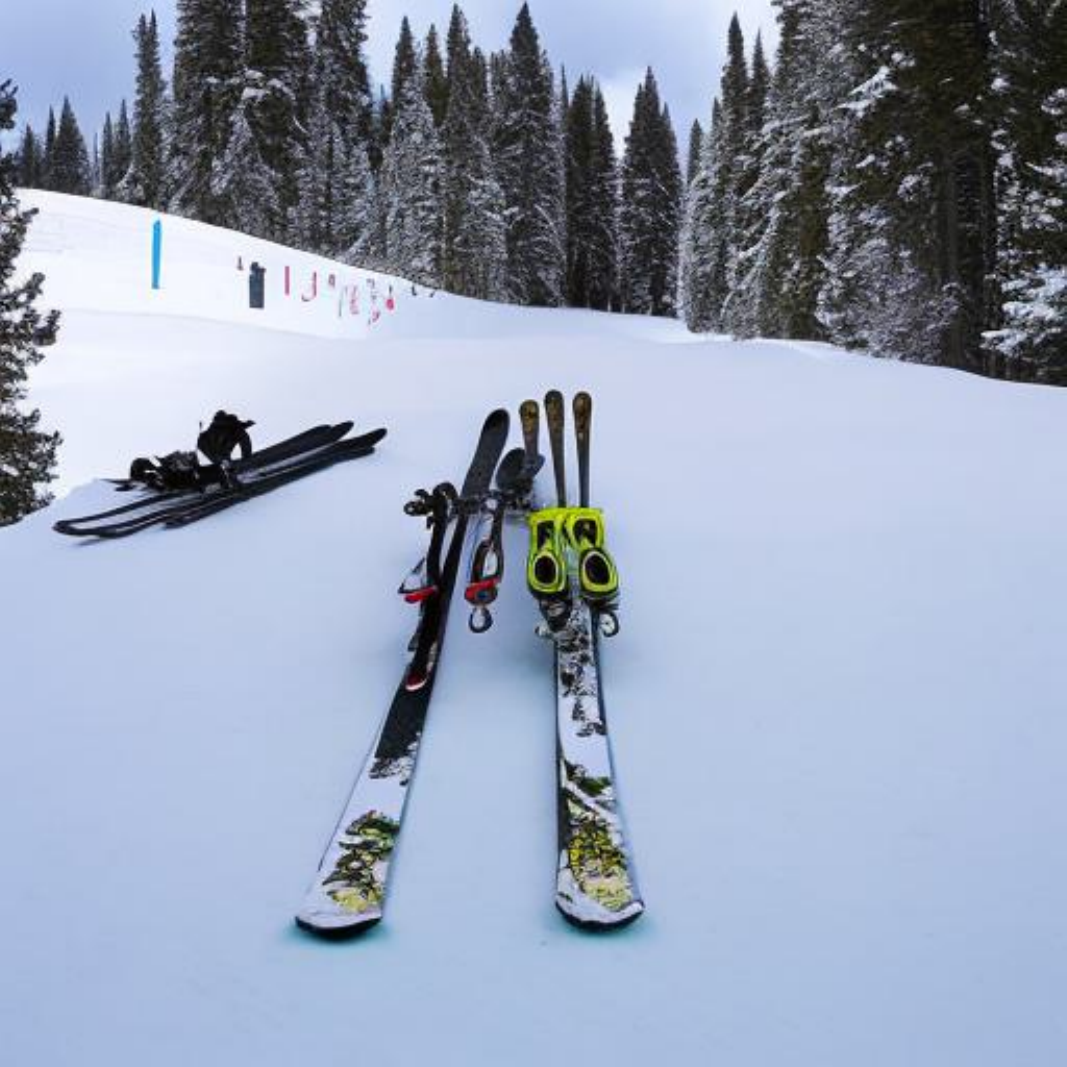}
        \hspace{-0.02\linewidth}
        \includegraphics[width=0.24\linewidth]{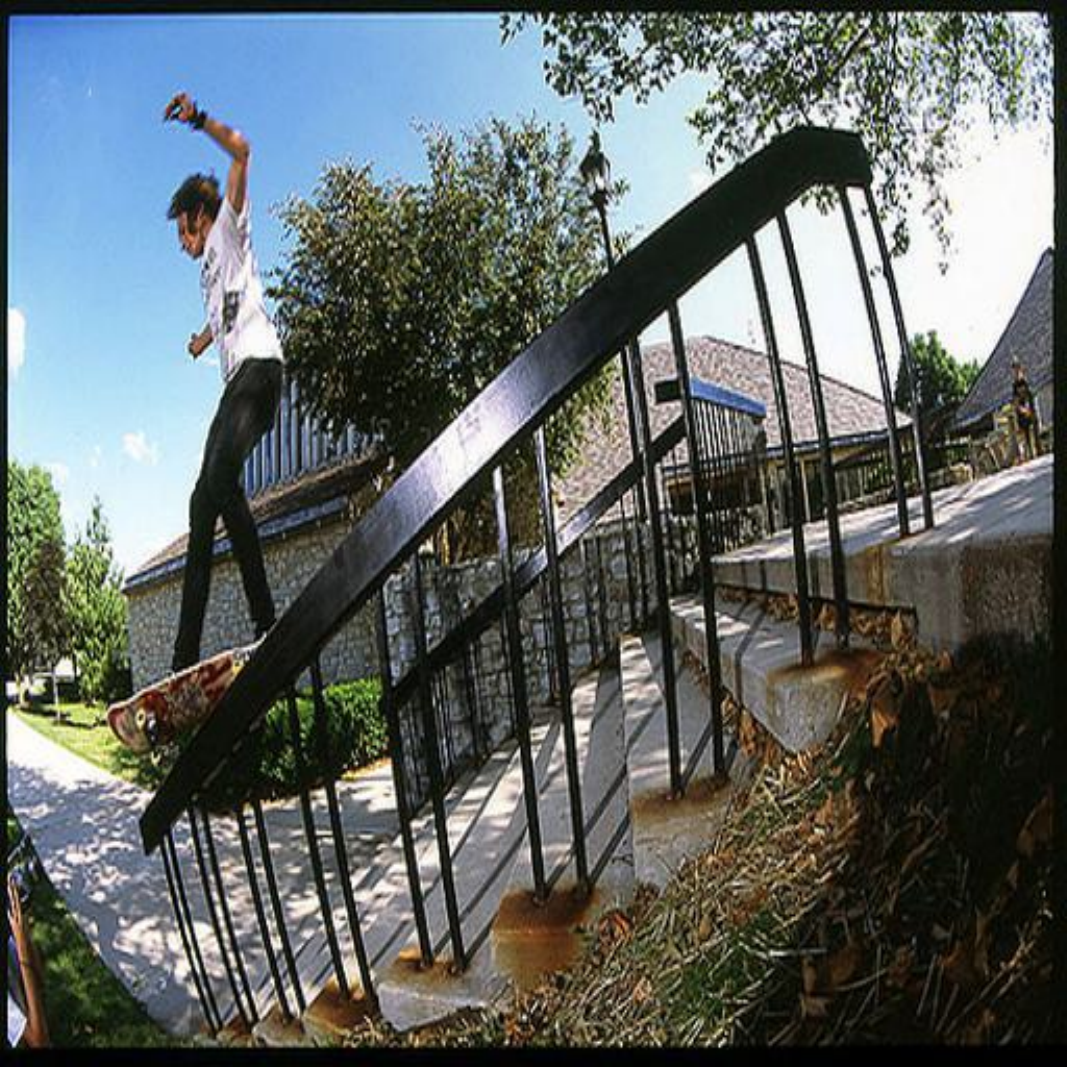}
        \hspace{-0.02\linewidth}
        \includegraphics[width=0.24\linewidth]{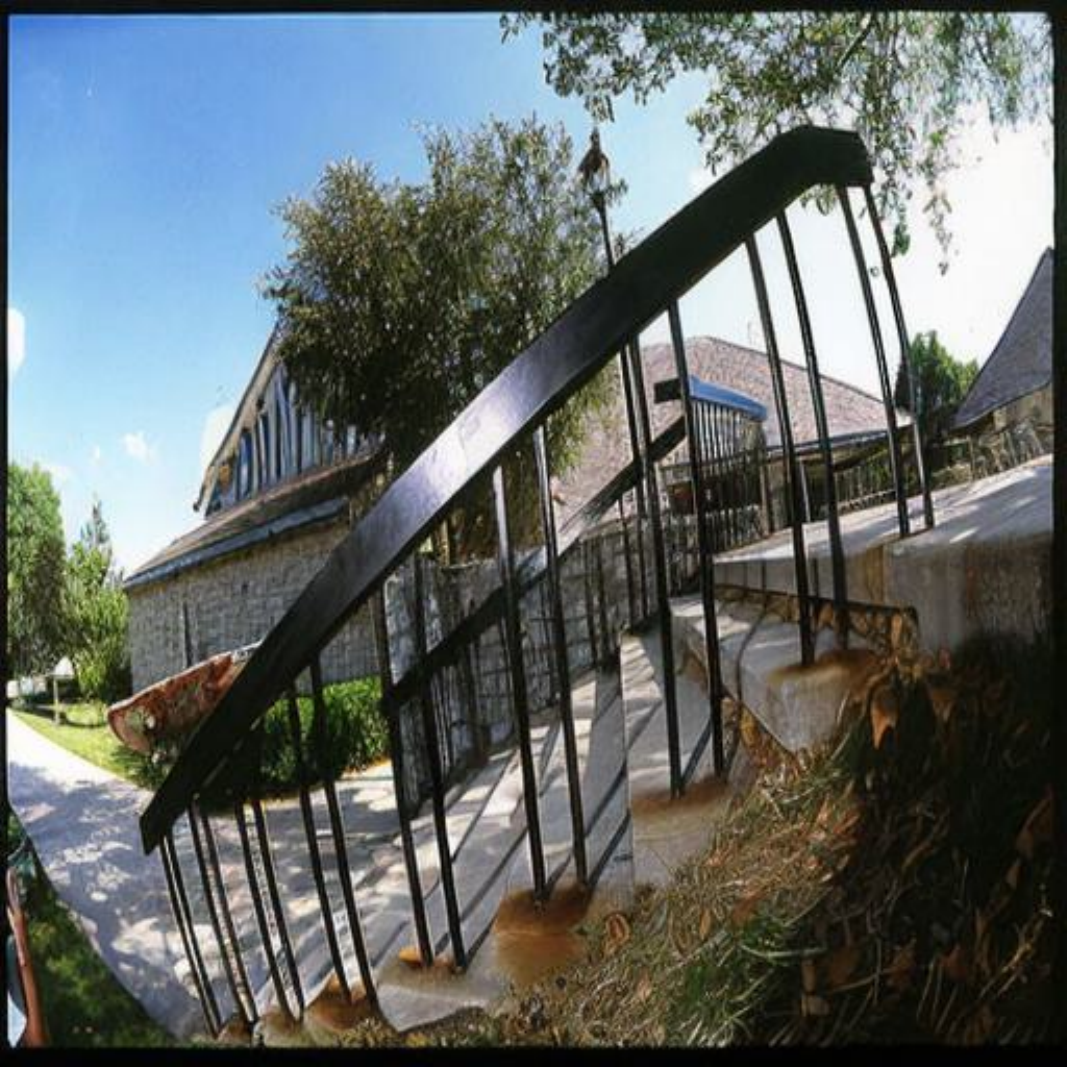}
        \hspace{-0.02\linewidth}
        \includegraphics[width=0.24\linewidth]{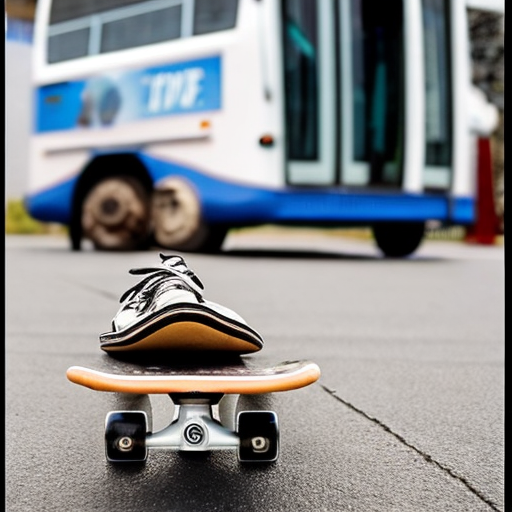}
        \hspace{-0.02\linewidth}
        \includegraphics[width=0.24\linewidth]{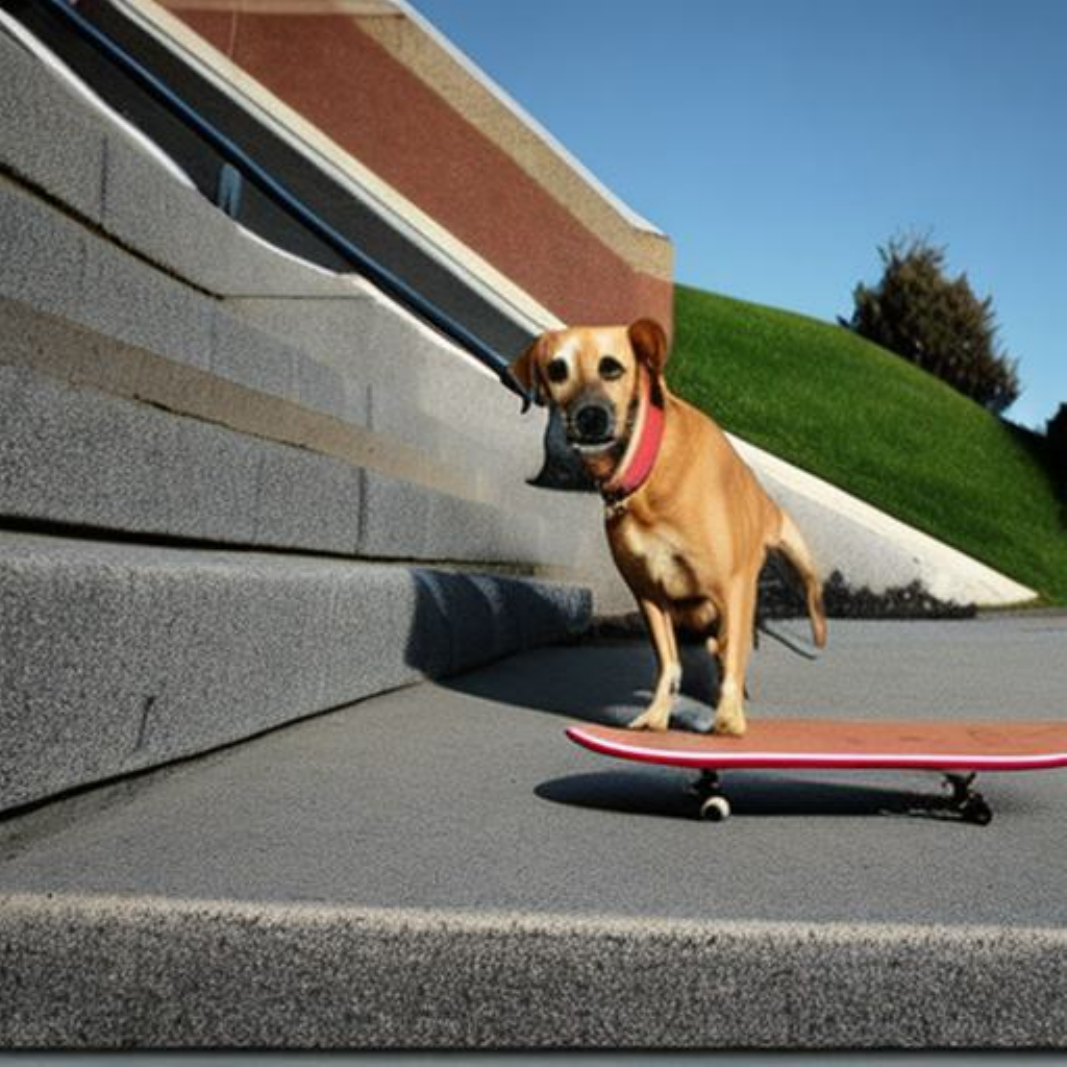}
        \hspace{-0.02\linewidth}
        \includegraphics[width=0.24\linewidth]{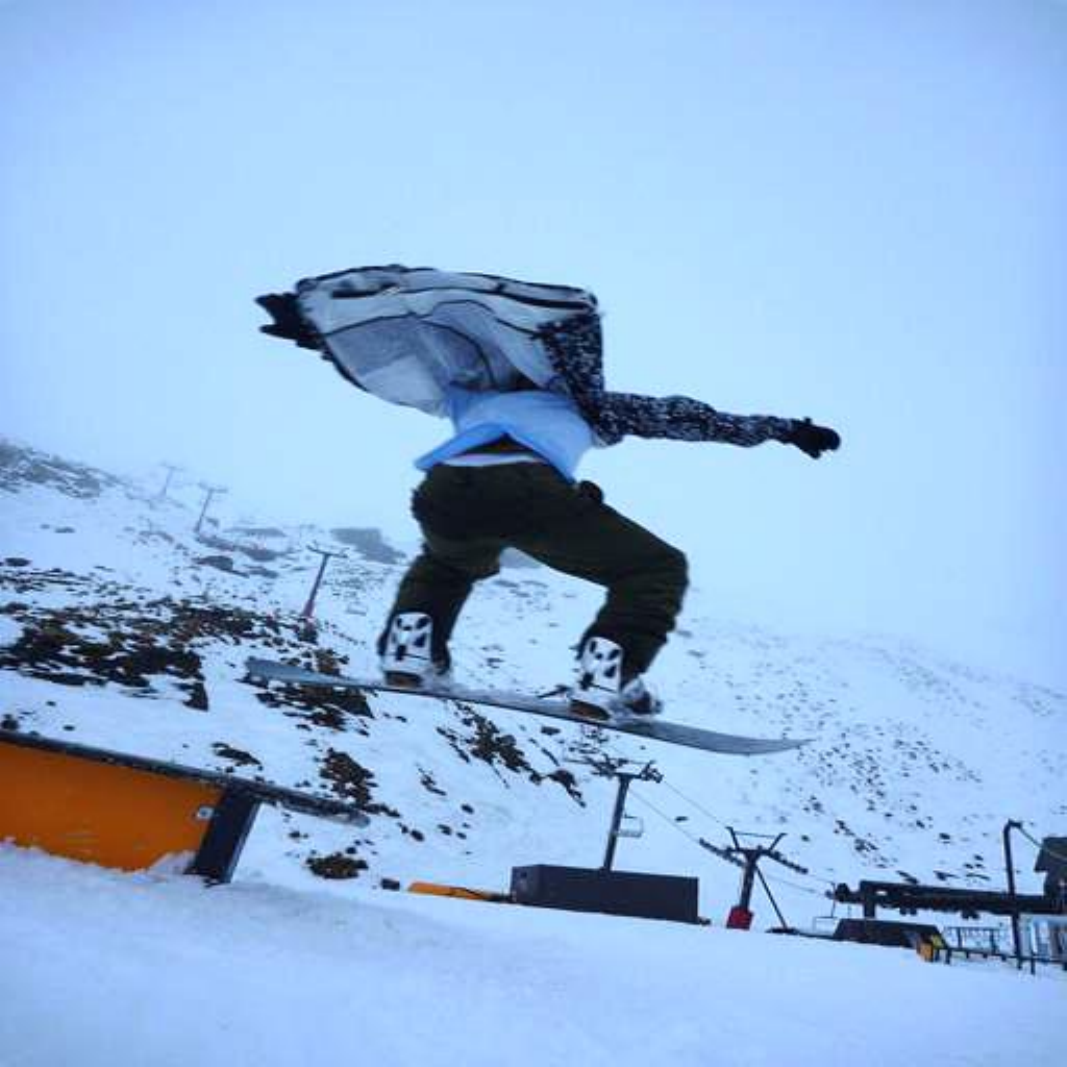}
        \hspace{-0.02\linewidth}
        \includegraphics[width=0.24\linewidth]{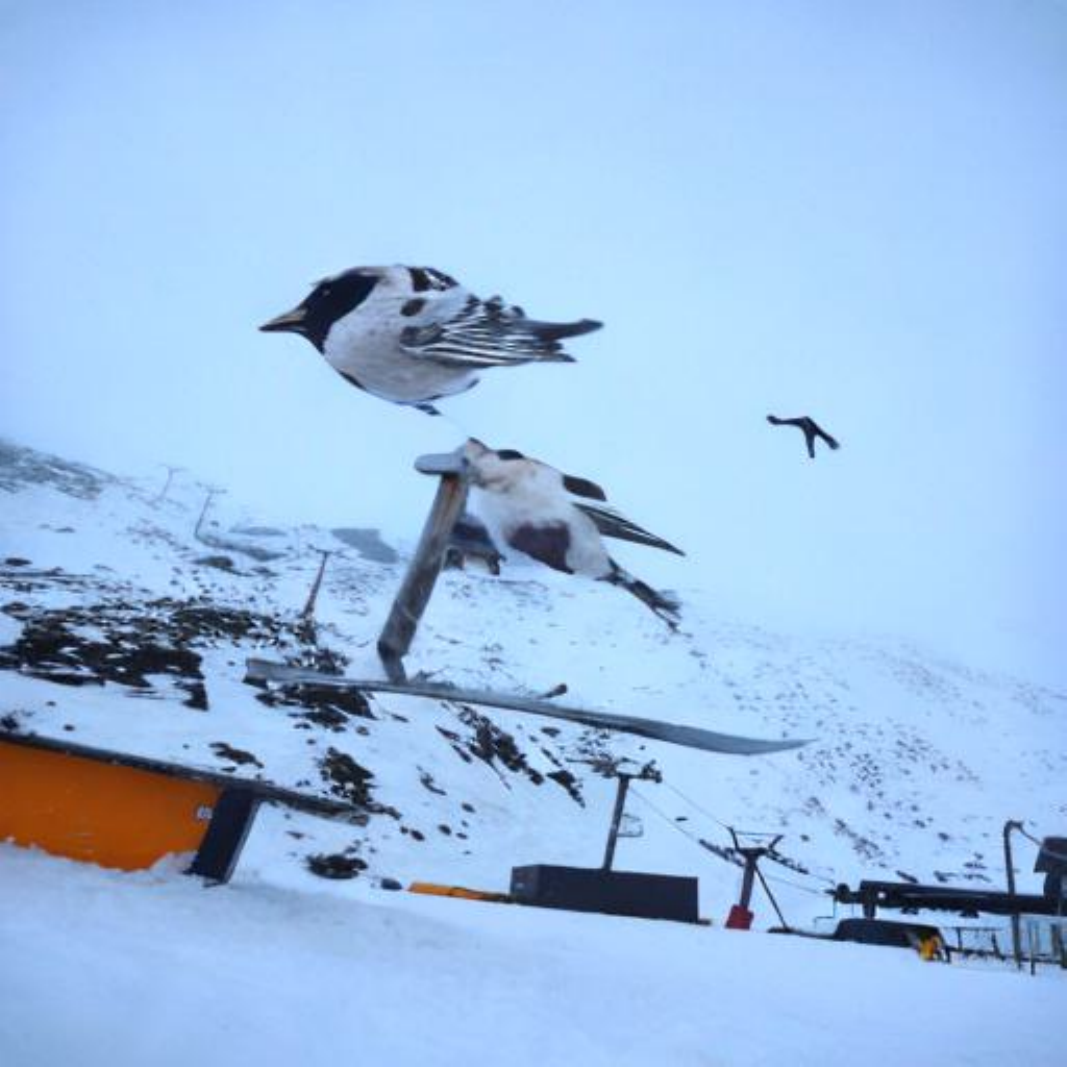}
        \hspace{-0.02\linewidth}
        \includegraphics[width=0.24\linewidth]{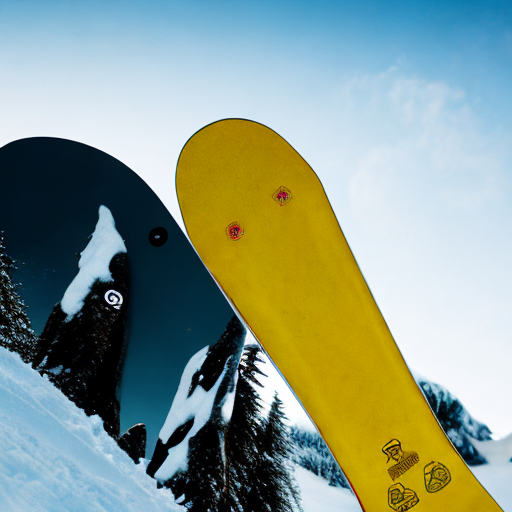}
        \hspace{-0.02\linewidth}
        \includegraphics[width=0.24\linewidth]{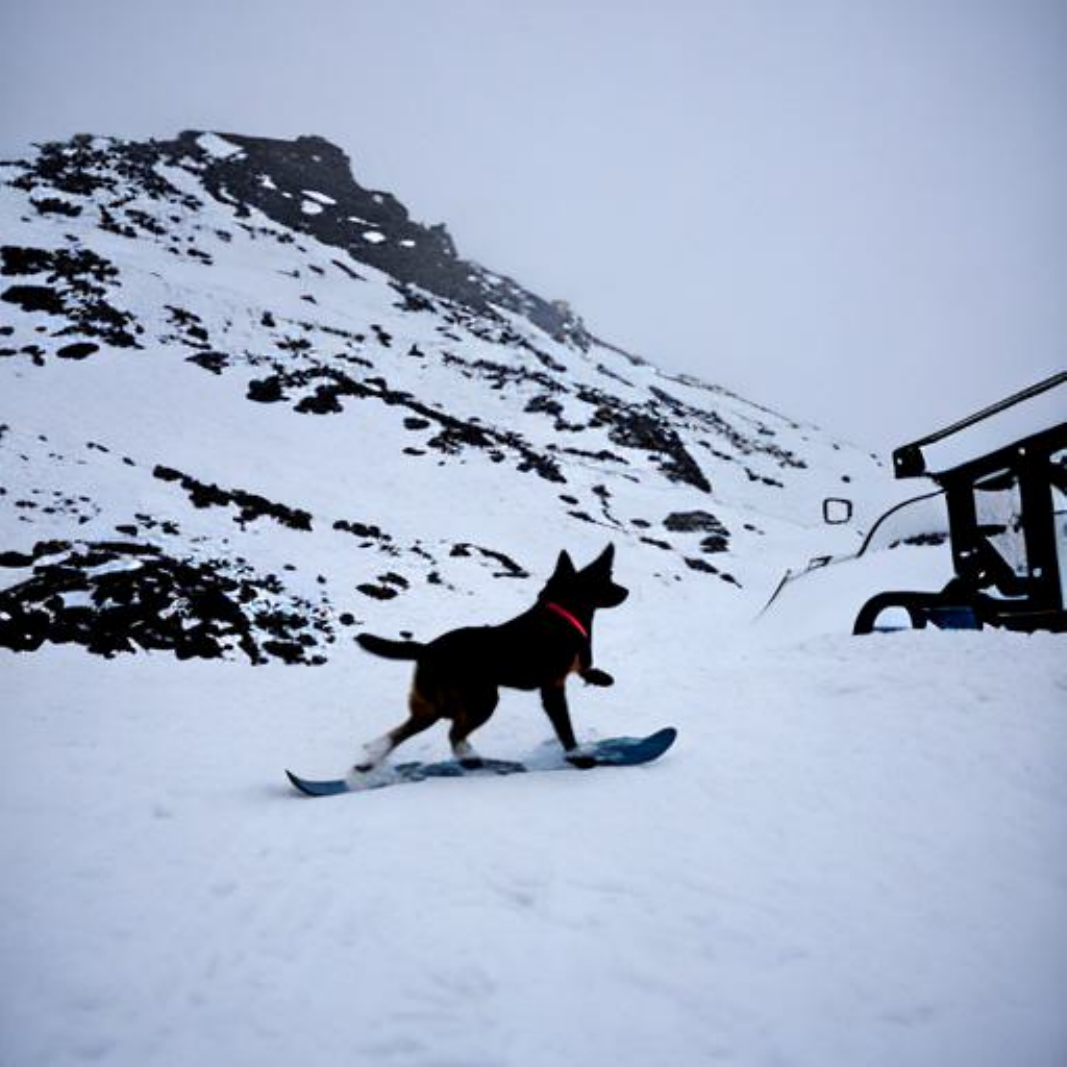}
        \hspace{-0.02\linewidth}
        \includegraphics[width=0.24\linewidth]{figures/result_hard/case2/18798_orig.jpg}
        \hspace{-0.02\linewidth}
        \includegraphics[width=0.24\linewidth]{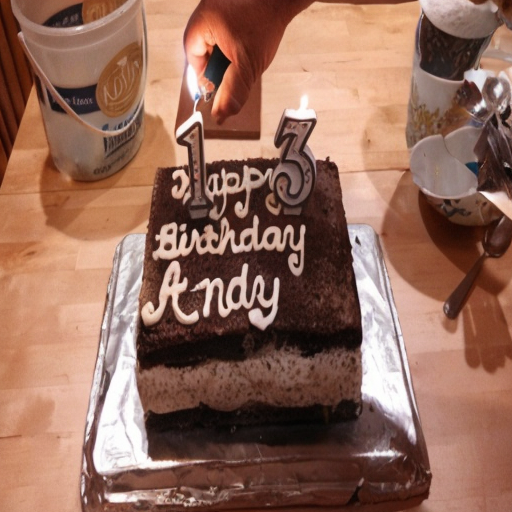}
        \hspace{-0.02\linewidth}
        \includegraphics[width=0.24\linewidth]{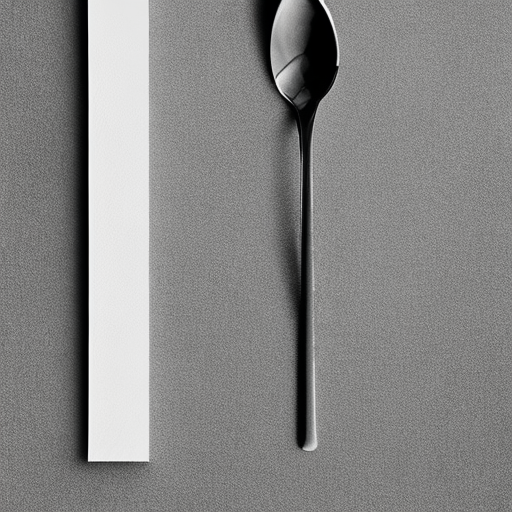}
        \hspace{-0.02\linewidth}
        \includegraphics[width=0.24\linewidth]{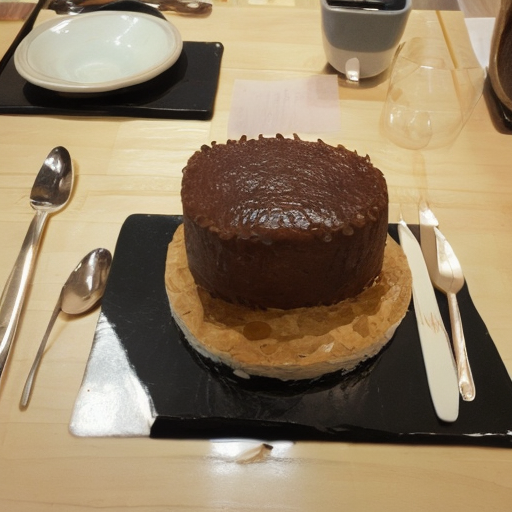}
        \hspace{-0.02\linewidth}
        \includegraphics[width=0.24\linewidth]{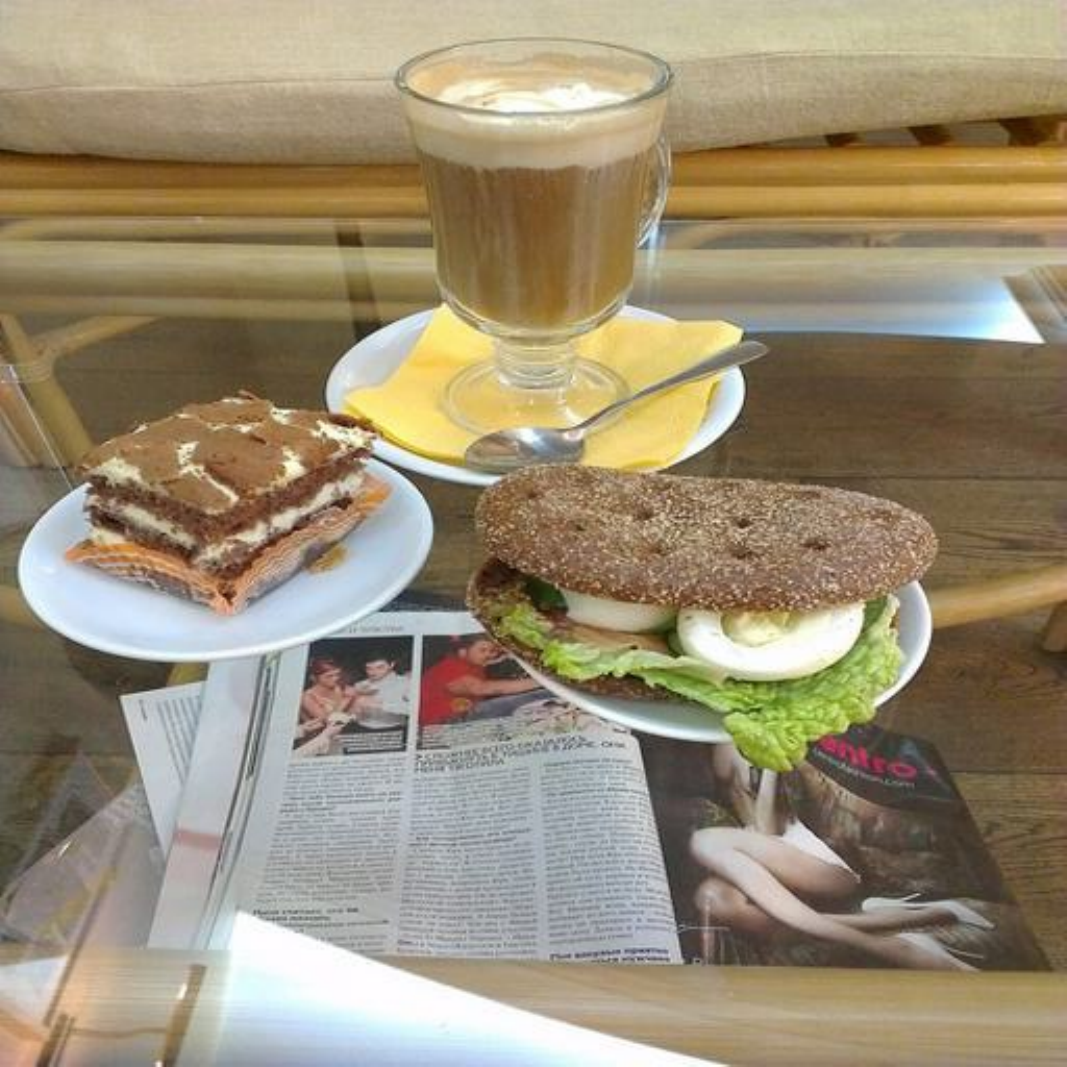}
        \hspace{-0.02\linewidth}
        \includegraphics[width=0.24\linewidth]{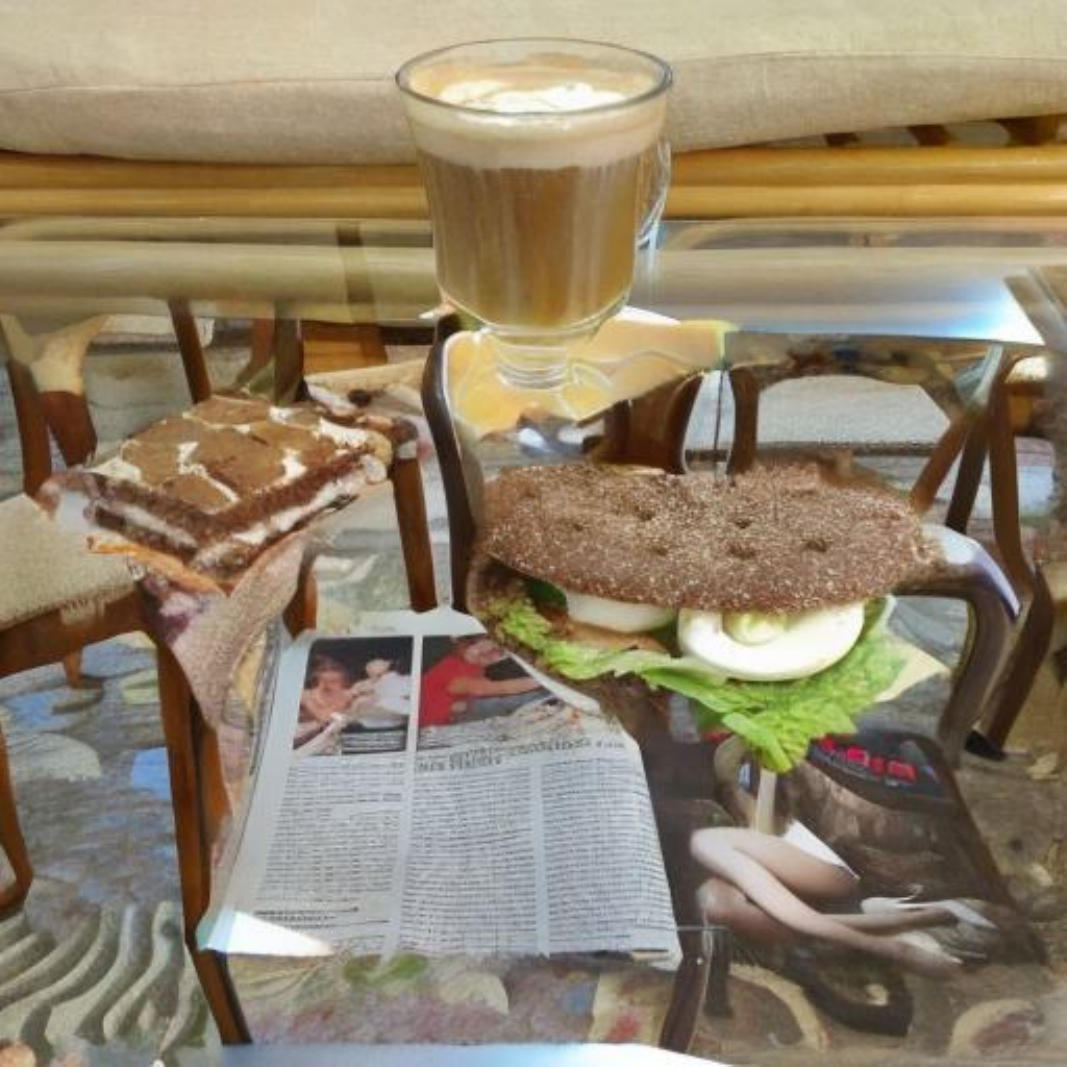}
        \hspace{-0.02\linewidth}
        \includegraphics[width=0.24\linewidth]{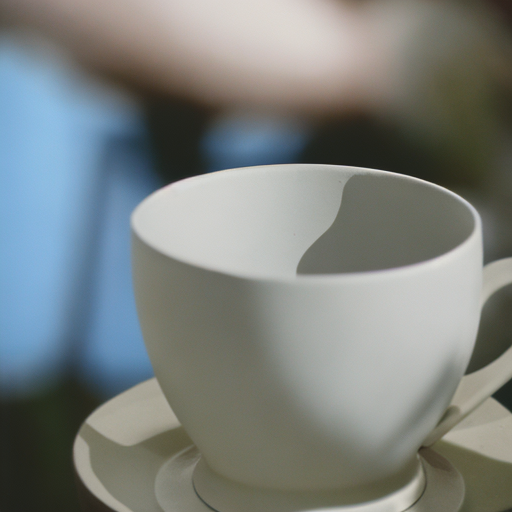}
        \hspace{-0.02\linewidth}
        \includegraphics[width=0.24\linewidth]{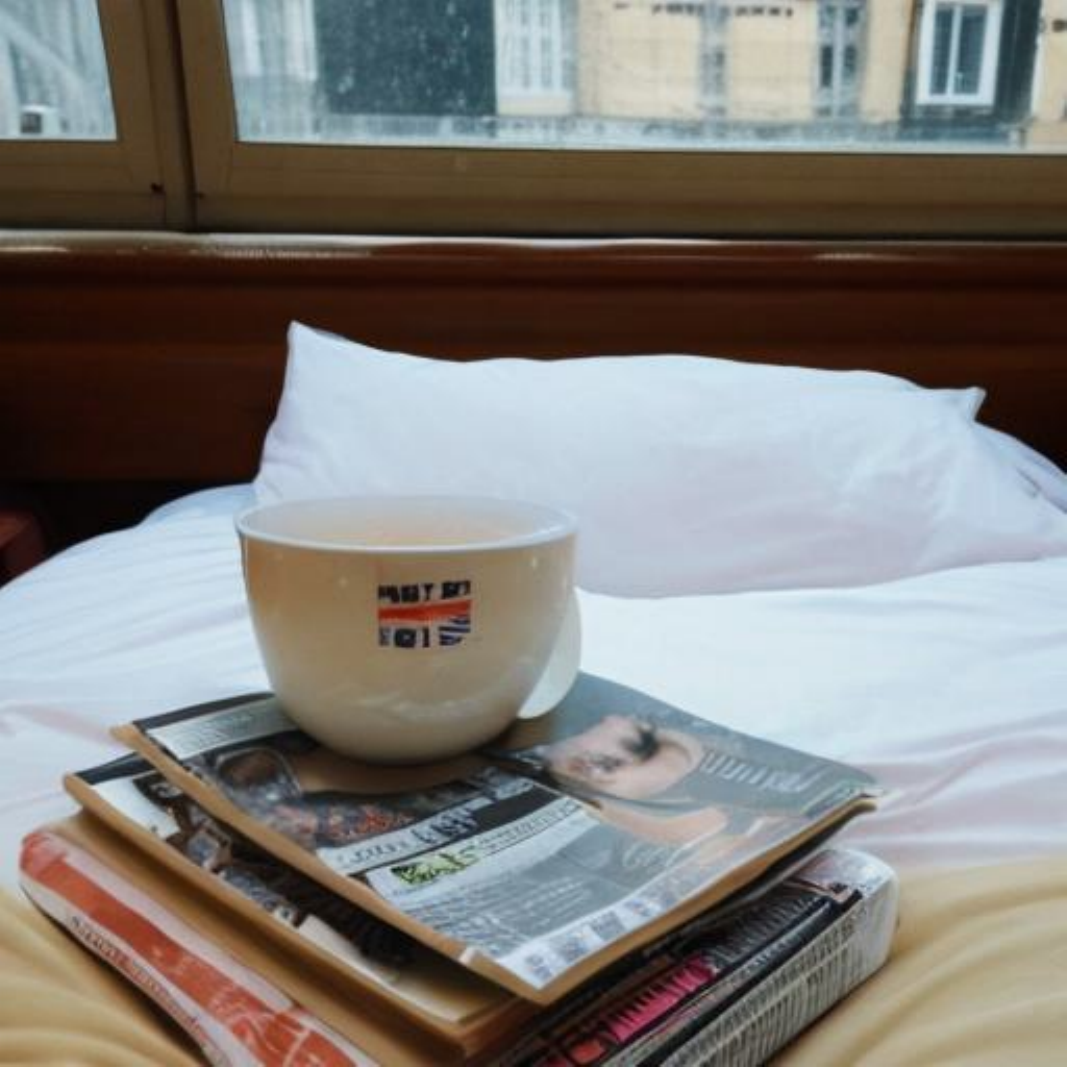}
        \hspace{-0.02\linewidth}
        \includegraphics[width=0.24\linewidth]{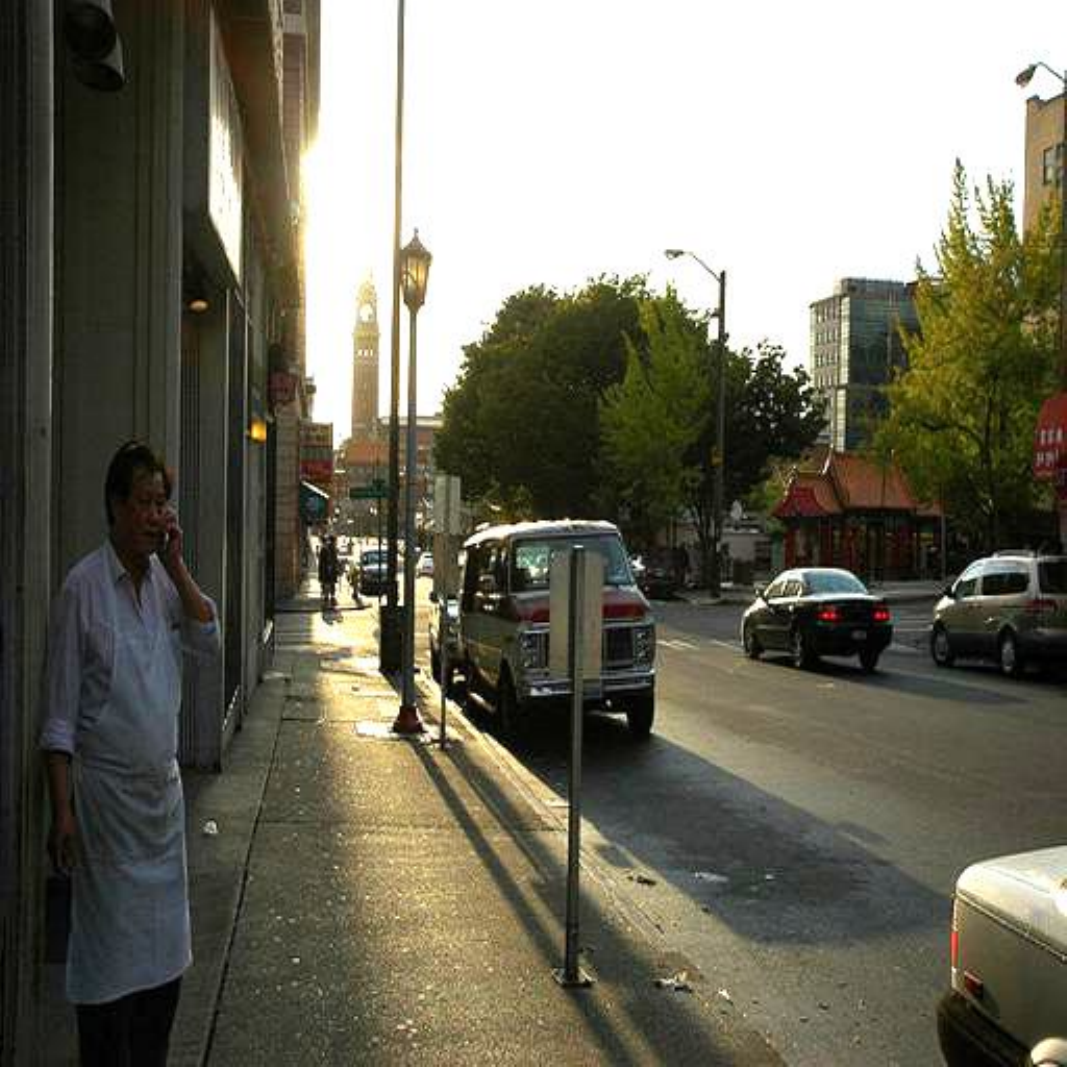}
        \hspace{-0.02\linewidth}
        \includegraphics[width=0.24\linewidth]{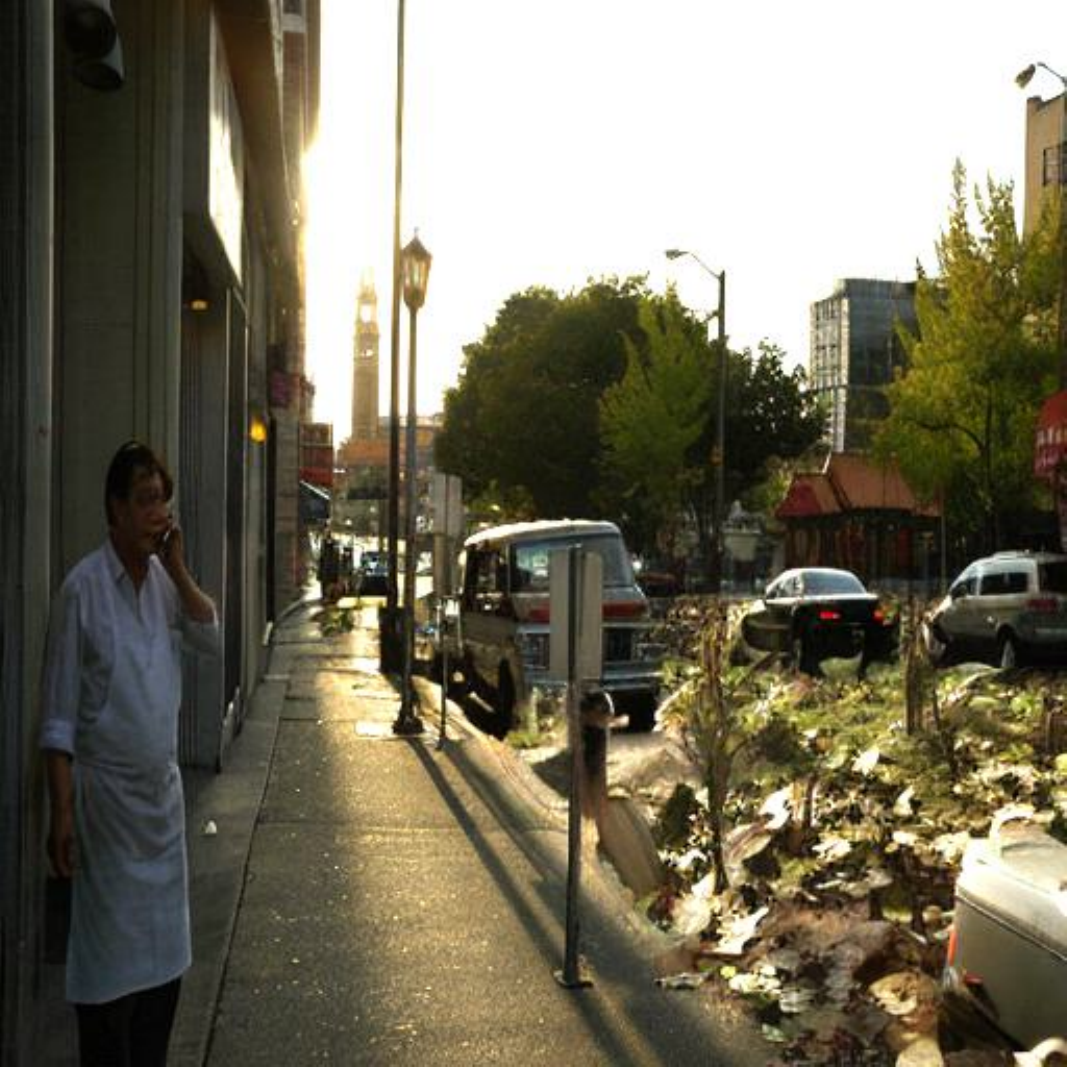}
        \hspace{-0.02\linewidth}
        \includegraphics[width=0.24\linewidth]{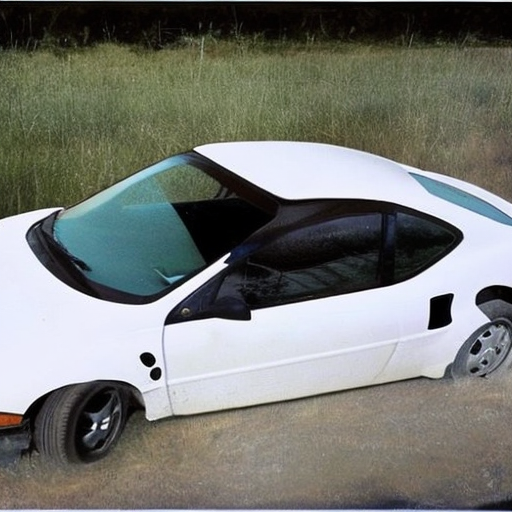}
        \hspace{-0.02\linewidth}
        \includegraphics[width=0.24\linewidth]{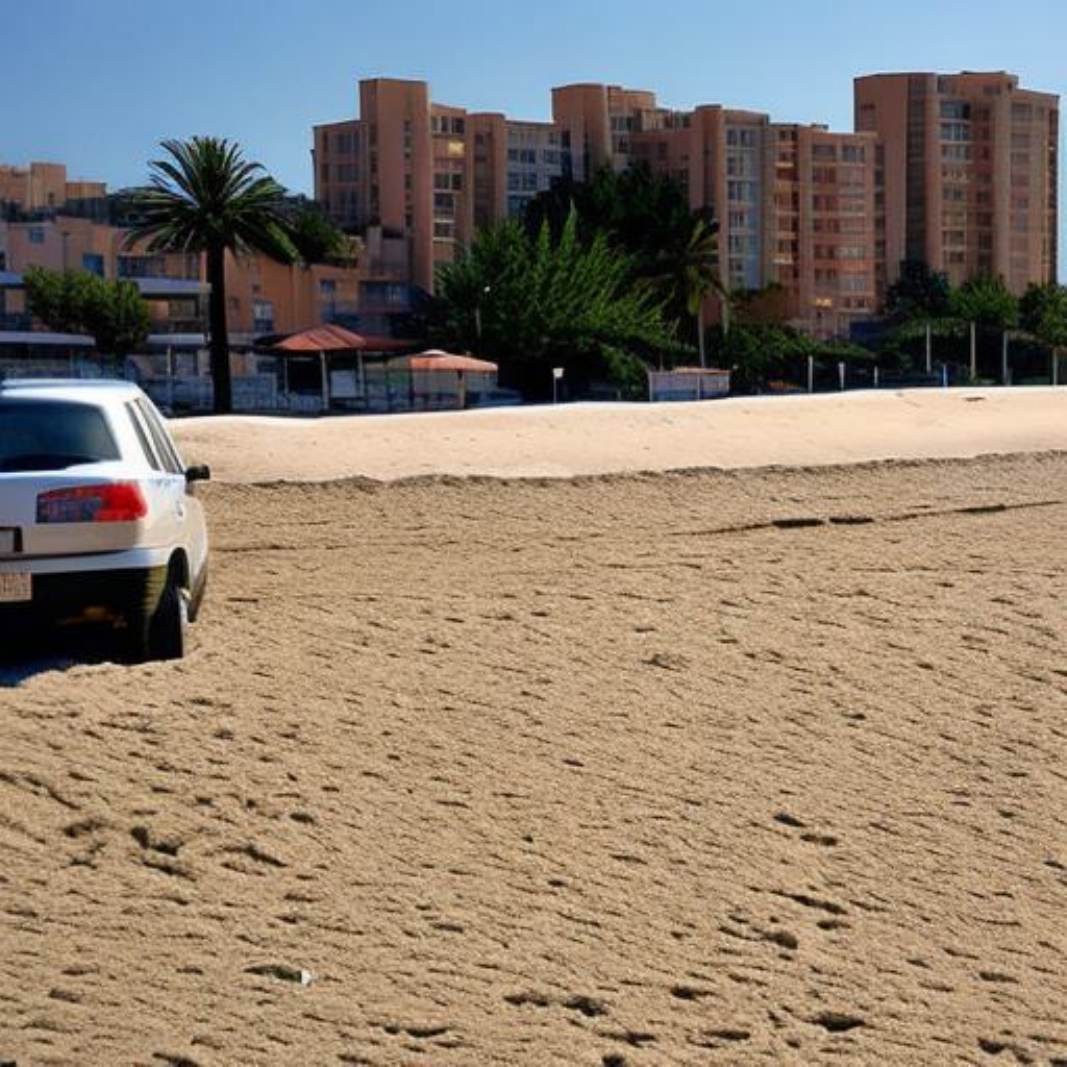}
        \hspace{-0.02\linewidth}
        \includegraphics[width=0.24\linewidth]{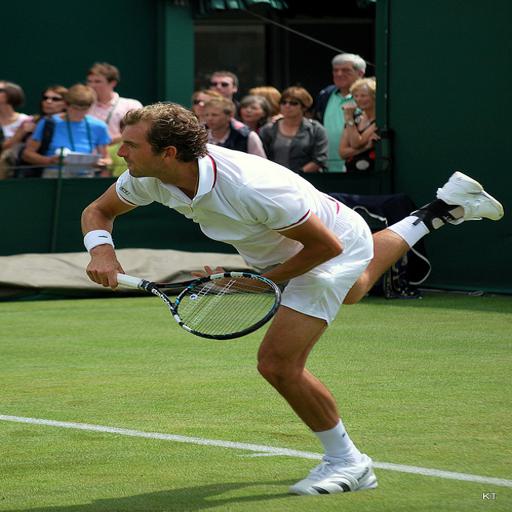}
        \hspace{-0.02\linewidth}
        \includegraphics[width=0.24\linewidth]{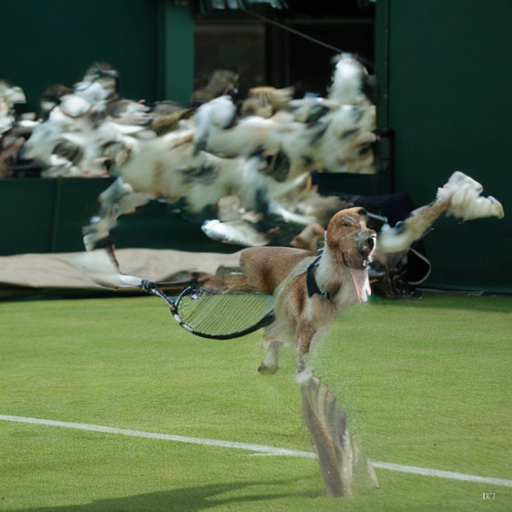}
        \hspace{-0.02\linewidth}
        \includegraphics[width=0.24\linewidth]{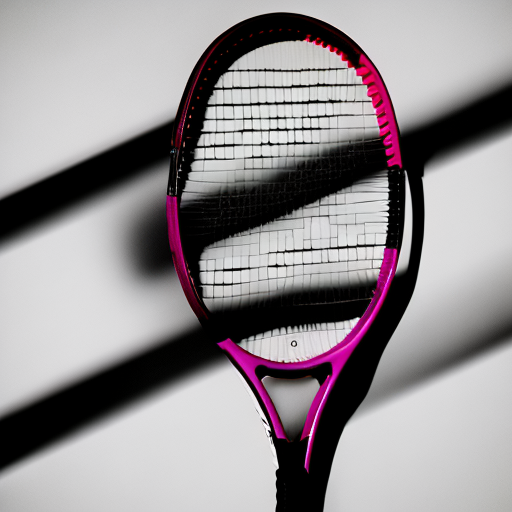}
        \hspace{-0.02\linewidth}
        \includegraphics[width=0.24\linewidth]{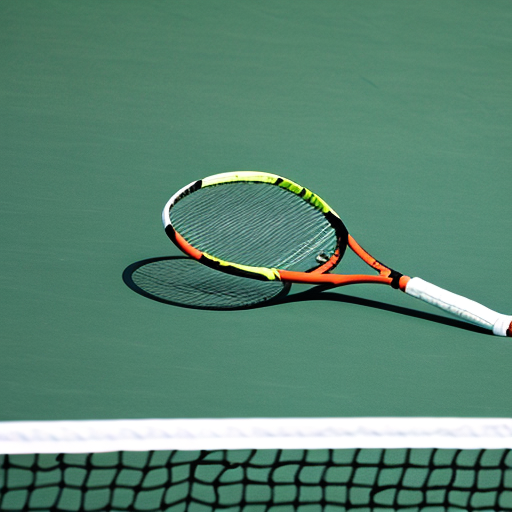}
        \hspace{-0.02\linewidth}
        \begin{subfigure}[b]{0.24\linewidth}
            \includegraphics[width=\linewidth]{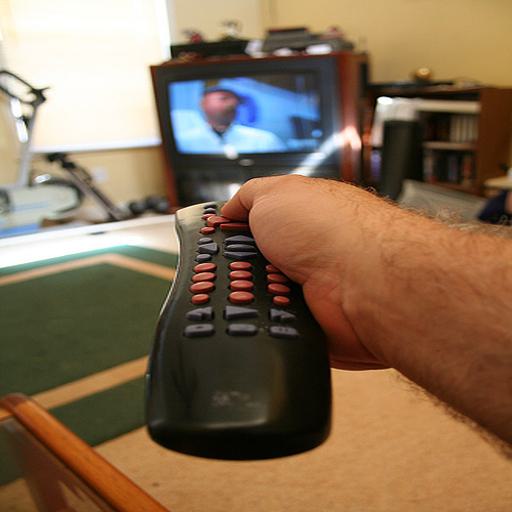}
            \caption{Original}
        \end{subfigure}
        \hspace{-0.02\linewidth}
        \begin{subfigure}[b]{0.24\linewidth}
            \includegraphics[width=\linewidth]{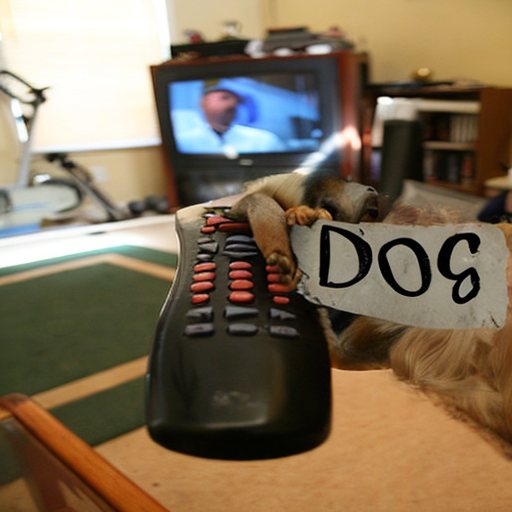}
            \caption{Blended LD}
            \label{fig:qualex_local_app}
        \end{subfigure}
        \hspace{-0.02\linewidth}
        \begin{subfigure}[b]{0.24\linewidth}
            \includegraphics[width=\linewidth]{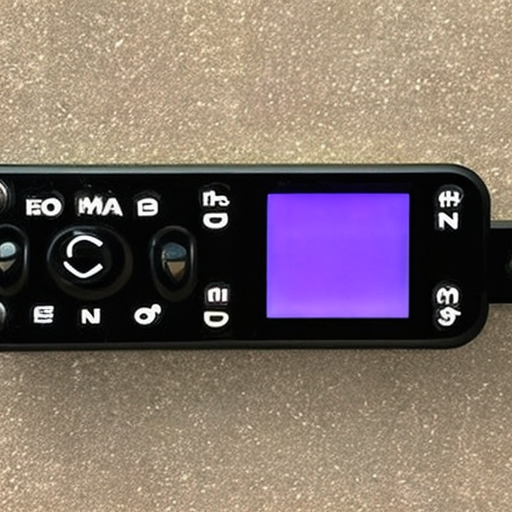}
            \caption{Txt2Img FG}
            \label{fig:qualex_txt2img_app}
        \end{subfigure}
        \hspace{-0.02\linewidth}
        \begin{subfigure}[b]{0.24\linewidth}
            \includegraphics[width=\linewidth]{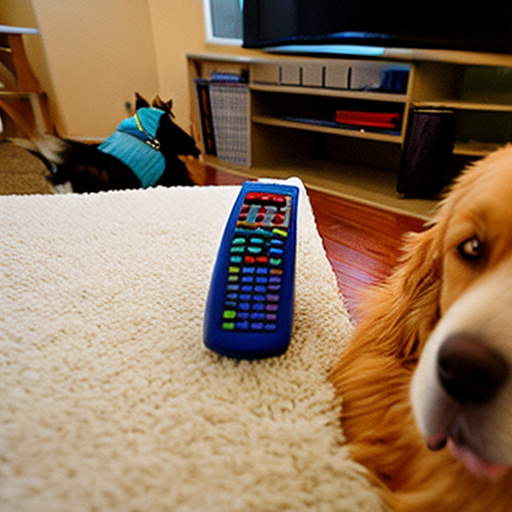}
            \caption{DecoupleGen}
            \label{fig:qualex_ours_app}
        \end{subfigure}
    \caption{\textbf{More examples of qualitative comparisons.}
    Local image editing models, e.g. Blended LD \citep{avrahami2023blended}, often fail to keep generate reasonable and coherent images of complex scenes. Feedback-guided generations \citep{hemmat2023feedback} fail to preserve similar visual appearance in natural images. Instead, DecoupleGen can handle complex relationship between objects and scene, while capable of maintaining similar visual details from original images.}
    \label{fig:qual_ex_app}
    \vspace{-5mm}
\end{figure}

\paragraph{Additional qualitative examples}
We provide additional qualitative comparisons in Figure \ref{fig:qual_ex_app}. Compared to local image editing models (Figure \ref{fig:qualex_local_app}), i.e. Blended Latent Diffusion \citep{avrahami2023blended}, \proj{} can handle complex relationships between objects and scenes, for example when objects interact with each other or when transforming images into a different contextual background (Figure \ref{fig:qualex_ours_app}). Compared to Txt2Img models with Feedback Guidance \citep{hemmat2023feedback} (Figure \ref{fig:qualex_txt2img_app}), our method is able to preserve similar visual information as in original samples (Figure \ref{fig:qualex_ours_app}).

\begin{figure}[!htb]
    \centering
    \includegraphics[width=0.49\linewidth]{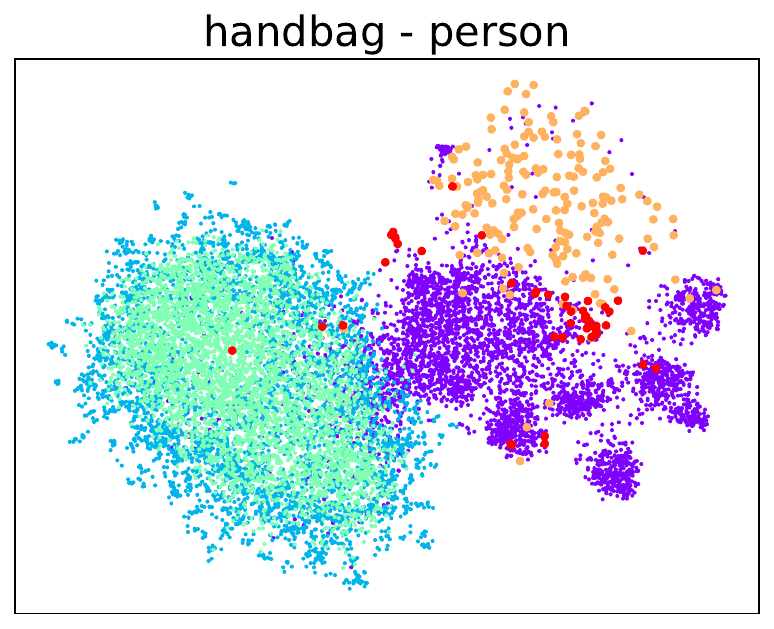}
    \includegraphics[width=0.49\linewidth]{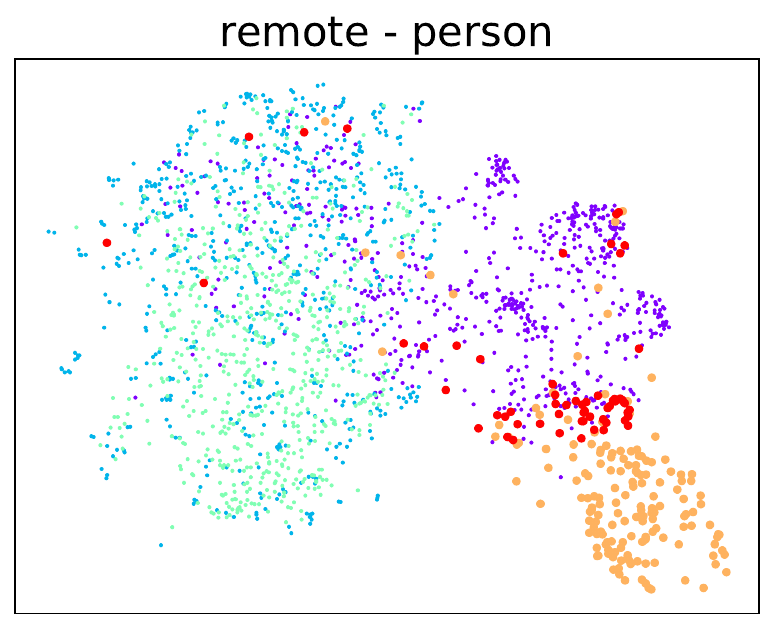}
    \includegraphics[width=0.49\linewidth]{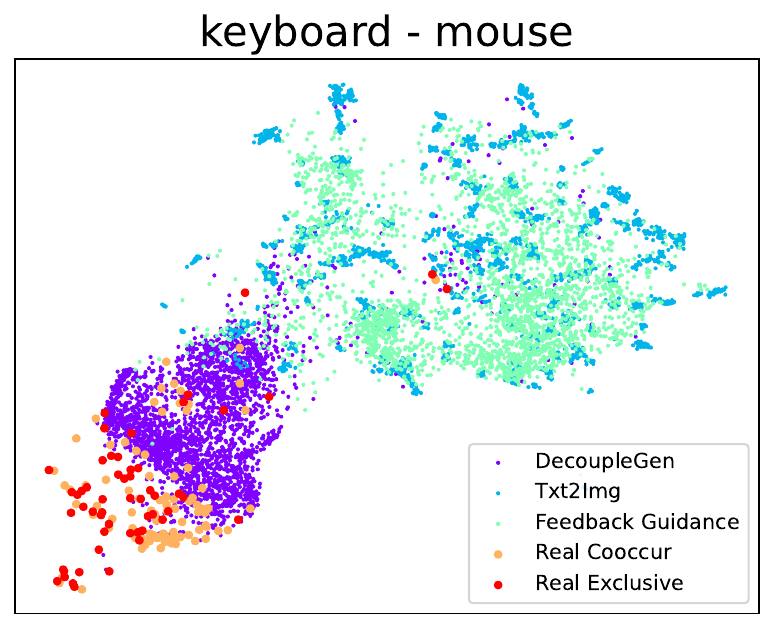}
    \includegraphics[width=0.49\linewidth]{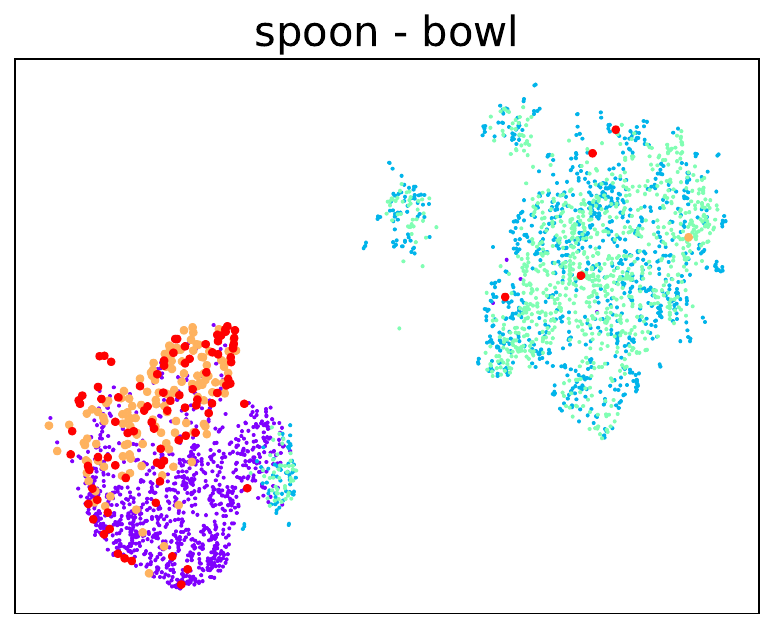}
    \caption{\textbf{UMAP visualizations of real samples and generations.}
    We plot real exclusive and cooccur images from validation split, together with synthetic samples from different generation approaches. Generations from DecoupleGen overall stay closer to real data distribution than those from other methods, thus providing information more relevant to downstream tasks.}
    \label{fig:umap_coco_app}
\end{figure}

\vspace{-3mm}
\paragraph{Additional UMAP visualizations}
To demonstrate that generated samples from \proj{} are able to better adapt to downstream datasets than common Txt2Img models, we provide UMAP visualizations for more classes in Figure~\ref{fig:umap_coco_app}. We extract visual features from ResNet50 pre-trained on ImageNet and use them as input to UMAP projections. We observe that real exclusive and cooccur images are more separable from each other for classes \texttt{handbag} and \texttt{remote}, while for classes \texttt{keyboard} and \texttt{spoon} there are not significantly distinct clusters. In both cases our method is able to generate samples more similar to original datasets in their UMAP projections, which demonstrates effectiveness of our adapted finetuning process.

\begin{figure}[!htb]
    \centering
    \includegraphics[width=0.24\linewidth]{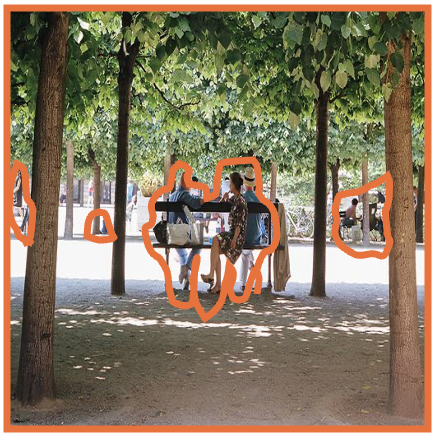}
    \hspace{-0.02\linewidth}
    \includegraphics[width=0.24\linewidth]{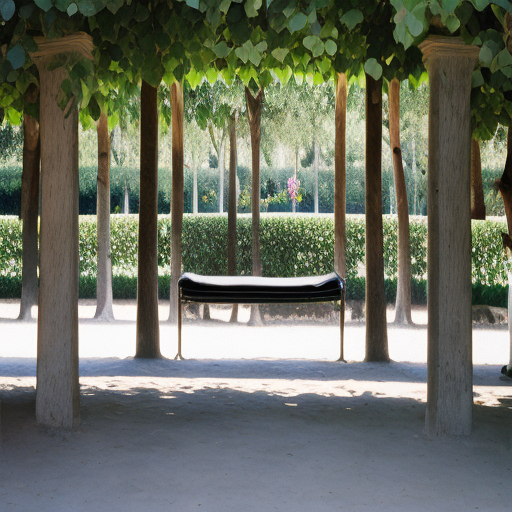}
    \hspace{-0.02\linewidth}
    \includegraphics[width=0.24\linewidth]{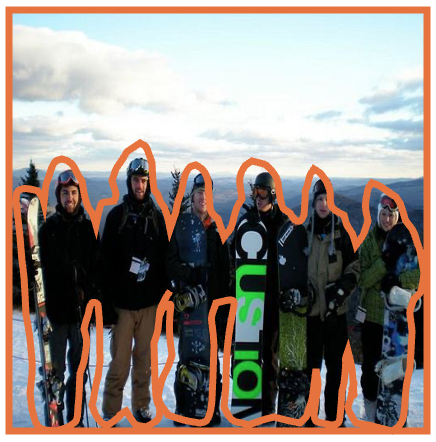}
    \hspace{-0.02\linewidth}
    \includegraphics[width=0.24\linewidth]{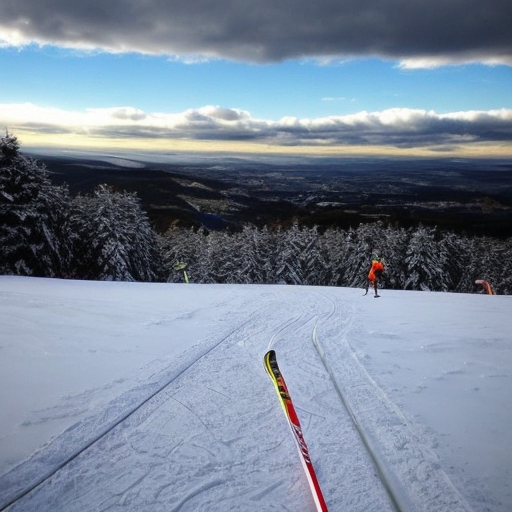}
    \caption{\textbf{Background reconstruction capability.}
    We provide background mask (orange) at training time and generate from text input ``A photo of \texttt{[Vbackground]}" at inference time. Our finetuned token can capture most visual details from original images.}
    \label{fig:exp_vbg}
    \vspace{-5mm}
\end{figure}

\vspace{-3mm}
\paragraph{How good is the capability of preserving visual details?} 
Our approach preserves visual details by learning a new special token \texttt{[Vbackground]} and use this word as part of text input at inference time. Hence, the background reconstruction ability could indicate whether generations would stay similar to original images. Figure \ref{fig:exp_vbg} shows generations from fine-tuned models with text input ``A photo of \texttt{[Vbackground]}". This further justifies that our adapted personalized finetuning process is able to preserve visual details from original datasets.

\begin{figure}[!htb]
    \centering
    \begin{subfigure}[b]{0.24\linewidth}
        \includegraphics[width=\linewidth]{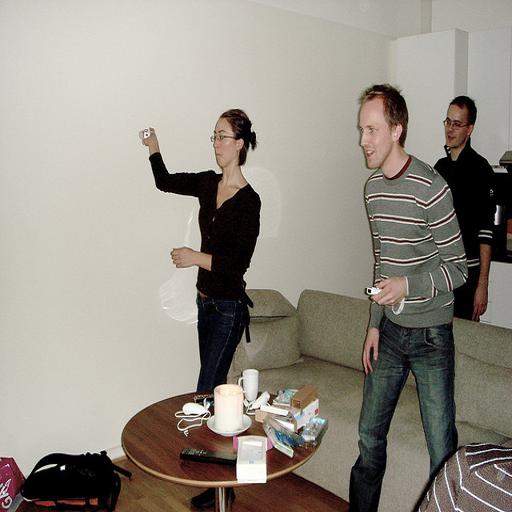}
        \caption{Original}
        \label{fig:exp_fail_orig1}
    \end{subfigure}
    \hspace{-0.02\linewidth}
    \begin{subfigure}[b]{0.24\linewidth}
        \includegraphics[width=\linewidth]{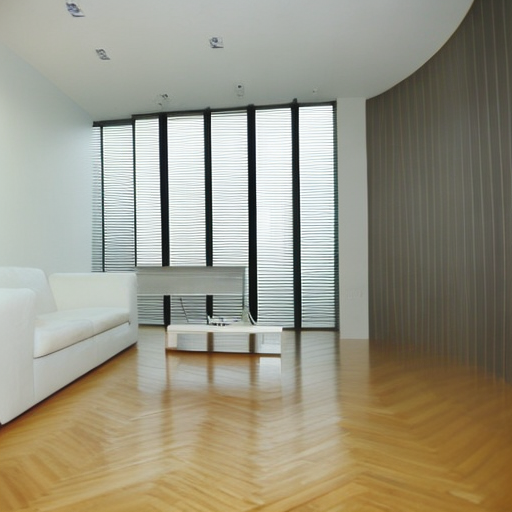}
        \caption{\texttt{[Vremote]}}
        \label{fig:exp_fail_gen1}
    \end{subfigure}
    \hspace{-0.02\linewidth}
    \begin{subfigure}[b]{0.24\linewidth}
        \includegraphics[width=\linewidth]{figures/result_fail/58283_orig.jpg}
        \caption{Original}
        \label{fig:exp_fail_orig2}
    \end{subfigure}
    \hspace{-0.02\linewidth}
    \begin{subfigure}[b]{0.24\linewidth}
        \includegraphics[width=\linewidth]{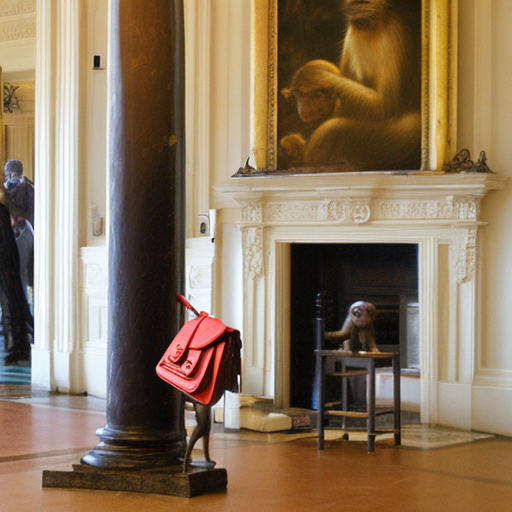}
        \caption{Miss details}
        \label{fig:exp_fail_gen2}
    \end{subfigure}
    \caption{\textbf{Examples of failure cases.}
    Limitations of our approach include when mask regions of object class, e.g. remote, are very small, and when original image, e.g. handbag and person, contains very complex visual information, generations fail to replicate some details, such as cake and dining table.}
    \label{fig:exp_fail}
    \vspace{-5mm}
\end{figure}

\paragraph{Failure cases and limitations} 
The personalization finetuning approach in every image is not perfect, as shown in Figure \ref{fig:exp_fail}. Specifically, it is not very good at scenarios where objects are very small, where such thin masks usually do not have enough learning signals, and our new word tokens fail to learn semantic information (Figure \ref{fig:exp_fail_orig1}). For example, prompted fine-tuned models to generate "A photo of \texttt{[Vremote]}", the new word token fails to learn semantic information corresponding to the remote object (Figure \ref{fig:exp_fail_gen1}). Therefore, it always fails to include the remote object in generations. 
Additionally, when the original image contains very complex visual details, for example a cooccur image of handbag and person containing cake and dining table (Figure \ref{fig:exp_fail_orig2}), generated samples usually cannot fully replicate all of contents. For example, such generations may fail to reconstruct the table and cake while removing person from the image (Figure \ref{fig:exp_fail_gen2}).

\subsection{Further Analysis of DecoupleGen}

\begin{figure}[!htb]
    \centering
    \begin{subfigure}[b]{0.24\linewidth}
        \includegraphics[width=\linewidth]{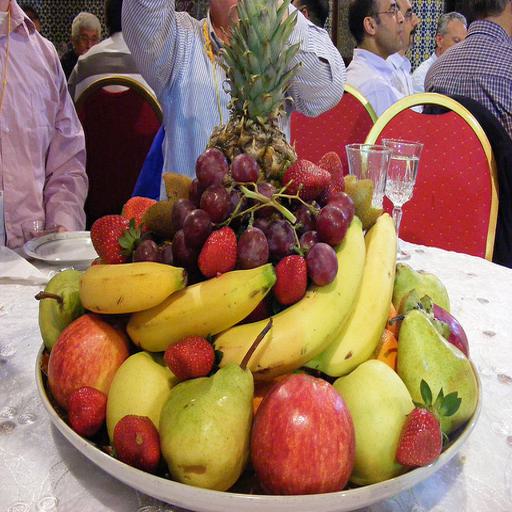}
        \caption{\centering Wine glass and Person}
        \label{fig:shift_qualex_orig1}
    \end{subfigure}
    \hspace{-0.02\linewidth}
    \begin{subfigure}[b]{0.24\linewidth}
        \includegraphics[width=\linewidth]{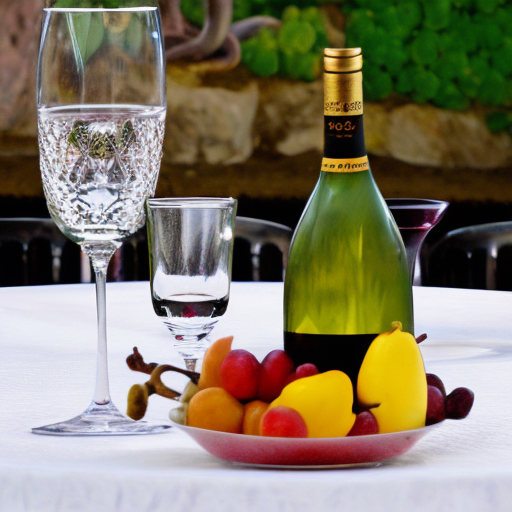}
        \caption{\centering Person removed}
        \label{fig:shift_qualex_ours1}
    \end{subfigure}
    \hspace{-0.02\linewidth}
    \begin{subfigure}[b]{0.24\linewidth}
        \includegraphics[width=\linewidth]{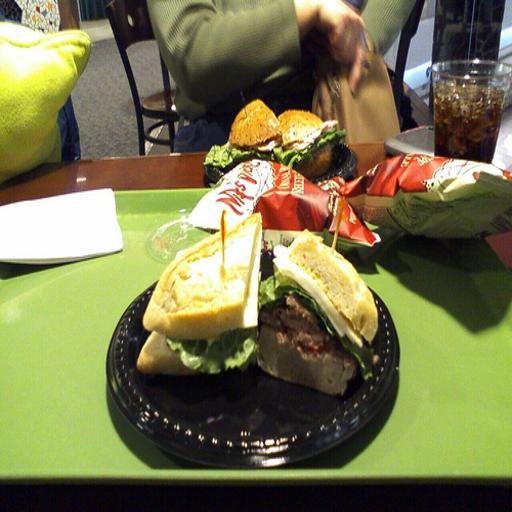}
        \caption{\centering Cup and Dining table}
        \label{fig:shift_qualex_orig2}
    \end{subfigure}
    \hspace{-0.02\linewidth}
    \begin{subfigure}[b]{0.24\linewidth}
        \includegraphics[width=\linewidth]{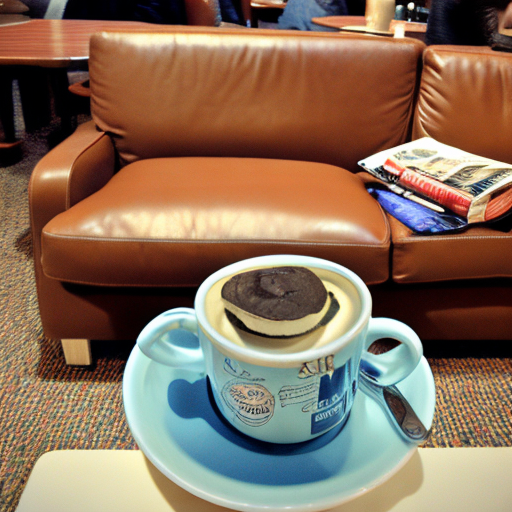}
        \caption{\centering Spoon included}
        \label{fig:shift_qualex_ours2}
    \end{subfigure}
    \caption{\textbf{Examples of newly introduced common contextual patterns.}
    We observe that such cooccurring patterns can either come from original images and become strengthened, or are implicitly introduced by generative models.}
    \label{fig:shift_qualex}
    \vspace{-5mm}
\end{figure}

\paragraph{Impact on contextual patterns.}
As shown in the main paper, while \proj{} is able to reduce contextual patterns that are targeted to improve, it may introduce new common cooccurring object-context pairs even though not specified. In our experiments, such examples include \texttt{wine glass} - \texttt{dining table} and \texttt{spoon} - \texttt{cup}. Figure \ref{fig:shift_qualex} shows some of these scenarios.

In some cases, such cooccurring patterns can come from original datasets, for example dining table exists as well in the cooccur image of wine glass and person (Figure \ref{fig:shift_qualex_orig1}). Generated samples from \proj{} remove person successfully but leave dining table preserved (Figure \ref{fig:shift_qualex_ours1}), further strengthening additional biases in original datasets. 
In other cases, such cooccurring patterns can be implicitly introduced by Txt2Img models, for example when handling a pair of cup and dining table (Figure \ref{fig:shift_qualex_orig2}), the generative model automatically includes a spoon together with the cup (Figure \ref{fig:shift_qualex_ours2}), even though spoons do not exist in original image or is not specified as part of text input to generative model. Therefore, although our experiments demonstrate improvements in target biased cooccurring pairs, we cannot completely avoid introducing new contextual patterns.

\paragraph{Model accuracy on unbiased classes also improves.}
We report AP values on each unbiased classes from models trained in our approach in Table~\ref{tab:unbiased_accs}.
Our results show that model accuracy on unbiased classes improves from standard classifier by 1.3\% increase in mAP as well. This is a smaller improvements compared to biased classes, where we have 3.68\% and 2.24\% improvement for exclusive and cooccur images respectively. Among 56 unbiased classes, after ignoring those containing very few data points, 1 class decreases, 15 classes stayed within error bar, and 40 classes significantly improved.

\begin{table}[!ht]
    \centering
    \resizebox{\linewidth}{!}{
    \begin{tabular}{lc|lc}
        \toprule
        Unbiased class & AP value & Unbiased class & AP value \\
        \midrule
        person & 98.04 $\pm$ 0.00 (\textcolor{blue}{+0.31}) & knife & 46.92 $\pm$ 0.48 (\textcolor{blue}{+1.59}) \\ 
        bicycle & 62.52 $\pm$ 0.51 (\textcolor{blue}{+3.58}) & bowl & 61.81 $\pm$ 0.40 (\textcolor{blue}{+1.99}) \\ 
        motorcycle & 87.41 $\pm$ 0.19 (\textcolor{blue}{+0.84}) & banana & 73.26 $\pm$ 0.11 (\textcolor{blue}{+0.58}) \\ 
        airplane & 95.09 $\pm$ 0.22 (\textcolor{gray}{+0.36}) & apple & 53.34 $\pm$ 0.28 (\textcolor{blue}{+1.28}) \\ 
        train & 95.06 $\pm$ 0.10 (\textcolor{blue}{+0.78}) & sandwich & 61.59 $\pm$ 0.29 (\textcolor{blue}{+1.43}) \\ 
        truck & 66.33 $\pm$ 0.20 (\textcolor{blue}{+3.29}) & orange & 64.61 $\pm$ 0.33 (\textcolor{blue}{+0.77}) \\ 
        boat & 83.94 $\pm$ 0.29 (\textcolor{blue}{+2.09}) & broccoli & 80.32 $\pm$ 0.34 (\textcolor{gray}{+0.03}) \\ 
        traffic light & 74.90 $\pm$ 0.17 (\textcolor{blue}{+1.90}) & carrot & 60.38 $\pm$ 0.36 (\textcolor{blue}{+2.15}) \\ 
        fire hydrant& 72.84 $\pm$ 0.67 (\textcolor{gray}{+0.16}) & hot dog & 60.71 $\pm$ 0.69 (\textcolor{gray}{-0.72}) \\ 
        stop sign & 67.85 $\pm$ 0.43 (\textcolor{gray}{-0.26}) & pizza & 84.60 $\pm$ 0.22 (\textcolor{gray}{+0.31}) \\ 
        bench & 59.70 $\pm$ 0.44 (\textcolor{blue}{+3.64}) & donut & 70.79 $\pm$ 0.13 (\textcolor{blue}{+0.78}) \\ 
        bird & 70.55 $\pm$ 0.30 (\textcolor{blue}{+1.76}) & cake & 71.25 $\pm$ 0.30 (\textcolor{blue}{+0.90}) \\ 
        cat & 88.62 $\pm$ 0.09 (\textcolor{gray}{+0.50}) & chair & 71.10 $\pm$ 0.08 (\textcolor{blue}{+3.30}) \\ 
        dog & 77.16 $\pm$ 0.32 (\textcolor{blue}{+1.24}) & couch & 77.12 $\pm$ 0.23 (\textcolor{blue}{+3.58}) \\ 
        horse & 88.18 $\pm$ 0.31 (\textcolor{gray}{+0.05}) & bed & 83.49 $\pm$ 0.14 (\textcolor{blue}{+2.18}) \\ 
        sheep & 90.75 $\pm$ 0.16 (\textcolor{blue}{+0.65}) & dining table & 75.15 $\pm$ 0.14 (\textcolor{blue}{+1.65}) \\ 
        cow & 83.35 $\pm$ 0.46 (\textcolor{gray}{-0.04}) & toilet & 93.58 $\pm$ 0.10 (\textcolor{blue}{+0.39}) \\ 
        elephant & 96.82 $\pm$ 0.12 (\textcolor{gray}{-0.20}) & tv & 80.43 $\pm$ 0.21 (\textcolor{blue}{+2.07}) \\ 
        bear & 94.52 $\pm$ 0.42 (\textcolor{gray}{+0.23}) & laptop & 81.45 $\pm$ 0.32 (\textcolor{blue}{+1.23}) \\ 
        zebra & 97.74 $\pm$ 0.15 (\textcolor{gray}{-0.05}) & cell phone & 42.47 $\pm$ 0.28 (\textcolor{blue}{+1.62}) \\ 
        giraffe & 98.57 $\pm$ 0.07 (\textcolor{gray}{+0.04}) & oven & 80.42 $\pm$ 0.67 (\textcolor{blue}{+3.85}) \\ 
        backpack & 38.29 $\pm$ 0.62 (\textcolor{blue}{+2.89}) & sink & 85.79 $\pm$ 0.33 (\textcolor{blue}{+1.90}) \\ 
        umbrella & 77.06 $\pm$ 0.11 (\textcolor{blue}{+1.66}) & refrigerator & 70.36 $\pm$ 0.27 (\textcolor{blue}{+1.82}) \\ 
        suitcase & 62.87 $\pm$ 0.63 (\textcolor{blue}{+2.57}) & book & 57.69 $\pm$ 0.18 (\textcolor{blue}{+2.52}) \\ 
        kite & 92.27 $\pm$ 0.05 (\textcolor{blue}{+0.67}) & clock & 75.02 $\pm$ 0.21 (\textcolor{blue}{+1.27}) \\ 
        baseball bat & 88.22 $\pm$ 0.42 (\textcolor{gray}{+0.23}) & vase & 71.48 $\pm$ 0.43 (\textcolor{blue}{+2.33}) \\ 
        surfboard & 93.21 $\pm$ 0.03 (\textcolor{blue}{+0.64}) & scissors & 37.59 $\pm$ 1.02 (\textcolor{red}{-2.82}) \\ 
        bottle & 58.81 $\pm$ 0.14 (\textcolor{blue}{+3.51}) & teddy bear & 76.43 $\pm$ 0.70 (\textcolor{blue}{+0.61}) \\ 
        \bottomrule
    \end{tabular}}
    \caption{\textbf{Model accuracy on each unbiased class.}
    We report AP values on each unbiased class from DecoupleGen, and change in AP values compared to standard classifier trained on original dataset, where \textcolor{blue}{blue} indicates improvement, \textcolor{red}{red} indicates drop, and \textcolor{gray}{gray} indicates similar performance within error bars.
    Model performance on most classes improve from standard classifier, even though these classes are not explicitly addressed in our approach.}
    \label{tab:unbiased_accs}
\end{table}

